\documentclass[11pt]{article}

\usepackage[preprint]{acl}

\usepackage{times}
\usepackage{latexsym}

\usepackage[T1]{fontenc}

\usepackage[utf8]{inputenc}

\usepackage{microtype}

\usepackage{inconsolata}

\usepackage{graphicx}
\usepackage{multirow}
\usepackage[normalem]{ulem}

\usepackage{hyperref}
\usepackage{url}
\usepackage{xspace}

\usepackage{booktabs}

\usepackage{listings}

\usepackage{enumitem}

\usepackage{comment}
\usepackage{titlesec}

\usepackage{soul}
\soulregister\ref{7} %
\soulregister\cite{7} %

\usepackage{xcolor} 

 \usepackage{amsmath} 

\definecolor{codegreen}{rgb}{0,0.6,0}
\definecolor{codegray}{rgb}{0.5,0.5,0.5}
\definecolor{codepurple}{rgb}{0.58,0,0.82}
\definecolor{backcolour}{rgb}{0.95,0.95,0.92}

\lstdefinestyle{mystyle}{
    backgroundcolor=\color{backcolour},   
    commentstyle=\color{codegreen},
    keywordstyle=\color{magenta},
    numberstyle=\tiny\color{codegray},
    stringstyle=\color{codepurple},
    basicstyle=\ttfamily\scriptsize,
    breakatwhitespace=false,         
    breaklines=true,                 
    captionpos=b,                    
    keepspaces=true,                 
    numbers=left,                    
    numbersep=5pt,                  
    showspaces=false,                
    showstringspaces=false,
    showtabs=false,                  
    tabsize=2
}

\lstdefinelanguage{yaml}{
  keywords={true,false,null,y,n},
  comment=[l]{\#},
  morecomment=[s]{/*}{*/},
  morestring=[b]",
  morestring=[b]',
  sensitive=true
}

\lstdefinelanguage{python}{
  keywords={def, return, if, else, for, while, break, continue, import, from, as, class, try, except, finally, with, yield, pass, lambda, and, or, not, is, in, True, False, None},
  keywordstyle=\color{magenta},
  ndkeywords={self},
  ndkeywordstyle=\color{codegreen},
  comment=[l]{\#},
  morecomment=[s]{"""}{"""},
  morestring=[b]',
  morestring=[b]",
  sensitive=true
}

\lstdefinelanguage{json}{
    basicstyle=\ttfamily\footnotesize,
    numbers=none,
    showstringspaces=false,
    breaklines=true,
    backgroundcolor=\color{gray!5},
    literate=
     *{0}{{{\color{blue}0}}}{1}
      {1}{{{\color{blue}1}}}{1}
      {2}{{{\color{blue}2}}}{1}
      {3}{{{\color{blue}3}}}{1}
      {4}{{{\color{blue}4}}}{1}
      {5}{{{\color{blue}5}}}{1}
      {6}{{{\color{blue}6}}}{1}
      {7}{{{\color{blue}7}}}{1}
      {8}{{{\color{blue}8}}}{1}
      {9}{{{\color{blue}9}}}{1}
      {:}{{{\color{black}:}}}{1}
      {,}{{{\color{black},}}}{1}
      {"}{{{\color{red}"}}}{1}
      {true}{{{\color{magenta}true}}}{1}
      {false}{{{\color{magenta}false}}}{1}
      {null}{{{\color{magenta}null}}}{1}
}

\newcommand{\bench}{Thunder-NUBench\xspace}

\title{\bench: A Benchmark for LLMs'\\ Sentence-Level Negation Understanding}

\author{
\textbf{Yeonkyoung So\textsuperscript{1}}, 
\textbf{Gyuseong Lee\textsuperscript{1}}, 
\textbf{Sungmok Jung\textsuperscript{1}}, 
\textbf{Joonhak Lee\textsuperscript{1}}, \\
\textbf{JiA Kang\textsuperscript{1}}, 
\textbf{Sangho Kim\textsuperscript{1}},  
\textbf{Jaejin Lee\textsuperscript{1,2}} \\ \\ 
\textsuperscript{1}Graduate School of Data Science, Seoul National University \\
\textsuperscript{2}Dept. of Computer Science and Engineering, Seoul National University \\
\texttt{\{kathy1028,ksnannaya,tjdahrwjd,hmjelee,jia6776,ksh4931,jaejin\}@snu.ac.kr}
}

\begin{document}
\maketitle

\begin{abstract}
Negation is a fundamental linguistic phenomenon that poses ongoing challenges for Large Language Models (LLMs), particularly in tasks requiring deep semantic understanding. Current benchmarks often treat negation as a minor detail within broader tasks, such as natural language inference. Consequently, there is a lack of benchmarks specifically designed to evaluate comprehension of negation. In this work, we introduce \textit{\bench} — a novel benchmark explicitly created to assess sentence-level understanding of negation in LLMs. \bench goes beyond identifying surface-level cues by contrasting standard negation with structurally diverse alternatives, such as local negation, contradiction, and paraphrase. This benchmark includes manually created sentence-negation pairs and a multiple-choice dataset, allowing for a comprehensive evaluation of models' understanding of negation. 
\end{abstract}

\section{Introduction}
\label{sec:1_introduction}
Negation is a fundamental and universal phenomenon found in languages worldwide. It is closely associated with various human communicative abilities, such as denial, contradiction, deception, misrepresentation, and irony. Although affirmative statements are more common, negation still plays a significant role in language; approximately 25\% of sentences in English texts contain some form of negation~\citep{sarabi2016understanding, hossain2020analysis, sep-negation}. This prevalence and its impact on meaning make accurate interpretation of negation crucial for several natural language processing (NLP) tasks, including sentiment analysis, question answering, knowledge base completion, and natural language inference (NLI)~\citep{khandelwal2020negbert, hosseini2021understanding, singh2023nlms}. Recent studies have shown that effectively managing negation is important even for multi-modal language models~\citep{quantmeyer2024and, alhamoud2025vision, park2025know}.

\begin{figure}[t]
    \begin{center}
    \includegraphics[width=1\columnwidth]{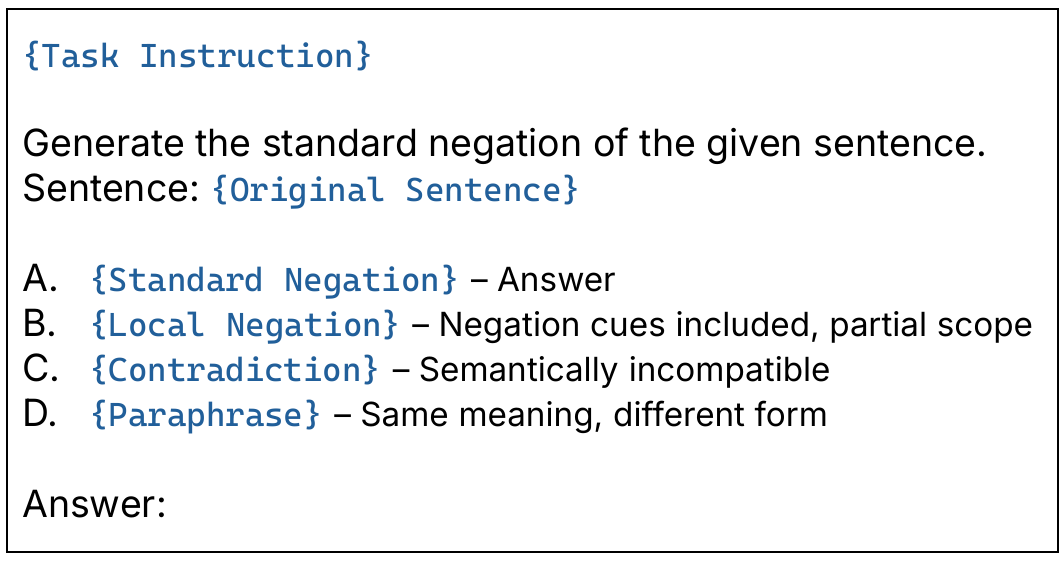}
    \caption{Illustration of the \bench task. Only the standard negation reverses the truth value of the original sentence, while other options differ in scope or semantics.}
    \vspace*{-1\baselineskip}
    \label{fig:intro_example}
    \end{center}
\end{figure}

Meanwhile, negation poses significant challenges for both humans and language models. Research shows that people often find it more difficult to process and comprehend negated statements compared to affirmative ones~\citep{wales1969so, sarabi2016understanding}. Similarly, many studies indicate that pretrained language models (PLMs) struggle to interpret negation accurately. For example, models like BERT~\citep{devlin2019bert} and even large language models (LLMs) such as GPT-3~\citep{brown2020language} often have difficulty distinguishing between negated and affirmative statements. These models tend to rely on superficial cues, which can result in incorrect outputs when negation is involved~\citep{kassner2020negated, hossain2022analysis, truong2023language}.

Despite its significance, there is a notable lack of dedicated evaluation benchmarks for understanding negation. Most existing resources either treat negation as a minor aspect of broader tasks or focus solely on narrow syntactic detection, often emphasizing encoder-based models~\citep{hossain2020analysis, geiger2020neural, truong2022not, anschutz2023not}. To address this gap, we introduce \textit{\bench} (\textit{N}egation \textit{U}nderstanding \textit{Bench}mark), a dataset explicitly designed to evaluate LLMs' sentence-level comprehension of negation.\footnote{\bench is publicly available at \url{https://huggingface.co/datasets/thunder-research-group/SNU_Thunder-NUBench}.} Our benchmark is structured as a multiple-choice question (MCQ) task: given an original sentence, the model must select the correct standard negation from four options. The other three choices (local negation, contradiction, and paraphrase) are carefully designed distractors that test whether models truly grasp semantic scope and logical oppositions.

The contributions of this paper are summarized as follows:
\begin{itemize}[nosep]
    \item We define standard negation within the framework of sentential logic. Grounding standard negation in logical structure not only clarifies its role in natural language but also supports the evaluation and enhancement of reasoning in LLMs.
    \item We introduce a manually created benchmark that includes a dataset of sentence-negation pairs for fine-tuning, along with a multiple-choice evaluation task.
    \item We conduct systematic evaluations of decoder-based LLMs across model families, scales, and training methods to analyze variation in negation understanding.
\end{itemize}

\bench provides valuable insights into the semantic reasoning abilities of language models and serves as a robust standard for future research focused on understanding negation. 

\section{Related Work}
\label{sec:2_related}

\begin{table*}[htbp]
\centering
\resizebox{\textwidth}{!}{%
\begin{tabular}{@{}ccll@{}}
\toprule
\textbf{Dimension} &
  \textbf{Negation Type} &
  \multicolumn{1}{c}{\textbf{Definition}} &
  \multicolumn{1}{c}{\textbf{Example}} \\ \midrule
\multirow{2}{*}{\textbf{Scope}} &
  \textbf{\begin{tabular}[c]{@{}c@{}}
   Clausal Negation \\ 
   ( = Sentential \\
   Negation)\end{tabular}} &
  \begin{tabular}[c]{@{}l@{}}Negation that applies to the entire clause or \\ sentence. This typically involves the use of "not", \\ or its contracted form "n't" with auxiliary verbs.\end{tabular} &
  \begin{tabular}[c]{@{}l@{}}He \textbf{speaks} English fluently.  \\ $\rightarrow$ He \textbf{doesn't speak} English fluently.\end{tabular} \\ \cmidrule(l){2-4} 
 &
  \textbf{\begin{tabular}[c]{@{}c@{}}
  Subclausal Negation \\ 
  ( = Constituent / \\ Local Negation)\end{tabular}} &
  \begin{tabular}[c]{@{}l@{}}Negation that focuses on negating a specific \\ part of a clause, such as a word or phrase, \\ rather than the entire clause.\end{tabular} &
  \begin{tabular}[c]{@{}l@{}}He speaks English \textbf{fluently}.\\ $\rightarrow$ He speaks English, \textbf{but}  \textbf{not}  \textbf{fluently}.\end{tabular} \\ \midrule
\multirow{2}{*}{\textbf{Form}} &
\begin{tabular}[c]{@{}c@{}}
\textbf{Morphological} \\ 
\textbf{Negation} \end{tabular}
  &
  \begin{tabular}[c]{@{}l@{}}Negation expressed through affixes attached to words \\ 
  such as prefixes like "un-", "in-", "dis-", or suffixes  \\ 
  like "-less".\end{tabular} &
  \begin{tabular}[c]{@{}l@{}}She is \textbf{happy}.\\ $\rightarrow$ She is \textbf{unhappy}.\end{tabular} \\ \cmidrule(l){2-4} 
 &
  \textbf{Syntactic Negation} &
  \begin{tabular}[c]{@{}l@{}}Negation expressed through separate words (particles) \\ 
  in the syntax, such as "not", "never", "no", etc.\end{tabular} &
  \begin{tabular}[c]{@{}l@{}}She is happy.\\ $\rightarrow$ She is not happy.\end{tabular} \\ \midrule
\multirow{2}{*}{\textbf{Target}} &
  \textbf{Verbal Negation} &
    \begin{tabular}[c]{@{}l@{}}  
    Negation that applies directly to the verb \\
    or verb phrase. \end{tabular} 
&
  \begin{tabular}[c]{@{}l@{}}They \textbf{have finished} the work.\\ $\rightarrow$ They \textbf{have not finished} the work.\end{tabular} \\ \cmidrule(l){2-4} 
 &
 \begin{tabular}[c]{@{}c@{}}
\textbf{Non-verbal } \\ 
\textbf{Negation} \end{tabular}
&
  Negation that negates elements other than the verb. &
  \begin{tabular}[c]{@{}l@{}}There is \textbf{milk} in the fridge.\\ $\rightarrow$ There is \textbf{no milk} in the fridge.\end{tabular} \\ \bottomrule
\end{tabular}%
}
\caption{Typology of negation.}
\vspace*{-0.5\baselineskip}
\label{tab:negtype}
\end{table*}

\paragraph{Negation detection and scope resolution.}
Early work in negation detection and scope resolution primarily relied on rule-based systems and handcrafted heuristics, especially in domain-specific contexts like clinical texts. While these systems are effective, they lack flexibility across different domains~\citep{chapman2001simple, de2012ucm, ballesteros2012ucm, basile2012ugroningen}. Traditional machine learning methods, such as Support Vector Machines (SVMs)~\citep{hearst1998support} and Conditional Random Fields (CRFs)~\citep{sutton2012introduction}, were introduced later; however, they too are limited to narrow domains~\citep{morante2008learning, morante2009metalearning, read2012uio1, li2018learning}.

More recently, deep learning approaches employing Convolutional Neural Networks (CNNs)~\citep{o2015introduction} and Bidirectional Long Short-Term Memory (BiLSTM) networks~\citep{siami2019performance} have enhanced performance by providing improved contextual embeddings and sequence modeling~\citep{fancellu2016neural, bhatia2019end}. Pretrained transformer models like BERT have been employed through transfer learning techniques (e.g., NegBERT~\citep{khandelwal2020negbert}), significantly increasing the accuracy of negation detection tasks. Nonetheless, these methods still largely focus on syntactic span detection, leaving deeper semantic understanding of negation a challenging area to tackle.

\paragraph{Negation-sensitive subtasks of NLU.}
Negation understanding has become increasingly important in natural language understanding (NLU) tasks~\citep{hosseini2021understanding}. However, existing NLU benchmarks, such as SNLI~\citep{bowman2015large} for natural language inference (NLI), CommonsenseQA~\citep{talmor2019commonsenseqa} for Question Answering (QA), SST-2~\citep{socher2013recursive} for sentiment analysis, STS-B~\citep{cer2017semeval} for textual similarity and paraphrasing, have been criticized for not adequately addressing the semantic impact of negation~\citep{hossain2022analysis, rezaei2024paraphrasing}. These datasets contain relatively few instances of negation or include negations that are not crucial to task performance, allowing language models to achieve high accuracy even when they completely ignore negation.

Recent studies, including NegNLI~\citep{hossain2020analysis}, MoNLI~\citep{geiger2020neural}, NaN-NLI~\citep{truong2022not}, NoFEVER-ML and NoSNLI-ML~\citep{vrabcova-etal-2025-towards}, have introduced benchmarks for NLU that includes negation. These studies show that model performance significantly declines when negation plays a crucial role in affecting the outcome~\citep{naik2018stress, yanaka2019can, hartmann2021multilingual, hossain2022question, hossain2022leveraging, she2023scone, anschutz2023not}. These findings suggest that current language models tend to depend on superficial linguistic patterns rather than a genuine understanding of semantics. 

\paragraph{Limitations of distributional semantics.}
\label{sec:2_3_distribution}
Distributional semantics~\citep{harris1954distributional, sahlgren2008distributional} aims to create models that learn semantic representations based on patterns of word co-occurrences~\citep{boleda2020distributional, lenci2022comparative} and capture broad semantic relationships; however, it encounters significant challenges with negation. Negated expressions, such as "not good," often appear in similar contexts as their affirmative counterparts, like "good." As a result, models tend to generate similar vector representations for these expressions, despite their opposing meanings. Previous research has pointed out this limitation, showing that PLMs struggle to capture the subtle semantic differences introduced by antonyms and the reversal of polarity~\citep{rimell2017learning, jumelet2018language, niwa2021predicting, jang2022beyond, vahtola2022not}. Studies have further suggested that models like BERT find it difficult to distinguish between affirmative and negated contexts~\citep{kassner2020negated, ettinger2020bert}.

\paragraph{Negations in generative language models.}
Recent research on understanding negation has primarily focused on bidirectional models, such as BERT~\citep{devlin2019bert} and RoBERTa~\citep{liu2019roberta}, which have demonstrated strong performance in NLU and negation detection tasks. However, with the emergence of generative foundation models like GPT~\citep{radford2018improving} and LLaMA~\citep{touvron2023llama}, attention has shifted towards evaluating how these models handle negation. Studies have shown that these generative models often exhibit a positive bias and struggle with producing or interpreting negated statements~\citep{truong2023language, chen2023say, garcia2023not}. Although some benchmarks, such as CONDAQA~\citep{ravichander2022condaqa} and ScoNe~\citep{she2023scone}, reveal these limitations, there is still a lack of robust evaluation resources specifically designed for negation understanding of generative models. 

\paragraph{ }
Building on previous studies, this paper focuses on sentence-level negation as a core linguistic operation. While prior negation-related NLI or QA benchmarks evaluate negation within broader inference or comprehension tasks, \bench specifically tests whether models can distinguish standard negation, the logical reversal of the sentence, from closely related distractors. As a result, \bench enables a more direct examination of sentence-level negation than is afforded by NLI or QA benchmarks, where negation is evaluated only as part of broader inference tasks.

\begin{table*}[htbp]
\centering
\resizebox{0.9\textwidth}{!}{%
\begin{tabular}{@{}ccp{12cm}@{}}
\toprule
\textbf{Type}      & \multicolumn{2}{l}{\textbf{Definition}}                                                                         \\ \midrule\midrule
\textbf{Base case} & \multicolumn{2}{l}{\begin{tabular}[c]{@{}l@{}}If $P$ is an atomic proposition, $\text{Neg}(P)$ is the proposition where the main predicate of $P$ is negated.\end{tabular}} \\ \midrule
\multirow{3}{*}{\textbf{Inductive step}} &
  \textbf{Conjunction} &
  \begin{tabular}[c]{@{}l@{}}$\text{Neg}(P \;\text{and}\; Q) \equiv \text{Neg}(P) \;\text{or}\; \text{Neg}(Q)$,\hspace{1em} $\text{Neg}(P \;\text{but}\; Q) \equiv \text{Neg}(P) \;\text{or}\; \text{Neg}(Q)$ \end{tabular} \\ \cmidrule(l){2-3} 
          & \textbf{Disjunction}    & $\text{Neg}(P \;\text{or}\; Q) \equiv \text{Neg}(P) \;\text{and}\; \text{Neg}(Q)$   \\ \cmidrule(l){2-3} 
 &
  \textbf{Implication} &
  \begin{tabular}[c]{@{}l@{}}$\text{Neg}(\text{if } P, Q) \equiv \text{Neg}(\text{Neg}(P) \;\text{or}\; Q) \equiv P \;\text{and}\; \text{Neg}(Q)$\\ $\text{Neg}(P \;\text{if and only if}\; Q) \equiv \text{Neg}(\text{if } P, Q \;\text{and}\; \text{if } Q, P)$\end{tabular} \\ 
\bottomrule
\end{tabular}%
}
\caption{Standard negation. $P$ and $Q$ stand for propositions. In addition to \textit{and}, \textit{or}, and \textit{if}, other natural language connectives such as \textit{when} are also considered, and their negations follow the same principles depending on their function.}
\vspace*{-0.5\baselineskip}
\label{tab:definition}
\end{table*}
\section{Scope and Categorization of Negation}
\label{sec:3_negation}
In this work, we aim to clarify the concept of negation by introducing a typology that clearly outlines its semantic boundaries and differentiates it from related, yet distinct, phenomena. This typology organizes various forms of meaning reversal into logically consistent categories, allowing for a more precise and systematic evaluation of how language models handle negation. 

\subsection{Typology of Negation}

Negation is a fundamental semantic and syntactic operation found in natural languages, used to convey denial, rejection, or the absence of a proposition. Hereafter, we denote our negation operation for a sentence $S$ as $\text{Neg}(S)$. In formal logic, negation flips the truth value of a proposition $P$: if $P$ is true, then $\text{Neg}(P)$ is false, and vice versa. Semantically, negation creates a binary opposition between a proposition and its affirmative counterpart, meaning that each one is the opposite of the other~\citep{sep-negation}.

Negation can be categorized along several dimensions: scope, form, and target (see Table~\ref{tab:negtype}). In terms of scope, negation may affect the entire clause (referred to as \textit{clausal negation}) or only part of it (known as \textit{subclausal negation}). Regarding form, negation can manifest as bound morphemes, such as prefixes and suffixes (\textit{morphological negation}), or as separate syntactic elements like "not" or "never" (\textit{syntactic negation}). Finally, depending on its target, negation can apply to the verb (\textit{verbal negation}) or to other elements in the sentence (\textit{non-verbal negation})~\citep{zanuttini2001sentential, miestamo2007negation, truong2022not, kletz2023self}.

\subsection{Negation and Contradiction}
Negation and contradiction are closely related concepts that are often conflated in NLP research~\citep{jiang2021m}. Contradiction refers to the incompatibility of two propositions, meaning that they cannot both be true at the same time. While negation frequently serves as a primary mechanism for creating contradictions (by reversing the truth value of a proposition), contradictions can also arise from antonymy, numeric mismatches, or differences in structure and lexicon (further details can be found in Appendix~\ref{appendix:contradiction}). For instance, the statements "An individual was born in France" and "An individual was born in Italy" are contradictory, but they are not negations, as the second statement does not reverse the truth of the first. 

Many previous studies have overlooked the possibility that contradictions can exist independently of explicit negation. Recognizing this gap, we specifically examine the ability of LLMs to differentiate between negations and non-negated contradictions, highlighting the nuanced semantic distinctions that are involved.

\begin{table*}[htbp]
\begin{center}
\resizebox{\textwidth}{!}{%
\begin{tabular}{@{}cll@{}}
\toprule
\textbf{Type} & \multicolumn{1}{c}{\textbf{Structure Explanation}} & \multicolumn{1}{c}{\textbf{Local Negation Example}} \\ \midrule 
\textbf{\begin{tabular}[c]{@{}c@{}}
Relative clause \\ 
negation\end{tabular}} &
  \begin{tabular}[c]{@{}l@{}}
  A relative clause is a type of dependent clause that gives \\ 
  extra details about a noun or noun phrase 
  in the main sentence. \\ It usually begins with a relative pronoun such as \\
  \textit{who, which, that, whom,} or \textit{whose}.\end{tabular} &
  \begin{tabular}[c]{@{}l@{}}
  The man \underline{who \textbf{owns} the car}
  is my \\ \hspace{1em} neighbor. \\ $\rightarrow$ The man 
  \underline{who \textbf{does} \textbf{not own} the car}\\ \hspace{2em} is my neighbor.\end{tabular} \\ \midrule
\textbf{\begin{tabular}[c]{@{}c@{}}Participle clause \\ negation\end{tabular}} &
  \begin{tabular}[c]{@{}l@{}}
  A participle clause is a type of dependent clause that begins with \\ a participle 
  (a verb form ending in \textit{-ing} or a past participle). \\ It acts like an adverb, giving extra details about the main \\ clause, 
  often showing time, reason, result, or sequence of actions.
  \end{tabular} &
  \begin{tabular}[c]{@{}l@{}}
  \underline{\textbf{Walking} through the park}, 
  she found \\ \hspace{1em} a lost wallet. \\ $\rightarrow$ \underline{\textbf{Not}} \underline{\textbf{walking} through 
  the park}, she \\
  \hspace{2em} found a lost wallet.\end{tabular} \\ \midrule
\textbf{\begin{tabular}[c]{@{}c@{}}
Adverbial clause \\ 
negation\end{tabular}} &
  \begin{tabular}[c]{@{}l@{}}
  An adverbial clause is a dependent clause that acts like an adverb, \\ 
  modifying a verb, adjective, or adverb. It gives information such \\ 
  as time, reason, condition, or contrast. These clauses are introduced \\
  by subordinating conjunctions like \textit{because, although}, or \textit{while}.\end{tabular} &
  \begin{tabular}[c]{@{}l@{}}She stayed inside \underline{because it \textbf{was}} \\
  \hspace{1em} \underline{\textbf{raining}}. \\ $\rightarrow$ She stayed inside \underline{because it \textbf{was}}\\  \hspace{2em}\underline{\textbf{not raining}}.\end{tabular} \\ \midrule
\textbf{\begin{tabular}[c]{@{}c@{}}
Compound \\
sentence \\ 
with \\
local negation\end{tabular}} &
  \begin{tabular}[c]{@{}l@{}}
  A compound sentence consists of two or more main clauses \\ joined by coordinating conjunctions such as \textit{and, but}, or \textit{or}. \\ If only one of these clauses is negated, the negation applies \\ only locally to that clause.\end{tabular} &
  \begin{tabular}[c]{@{}l@{}}He submitted the report and \\ \hspace{1em} \underline{\textbf{attended}
  the meeting}. \\ $\rightarrow$ He  
  submitted the 
  report and \textbf{did} \\
  \hspace{2em} \underline{\textbf{not attend} the meeting}.\end{tabular} \\ \bottomrule
\end{tabular}%
}
\end{center}
\caption{Typology of local negation.}
\vspace*{-0.5\baselineskip}
\label{tab:localtype}
\end{table*}

\subsection{Standard Negation}
\label{sec:3_3_standard_negation}
\textit{Standard negation} refers to the typical form of negation applied to the declarative verbal main clause. It specifically negates the verb in a \textit{main clause}~\citep{miestamo2000towards}. A main clause can function as a complete sentence on its own, consisting at a minimum of a subject and a predicate. This definition is grounded in the notion that the verb acts as the head of the clause~\citep{miller2011critical}. 

Building on this traditional understanding, we treat standard negation as \textit{the process of reversing the truth value of the verb phrase in the main clause}, which we will refer to as the \textit{main predicate} in this paper. A verb phrase is headed by a verb and can consist of a single verb or a combination of auxiliaries, complements, and modifiers (e.g., "will call" and "is being promoted")~\citep{lakoff1966criterion}. Since the main predicate conveys the core action or state of the clause, negating it effectively reverses the proposition of the entire sentence. In this context, this paper treats standard negation as a \textit{truth-functional operation that maps the main predicate to its complement set within the semantic space}.

We further clarify the scope of standard negation within the typology presented in Table~\ref{tab:negtype}. Standard negation includes both \textit{clausal negation} and \textit{verbal negation}, as it reverses the meaning of the entire sentence by negating the main predicate. In terms of form, standard negation can employ both \textit{syntactic} and \textit{morphological negation}. Syntactically, standard negation often uses explicit negation particles, such as "not." Morphologically, it can involve \textit{complementary antonyms} (for example, "alive" vs. "dead" or "true" vs. "false"), which occupy mutually exclusive semantic spaces, thus reversing the truth value of the proposition. In contrast, other types of antonyms, such as \textit{gradable antonyms} (e.g., "happy" vs. "unhappy") and \textit{relational antonyms} (e.g., "buy" vs. "sell")~\citep{lehrer1982antonymy}, do not strictly reverse truth values. Therefore, they are classified as contradictions rather than standard negation in this paper.

\paragraph{Atomic and Complex Propositions.}
While this characterization of standard negation effectively defines standard negation for atomic propositions (elementary sentences that cannot be further decomposed)~\citep{davis2004semantics}, its application to complex sentences with multiple clauses requires a more thorough approach. In this paper, we treat an atomic proposition as \textit{a sentence that contains a single main predicate}. Specifically, for propositions composed of multiple logically connected atomic statements, the method for reversing the truth value of the entire complex proposition can be ambiguous. In natural language, such logical structures typically appear as coordinated clauses (e.g., "$P$ and $Q$ or $R$") or comma-separated lists connected by "and" or "or" (e.g., "$P, Q,$ and $R$"). We treat these as equivalent to a sequence of binary conjunctions or disjunctions. 

\paragraph{Definition of standard negation.} 
In this paper, standard negation refers to natural-language sentential negation, which is formally treated as logical negation within the framework of sentential logic~\citep{enderton2001}. To address the complexities involved, we define standard negation recursively by applying it pairwise over the logical structure of a sentence until only atomic propositions remain, ensuring that the truth value of the entire sentence is reversed even when it contains multiple coordinated clauses. Our definition of $\text{Neg}(\cdot)$ is presented in Table~\ref{tab:definition}. 
Conditionals of the form "if $P$, $Q$" are equivalent to "$\text{Neg}(P) \;\text{or}\; Q$" in logic, and we adhere to this equivalence when defining their negation (more details can be found in Appendix~\ref{appendix:implication}).

\begin{figure*}[htbp]
  \centering
  \includegraphics[width=0.9\textwidth]{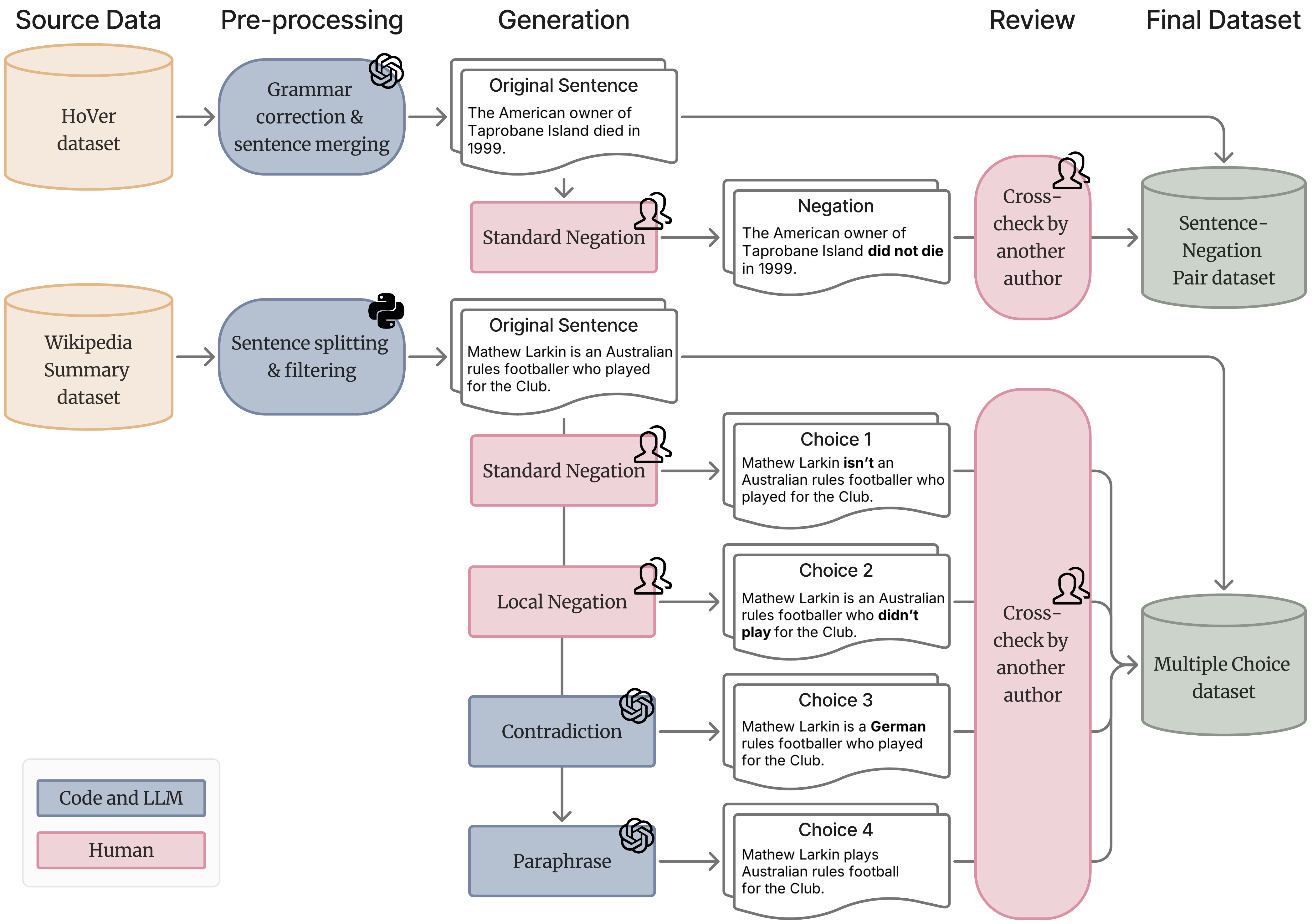}
  \caption{Dataset generation process.}
  \label{fig:data-const}
\end{figure*}

\subsection{Local Negation}
We define \textit{local negation} as a form of negation that specifically targets a verb phrase outside the main clause. While the term is often used interchangeably with subclausal negation, our focus is solely on local negation relating to subclausal and verbal negation. This concept applies to four types of sentence structures: relative clauses, participle clauses, adverbial clauses, and compound sentences (refer to Table~\ref{tab:localtype} for more details). 

In particular, conditional clauses, such as the "if $P$" part in "if $P$, $Q$" are categorized as adverbial clauses. In compound sentences, standard negation requires all main clauses to be negated in order to achieve sentence-level negation. If only a subset of the clauses is negated, this is considered local negation.

Local negation, in terms of structure, resembles standard negation, typically using explicit negation markers like "not." However, its scope is confined to a specific part of the sentence rather than encompassing the entire main clause. Because explicit cues such as "not" are still present, models that depend on shallow cue detection may be misled, failing to distinguish between standard negation and local negation.

\section{\bench Dataset}
\label{sec:4_benchmark}

We construct the \bench dataset through three main stages: (1) pre-processing, (2) generation, and (3) review. The overall workflow is illustrated in Figure~\ref{fig:data-const}.

\paragraph{Pre-processing.}
We begin by extracting sentences from two primary corpora: (1) the Hover dataset~\citep{jiang2020hover}, designed for multi-hop fact extraction and claim verification, and (2) the Wikipedia Summary dataset~\citep{scheepers2017compositionality}, which contains concise summaries from English Wikipedia. We chose these datasets because their factual content and complex sentence structures are well-suited for developing a dataset aimed at understanding standard negation in complex, sufficiently lengthy sentences. Additionally, we automatically correct any grammatical errors and merge or split sentences as needed to create well-formed single-sentence units.

\paragraph{Generation.}
We create two types of datasets from the pre-processed sentences: the \textit{sentence-negation pair dataset} and the \textit{multiple choice dataset}. In the sentence-negation pair dataset, each original sentence is paired with a manually crafted standard negation, as detailed in Section~\ref{sec:3_3_standard_negation}. In the multiple-choice dataset, each original sentence is presented with four options: a standard negation, a local negation, a contradiction, and a paraphrase. Each of them is described in Table~\ref{tab:category}. Together, these categories assess whether models truly understand semantic negation rather than relying on superficial cues.

Standard and local negation options are manually created rather than generated by LLMs. We have observed that LLMs often struggle to produce correct standard negations, frequently resulting in subclausal or local negations instead. They can also generate incorrect local negations, even when explicitly prompted to do otherwise. Since precise negation is essential to our benchmark, these options must be developed by humans to ensure the quality of the dataset. In contrast, contradiction and paraphrase options are initially created automatically using carefully designed prompts with the OpenAI API~\citep{OpenAI} and are then refined during the review process. Details of the GPT models and code used at each stage of the data generation process are provided in Appendix~\ref{appendix:datacode}.

\begin{table*}[htbp]
\centering
\resizebox{\textwidth}{!}{%
\begin{tabular}{lp{16cm}}
\hline
\textbf{Category} & \textbf{Description} \\ \hline
\textbf{Standard Negation} 
& This category involves reversing the truth value of the main clause, which is the primary focus of the benchmark.\\\hline
\textbf{Local Negation} 
& In this case, negation is applied to a subordinate clause or a partial structure, which does not reverse the entire sentence.
\\\hline
\textbf{Contradiction} 
& This category introduces conflicts with the original meaning through semantic changes, such as the use of antonyms, different numbers, or other entities, without employing explicit negation.
\\\hline
\textbf{Paraphrase}
& Here, the original meaning is preserved while the surface form is altered. Examples of paraphrases are intentionally constructed to vary the sentence structure and word choice significantly, ensuring that no additional information is added. As a result, the original sentence still entails its paraphrase. This category tests whether models mistakenly consider different surface forms as meaning reversals, even when the semantic meanings remain equivalent.\\\hline
\end{tabular}
}
\caption{Multiple choice categories included in \bench.}
\label{tab:category}
\end{table*}

\paragraph{Review.}
All constructed data undergo a multi-stage human review process (see Appendix~\ref{appendix:review_protocol}). A different author, separated from the creator, cross-checks each instance, and any disagreements are addressed in regular meetings to ensure consistency. Options for contradictions are reviewed only after the corresponding standard and local negations are finalized, as they must not overlap semantically. Consequently, the earlier negations are re-examined during the contradiction review and are cross-checked by multiple authors.

The guidelines for data generation and review are continuously updated, and any previously created data are revised accordingly (see Appendix~\ref{appendix:guidelines}). This protocol ensures rigorous quality control and consistency throughout the benchmark.

\paragraph{Dataset statistics.}
The final dataset includes a training set of sentence-negation pairs and a multiple-choice evaluation set (see Table~\ref{tab:data-stat}). For few-shot prompting, we construct a demonstration set of 50 examples. These are carefully selected to have unique Wikipedia page indices to avoid any overlap with the test set. Furthermore, to provide the model with a balanced overview of the task, we match the distribution of local negation types (\texttt{\small choice2\_type}) in the demonstration set to that of the overall dataset. This ensures that the demonstrations are representative and prevents the model from developing a biased strategy for specific negation types.

\begin{table}[htbp]
\centering
\resizebox{0.8\columnwidth}{!}{%
\begin{tabular}{@{}ccr@{}}
\toprule
\textbf{Dataset} & \textbf{Split} & \textbf{Count} \\ \midrule \midrule
Sentence-Negation & Train      & 3,772 \\ \midrule
Multiple Choice   & Demonstration & 50   \\
                  & Test       & 1,261 \\ \midrule
                  & Total      & 5,083 \\ \cmidrule(l){2-3}
\end{tabular}%
}
\caption{\bench statistics.}
\vspace{-1\baselineskip}
\label{tab:data-stat}
\end{table}
\section{Experiments}
\label{5_experiments}

\subsection{Evaluation Setup}
We evaluate models under two common Multiple-Choice Question Answering (MCQA) settings: (1) a completion-based evaluation, where the model assigns probabilities to each candidate by appending it as a continuation of the prompt, and (2) an option-selection evaluation, where the model selects from labeled options (A/B/C/D). To mitigate known issues such as selection or position bias in the option-selection setting, we randomly shuffle the order of the options (using random seed 42). In both settings, we use two instruction variants: a definition-based instruction (referred to as the \textit{definition} instruction) and a step-by-step instruction (referred to as the \textit{detailed} instruction). Details of the prompt templates and formatting are provided in Appendix~\ref{appendix:prompt_select}.

We report results for a diverse set of pretrained and instruction-tuned models, evaluated under zero-shot and few-shot settings with 1, 5, and 10 examples, as well as supervised fine-tuning (SFT). For SFT, models are trained on the sentence-negation pair dataset introduced in \bench. The full list of models, evaluation protocols, and fine-tuning details is provided in Appendix~\ref{appendix:details_eval}.

\subsection{Overall Performance and Key Findings}
\label{sec:5_2_performance}

\begin{figure}[htbp]
    \begin{center}
    \includegraphics[width=1\columnwidth]{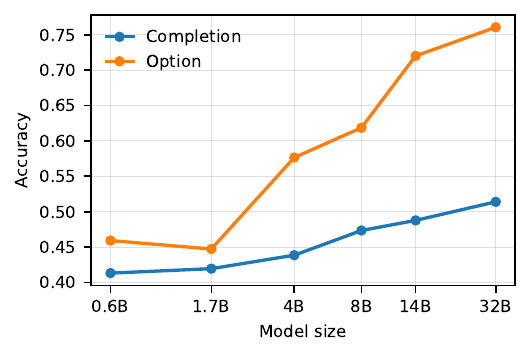}
    \vspace*{-1\baselineskip}
    \caption{Zero-shot accuracy of instruction-tuned Qwen3 models under the detailed prompt setting.}
    \label{fig:model_size}
    \vspace*{-1\baselineskip}
    \end{center}
\end{figure}

\paragraph{Effect of Model Size.}
Performance tends to improve as model size increases, indicating that negation understanding benefits from larger models. The improvement is more pronounced in the option-selection evaluation than in the completion-based setting, with larger models showing higher accuracy gains (as seen in Figure~\ref{fig:model_size}). This suggests that option selection benefits more from larger models, since it requires comparing multiple candidate answers rather than generating a single continuation.

\begin{figure}[b!]
    \begin{center}
    \includegraphics[width=1\columnwidth]{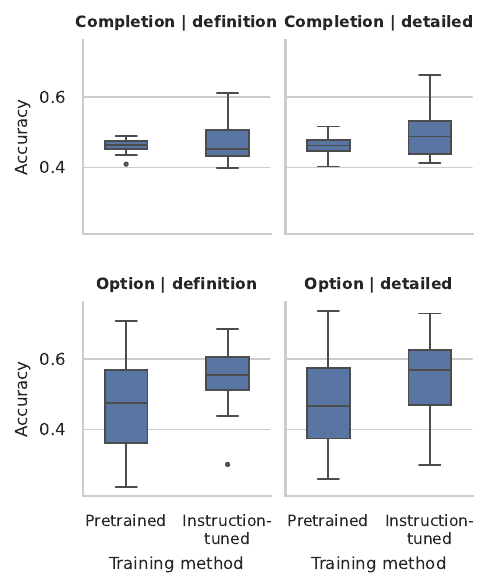}
    \vspace*{-1\baselineskip}
    \caption{Boxplots showing the distribution of zero-shot accuracy for pretrained and instruction-tuned models across evaluation settings, using both definition-based and detailed prompt instructions.}
    \label{fig:instruct-impact}
    \vspace*{-1\baselineskip}
    \end{center}
\end{figure}

\paragraph{Effect of Instruction Tuning.}
Instruction tuning generally improves performance across most model families and evaluation settings. As shown in Figure~\ref{fig:instruct-impact}, this improvement is observed in both completion-based and option-selection evaluations. We further observe that instruction tuning yields larger improvements under the detailed prompt setting than under the definition-based instruction. This suggests that instruction-tuned models follow step-by-step instructions more reliably, leading to larger gains in negation understanding.

However, the effect of instruction tuning is not uniform across all models. In particular, some Qwen3 models exhibit comparable or higher performance in their pretrained versions, indicating that instruction tuning does not always guarantee gains for negation understanding. 

\begin{figure}[htbp]
    \begin{center}
    \includegraphics[width=1\columnwidth]{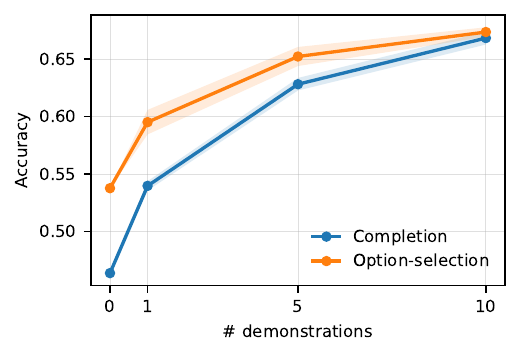}
    \vspace*{-1\baselineskip}
    \caption{Mean accuracy of Llama-3.1-8B-Instruct under the definition instruction across different numbers of in-context demonstrations.}
    \label{fig:fewshot-impact}
    \vspace*{-1\baselineskip}
    \end{center}
\end{figure}

\paragraph{Effect of Few-Shot Learning.}
Overall, increasing the number of in-context demonstrations tends to improve performance across few-shot settings. This indicates that few-shot learning generally benefits sentence-level negation understanding. Figure~\ref{fig:fewshot-impact} illustrates this trend for the Llama-3.1-8B-Instruct model under definition-based instruction. The initial introduction of demonstrations leads to a clear performance improvement, while the incremental benefits decrease as more examples are added.

This trend, however, is not uniform. For some API-based models under the detailed instruction, additional demonstrations do not always yield further gains. One possible reason is that the detailed instructions already provide strong guidance for API-based large models, and adding demonstrations may introduce noise or encourage unnecessary pattern following.

\begin{table*}[htbp]
\centering
\resizebox{\textwidth}{!}{%
\begin{tabular}{@{}ccccr|rrr|rrrr@{}}
\toprule
\multirow[b]{2}{*}{\textbf{\begin{tabular}[c]{@{}c@{}}Evaluation\\ Formats\end{tabular}}} & \multirow[b]{2}{*}{\textbf{\begin{tabular}[c]{@{}c@{}}Instruction\\ Formats\end{tabular}}} & \multirow[b]{2}{*}{\textbf{\begin{tabular}[c]{@{}c@{}}Training\\ Setting\end{tabular}}} & \multirow[b]{2}{*}{\textbf{N Shot}} & \multicolumn{1}{c|}{\multirow{2}{*}{\textbf{\begin{tabular}[c]{@{}c@{}}Error\\ Rate\\ (1-acc)\end{tabular}}}} & \multicolumn{3}{c|}{\textbf{Incorrect Choice Distribution}} & \multicolumn{4}{c}{\textbf{Local Negation Confusion Rate}} \\ \cmidrule(l){6-12} 
 &  &  &  & \multicolumn{1}{c|}{} & \multicolumn{1}{c}{\textbf{\begin{tabular}[c]{@{}c@{}}Local\\ Negation\\ (\%)\end{tabular}}} & \multicolumn{1}{c}{\textbf{\begin{tabular}[c]{@{}c@{}}Contra-\\ diction\\ (\%)\end{tabular}}} & \multicolumn{1}{c|}{\textbf{\begin{tabular}[c]{@{}c@{}}Para-\\ phrase\\ (\%)\end{tabular}}} & \multicolumn{1}{c}{\textbf{\begin{tabular}[c]{@{}c@{}}Relative\\ Clause\\ (\%)\end{tabular}}} & \multicolumn{1}{c}{\textbf{\begin{tabular}[c]{@{}c@{}}Participle\\ Clause\\ (\%)\end{tabular}}} & \multicolumn{1}{c}{\textbf{\begin{tabular}[c]{@{}c@{}}Compound\\ Sentence\\ (\%)\end{tabular}}} & \multicolumn{1}{c}{\textbf{\begin{tabular}[c]{@{}c@{}}Adverbial\\ Clause\\ (\%)\end{tabular}}} \\ \midrule
\multirow{6}{*}{\textbf{\begin{tabular}[c]{@{}c@{}}completion-\\ based\end{tabular}}} & \multirow{3}{*}{\textbf{definition}} & \multirow{2}{*}{\textbf{baseline}} & \textbf{zero-shot} & 0.536 & 73.82 & 23.96 & 2.22 & 25.00 & 31.82 & 58.50 & 48.71 \\
 &  &  & \textbf{5-shot} & 0.367 & 85.31 & 14.25 & 0.43 & 18.91 & 22.40 & 40.14 & 48.06 \\
 &  & \textbf{after SFT} & \textbf{zero-shot} & 0.229 & 86.51 & 12.46 & 1.04 & 10.90 & 14.29 & 23.13 & 33.55 \\ \cmidrule(l){2-12} 
 & \multirow{3}{*}{\textbf{detailed}} & \multirow{2}{*}{\textbf{baseline}} & \textbf{zero-shot} & 0.468 & 71.86 & 26.44 & 1.69 & 23.08 & 29.87 & 47.62 & 38.71 \\
 &  &  & \textbf{5-shot} & 0.347 & 84.70 & 14.84 & 0.46 & 19.55 & 23.05 & 37.76 & 41.29 \\
 &  & \textbf{after SFT} & \textbf{zero-shot} & 0.211 & 87.97 & 10.90 & 1.13 & 8.97 & 13.31 & 21.09 & 33.23 \\ \midrule
\multirow{6}{*}{\textbf{\begin{tabular}[c]{@{}c@{}}option-\\ selection\end{tabular}}} & \multirow{3}{*}{\textbf{definition}} & \multirow{2}{*}{\textbf{baseline}} & \textbf{zero-shot} & 0.462 & 70.67 & 18.52 & 10.81 & 27.56 & 24.68 & 53.40 & 30.00 \\
 &  &  & \textbf{5-shot} & 0.332 & 81.62 & 11.22 & 7.16 & 25.64 & 18.83 & 40.48 & 27.42 \\
 &  & \textbf{after SFT} & \textbf{zero-shot} & 0.356 & 71.94 & 15.37 & 12.69 & 25.00 & 19.81 & 44.56 & 17.10 \\ \cmidrule(l){2-12} 
 & \multirow{3}{*}{\textbf{detailed}} & \multirow{2}{*}{\textbf{baseline}} & \textbf{zero-shot} & 0.486 & 71.45 & 14.85 & 13.70 & 33.01 & 25.97 & 56.12 & 29.03 \\
 &  &  & \textbf{5-shot} & 0.353 & 77.53 & 11.24 & 11.24 & 27.56 & 19.81 & 39.12 & 26.77 \\
 &  & \textbf{after SFT} & \textbf{zero-shot} & 0.270 & 65.69 & 21.70 & 12.61 & 14.74 & 14.61 & 29.25 & 15.16 \\ \bottomrule
\end{tabular}%
}
\caption{Error distribution and confusion analysis of Llama-3.1-8B-Instruct model across various evaluation settings.}
\vspace{-0.5\baselineskip}
\label{tab:neg_analysis}
\end{table*}

\paragraph{Effect of Supervised Fine-Tuning.}
Supervised fine-tuning (SFT) generally leads to improved zero-shot performance across most models, indicating that task-specific training further enhances negation understanding. The performance gains from SFT are more pronounced in the completion-based evaluation than in the option-selection setting, suggesting that the effects of fine-tuning may interact with the evaluation format. We observe that SFT does not degrade models' general language abilities, as confirmed by evaluations on general benchmarks (Appendix~\ref{appendix:general_sft}).

\paragraph{}
Detailed results for all models and training settings are provided in
Appendix~\ref{appendix:result_bench}. In addition, we conducted a human evaluation, showing that although humans perform well on average, distinguishing standard negation from closely related alternatives is not uniformly easy across participants. Detailed results are provided in Appendix~\ref{appendix:human_eval}.

\subsection{Error Analysis}
\label{sec:5_3_analysis}

We analyze model errors to evaluate the ability of our models to differentiate standard negation from similar semantic variants. Each type of local negation in our dataset is explicitly labeled based on its sentence structure, as defined in Table~\ref{tab:localtype}.

We measure the \textit{confusion rate}, defined as the proportion of examples within each subtype where the model incorrectly selects the local negation option instead of the correct standard negation. 
For example, if 320 items are labeled as participle clause negation and the model incorrectly chooses the local negation option instead of the correct standard negation option in 32 of these cases, the confusion rate for participle clause negation would be 10\%. Complete analysis results are provided in Appendix~\ref{appendix:result_analysis}.

We focus on the results of Llama-3.1-8B-Instruct model as shown in Table~\ref{tab:neg_analysis}. In the completion-based setting, error rates generally decrease from zero-shot to 5-shot and continue to improve after SFT, with most errors concentrated in local negation options. Within local negation, compound sentences exhibit the highest confusion but also show the largest relative improvement after SFT.

In the option-selection setting, errors are likewise dominated by local negation, while the model shows a higher tendency to confuse paraphrase options compared to the completion-based setting. Comparatively, the option-selection setting results in lower confusion rates for adverbial clause negation. In addition, performance improvements in the option-selection setting do not consistently follow the expected progression across training configurations. In some cases, SFT yields higher error rates than the few-shot baselines, and this pattern is observed in some other models as well. 

Overall, these patterns highlight how different evaluation settings and model configurations lead to distinct types of errors, and how the addition of more examples or SFT affects error distribution.

\section{Conclusion}
\label{6_conclusion}
In this work, we introduce \bench, a benchmark designed to evaluate LLMs' sentence-level understanding of negation, going beyond surface cue detection. By distinguishing between standard negation, local negation, contradiction, and paraphrase, \bench offers a comprehensive assessment of semantic comprehension. Our experiments demonstrate that while supervised fine-tuning and in-context learning can help reduce specific errors, these approaches still struggle to differentiate standard negation from closely related semantic variants. \bench serves as a valuable diagnostic tool for analyzing the limitations of models' understanding of negation and stands as a robust benchmark for future research. Its design enables evaluation across diverse model families and settings, making it broadly applicable for studying semantic reasoning in LLMs.
\section*{Limitations}
\bench is exclusively constructed in English, despite negation being a universal linguistic phenomenon demonstrated in diverse forms across languages. The syntactic and semantic expressions of negation may vary in other languages, meaning that our current findings may not generalize to multilingual or cross-lingual settings. In future work, we aim to extend the research to a broader range of languages to enable cross-linguistic evaluation of negation understanding in language models.

Although we built the dataset using two distinct sources (HoVer and Wikipedia summaries), both are derived from encyclopedic, formal domains, which may not fully represent the variety of sentence structures and informal language found in real-world use cases. Moreover, while all examples were systematically generated and reviewed, some bias may persist due to subjective decisions in the human review process. We attempt to mitigate this through cross-checking by an independent group of authors, but some residual bias may remain.

\bench primarily focuses on standard (sentence-level) negation and its distinction from local negation, contradiction, and paraphrase. Other important negation phenomena, such as double negation and negative polarity items (NPIs), are not directly addressed in this benchmark. Our current focus is on establishing a strong foundation for evaluating models’ understanding of standard negation. However, we aim to expand the evaluation to a broader range of negation phenomena in future work.

\section*{Ethical Considerations} 
This work does not involve the use of crowd-sourcing methods. Instead, all data included in the \bench benchmark has been carefully reviewed by the authors to ensure quality, relevance, and adherence to ethical standards. The datasets and tools used for training and evaluation are publicly available and used in compliance with their respective licenses.

When leveraging OpenAI’s text generation models, we take additional care to avoid generating or including any content that is harmful, biased, or violates privacy. All generated examples are manually reviewed to meet ethical and safety standards. We ensure no personally identifiable information or offensive content is present in the final dataset. 

The \bench dataset is released under the CC BY-NC-SA 4.0 license, ensuring transparency, reproducibility, and accessibility for future research. We believe our work contributes positively to developing trustworthy and interpretable language models.
\section*{Acknowledgments}
We thank the anonymous reviewers and the meta-reviewer for their valuable feedback on this paper. We also sincerely thank Sungeun Hahm, Suyoung Park, Jongmin Kim, Yelim Ahn, Hyunji M. Park, Seorin Kim, and Jisoo Kim for their valuable contributions to the initial design of the negation dataset.

This work was partially supported by the National Research Foundation of Korea (NRF) under Grant No. RS-2023-00222663 (Center for Optimizing Hyperscale AI Models and Platforms), and by the Institute for Information and Communications Technology Promotion (IITP) under Grant No. 2018-0-00581 (CUDA Programming Environment for FPGA Clusters) and No. RS-2025-02304554 (Efficient and Scalable Framework for AI Heterogeneous Cluster Systems), all funded by the Ministry of Science and ICT (MSIT) of Korea. It was also partially supported by the Korea Health Industry Development Institute (KHIDI) under Grant No. RS-2025-25454559 (Frailty Risk Assessment and Intervention Leveraging Multimodal Intelligence for Networked Deployment in Community Care), funded by the Ministry of Health and Welfare (MOHW) of Korea. Additional support was provided by the BK21 Plus Program for Innovative Data Science Talent Education (Department of Data Science, Seoul National University, No. 5199990914569) and the BK21 FOUR Program for Intelligent Computing (Department of Computer Science and Engineering, Seoul National University, No. 4199990214639), both funded by the Ministry of Education (MOE) of Korea. This work was also partially supported by the Artificial Intelligence Industrial Convergence Cluster Development Project, funded by the MSIT and Gwangju Metropolitan City. Research facilities were provided by the Institute of Computer Technology (ICT) at Seoul National University.

\bibliography{custom}

\appendix

\section{Typology of Contradiction}
\label{appendix:contradiction}

\begin{table*}[htbp]
\begin{center}
\resizebox{0.9\textwidth}{!}{%
\begin{tabular}{@{}cll@{}}
\toprule
\textbf{Contradiction Type} &
  \multicolumn{1}{c}{\textbf{Definition}} &
  \multicolumn{1}{c}{\textbf{Example}} \\ \midrule \midrule
\textbf{Antonym} &
  \begin{tabular}[c]{@{}l@{}}Contradiction caused by opposing \\ meanings of aligned words.\end{tabular} &
  \begin{tabular}[c]{@{}l@{}}The policy was a \textbf{success}.\\ $\rightarrow$ The policy was a \textbf{failure}.\end{tabular} \\ \midrule
\textbf{Negation} &
  \begin{tabular}[c]{@{}l@{}}One sentence explicitly negates \\ a statement in the other.\end{tabular} &
  \begin{tabular}[c]{@{}l@{}}She \textbf{attended} the meeting.\\ $\rightarrow$ She \textbf{did not attend }the meeting.\end{tabular} \\ \midrule
\textbf{Numeric} &
  \begin{tabular}[c]{@{}l@{}}Inconsistent numbers, dates, or \\ quantities in related statements.\end{tabular} &
  \begin{tabular}[c]{@{}l@{}}Totally, \textbf{ten} people were injured.\\ $\rightarrow$ Totally, \textbf{five} people were injured.\end{tabular} \\ \midrule
\textbf{Factive/Modal} &
  \begin{tabular}[c]{@{}l@{}}Conflict in implied facts or modal \\ possibilities due to verbs or auxiliaries.\end{tabular} &
  \begin{tabular}[c]{@{}l@{}}He \textbf{managed to} enter the building.\\ $\rightarrow$ He \textbf{did not enter} the building.\end{tabular} \\ \midrule
\textbf{Structure} &
  \begin{tabular}[c]{@{}l@{}}Syntactic rearrangement or argument \\ swapping  causes contradiction.\end{tabular} &
  \begin{tabular}[c]{@{}l@{}}\textbf{Alice} hired \textbf{Bob}.\\ $\rightarrow$ \textbf{Bob} hired \textbf{Alice}.\end{tabular} \\ \midrule
\textbf{Lexical} &
  \begin{tabular}[c]{@{}l@{}}Contradiction through incompatible \\ verbs or phrases, not strictly antonyms.\end{tabular} &
  \begin{tabular}[c]{@{}l@{}}The manager \textbf{praised} her performance.\\ $\rightarrow$ The manager \textbf{expressed} \\
  \hspace{2em} \textbf{disappointment in} her performance.\end{tabular} \\ \midrule
\begin{tabular}[c]{@{}c@{}}
\textbf{World}\\ 
\textbf{Knowledge} 
\end{tabular} &
  \begin{tabular}[c]{@{}l@{}}Contradiction relies on common-sense \\ or background knowledge.\end{tabular} &
  \begin{tabular}[c]{@{}l@{}}The Eiffel Tower is in \textbf{Paris}.\\ $\rightarrow$ The Eiffel Tower is in \textbf{Berlin}.\end{tabular} \\ \bottomrule
\end{tabular}
}
\caption{Contradiction types from~\citet{de2008finding}. Contradiction covers a broader scope than negation.}
\label{tab:contratype}
\end{center}
\end{table*}

Contradictions in natural language can arise in diverse ways that go beyond simple negation. Following the typology of~\citet{de2008finding}, contradictions can be grouped into seven categories: antonymy, explicit negation, numeric mismatch, factive/modal inconsistencies, structural reversals, lexical incompatibilities, and conflicts based on world knowledge. These categories reflect the fact that contradiction covers a broader semantic scope than negation alone. Table~\ref{tab:contratype} summarizes these types with definitions and examples.

\section{Copular Verbs}
Copular verbs, also known as linking verbs, are verbs that connect the subject of a sentence to a subject complement, which can be a noun, adjective, or other expression that describes or identifies the subject. Unlike action verbs, copular verbs do not express actions but rather states or conditions. The most common copular verb in English is "to be" in its various forms (am, is, are, was, were). Other examples include "seem," "appear," "become," "feel," "look," "sound," "taste," and "smell" when used to describe the subject's state~\citep{hengeveld1986copular}.

As discussed in Section~\ref{sec:3_3_standard_negation}, standard negation in this work targets the main predicate of a clause. For sentences with copular verbs, this means that the entire verb phrase, including the copular verb and its complement, is subject to negation. For example, in the sentence "She is a doctor," the main predicate is "is a doctor." Negating this sentence results in "She is not a doctor," where the negation applies to the entire predicate, not just the verb "is."

Negation of a verb phrase including a copular verb can be realized either syntactically (e.g., "is \textbf{not} an expert") or by replacing the complement with its complementary antonym (e.g., "is a \textbf{non-expert}"), both of which result in the reversal of the main predicate’s truth value. Although such constructions may superficially appear to be non-verbal negation, especially when the complement is a noun or adjective, they are, in fact, instances of verbal negation, since the negation applies to the predicate as a whole.

\section{Negation of Implications}
\label{appendix:implication}

Negating implications presents challenges, as natural language intuitions often diverge from the rules of formal logic. Let's say there is a conditional statement, "If I study hard, I will pass the bar exam." Formally, let $P$ denote "I study hard" and $Q$ denote "I will pass the bar exam." In classical logic, the conditional "if $P$, $Q$" can be false only when $P$ is true and $Q$ is false. This implies that the negation of the conditional is "$P \;\text{and}\; \text{Neg}(Q)$" ("I study hard and I won't pass the bar exam,) while the conditional itself is equivalent to "$\text{Neg}(P) \;\text{or}\; Q$" ("I don't study hard or I will pass the exam)~\citep{nguyen2023negation}. 

Psychological studies confirm that people often accept both "if $P$, Neg($Q$)" ("If I study hard, I won't pass the exam.") and "if Neg($P$), $Q$" ("If I don't study hard, I will pass the exam."). However, the former can be interpreted as "$\text{Neg}(P) \;\text{or}\; \text{Neg}(Q)$", and the latter "$P \;\text{or}\; Q$", both of which are not equivalent to the original statement's negation, "$P \;\text{and}\; \text{Neg}(Q)$". "if Neg($P$), Neg($Q$)" ("If I don't study hard, I won't pass the exam.") is not the correct negation as well, as it is equivalent to "$P \;\text{or}\; \text{Neg}(Q)$"~\citep{espino2012not}.

While humans often struggle to distinguish the correct negation of a conditional from invalid alternatives, the logical form is unambiguous. We therefore include conditional statements in our benchmark to test whether language models, like humans, are prone to intuitive but invalid interpretations, or whether they can correctly apply truth-functional reasoning. 

Note that implications in natural language are not limited to the explicit "if $P$, $Q$" form, 
but may also appear with connectives such as \textit{when}, \textit{as long as}, or \textit{unless}, 
which functionally convey conditional meaning and are treated under the same negation principle.

\section{Compound Sentences and Coordinating Conjunction}

A compound sentence consists of two or more independent clauses joined by a coordinating conjunction. Each clause can stand alone, but they are combined to express related ideas~\citep{gleitman1965coordinating}.

Coordinating conjunctions connect elements of equal grammatical rank. The seven common ones in English are: \textit{for, and, nor, but, or, yet, so} (often remembered as FANBOYS). Among these, \textit{and}, \textit{or}, and \textit{but} are indisputably used to coordinate clauses. The others can be ambiguous or function in non-coordinating roles(e.g., indicating cause or result rather than logical structure). These are the examples using \textit{and}, \textit{or}, and \textit{but} to connect sentences equally.

\begin{itemize}[nosep]
    \item "She studied hard, \textbf{and} she passed the exam."
    \item "I wanted to go, \textbf{but} it was raining."
    \item "You can call me, \textbf{or} you can send an email."
\end{itemize}

We consider only the coordinating conjunctions \textit{and}, \textit{or}, and \textit{but} as indicators of compound sentences, in which two or more independent clauses are equally connected.
Although \textit{but} introduces a contrast semantically, in terms of logical structure, it functions as a conjunction equivalent to \textit{and}; therefore, its negation follows the same principle.

\section{Local Negation Constructions Excluded}

In constructing the \bench dataset, we consider various types of local (i.e., subclausal) negation, where negation applies to a phrase or constituent rather than the main predicate. However, several constructions are excluded due to their semantic ambiguity, syntactic irregularity, or misalignment with the benchmark’s focus on verbal negation.

\paragraph{Infinitive Phrase Negation.}  
Infinitive phrases (e.g., "to go") can be negated with "not" (e.g., "not to go" or "to not go"). Unlike the clause-level structures that define our local negation category,  infinitive phrases are not full clauses but simply part of a verb phrase, making them less compatible with our definition. Moreover, although grammatically correct, this construction is relatively rare and sounds awkward depending on the context.  

\begin{itemize}[nosep]
    \item \textbf{Original}: George wants to go to the park.  
    \item \textbf{Negated (infinitive)}: George wants not to go / George wants to not go to the park.  
\end{itemize}

For these reasons, we exclude infinitive phrase negation from the benchmark.

\paragraph{Appositive Clause Negation.}  
Appositive clauses are noun phrases that provide descriptive clarification. Attempting to negate an appositive typically involves lexical replacement rather than syntactic negation.  
\begin{itemize}[nosep]
    \item \textbf{Original:} My brother, a talented musician, plays the guitar.  
    \item \textbf{Negated (appositive):} My brother, not a talented musician, plays the guitar.
\end{itemize}
Such changes alter descriptive content rather than reversing the meaning of the predicate, and often fall into the domain of contradiction. Accordingly, they are excluded from the dataset.

\paragraph{Prepositional Phrase Negation.}  
Negating a prepositional phrase often involves replacing the preposition with its antonym (e.g., "with" → "without", "in" → "outside"), which results in a sentence that differs in content, rather than reversing the meaning of the predicate.  
\begin{itemize}[nosep]
    \item \textbf{Original}: She went to the park with her bird.  
    \item \textbf{Negated (preposition)}: She went to the park without her bird.
\end{itemize}
Since such modifications do not negate the verb but instead change the nature of an adjunct or argument, they fall outside the scope of standard negation or local negation in this work and are excluded.

In all of the above cases, the negation does not target the whole verb phrase but rather peripheral elements within the sentence. As the \bench is designed to evaluate verbal negation, these local or phrase-level forms of negation were intentionally left out.

\section{Double Negation}
Double negation refers to the use of two forms of grammatical negation within a single sentence. In standard English, only one negative form should be present in a subject-predicate construction; the presence of two negatives is generally considered non-standard and often results in an unintended meaning. For example, while "He's going nowhere" is correct, "He's not going nowhere" is ungrammatical. Another example is "I won't bake no cake," which combines verb negation ("won't") with object negation ("no cake"), resulting in a grammatically incorrect construction~\citep{deprez2015double}.

In English, certain double negation constructions convey affirmative meanings rather than intensifying negation, effectively paraphrasing the original positive statement (e.g., $\neg\neg p \approx p$)~\citep{van1996litotes}. This rhetorical device, known as litotes, often manifests in expressions such as "not bad," implying "good," or "not unhappy," implying "happy." Leveraging this phenomenon, we have generated paraphrase candidates for our dataset using such double negation patterns. For example,

\begin{itemize}[nosep]
  \item \textbf{Sentence:} His characteristic style fuses samba, funk, rock and bossa nova with lyrics that blend humor and satire with often esoteric subject matter.\\
        \textbf{Double Negation:} His characteristic style \textbf{does not fail to fuse} samba, funk, rock, and bossa nova with lyrics that blend humor and satire with often esoteric topics.

  \item \textbf{Sentence:} It covers a broad range of fields, including the humanities, social sciences, exact sciences, applied sciences, and life sciences.\\
        \textbf{Double Negation:} It\textbf{ does not exclude} a broad range of fields, including the humanities, social sciences, exact sciences, applied sciences, and life sciences.

  \item \textbf{Sentence:} Sanders was honoured to meet with many world dignitaries and representatives of UNESCO member nations, and delighted when delegates from UNESCO, visited Toowoomba in 2018 in return.\\
        \textbf{Double Negation:} Sanders\textbf{ was not unhappy }to meet with many world dignitaries and representatives of UNESCO member nations, and not displeased when delegates from UNESCO visited Toowoomba in 2018 in return.
\end{itemize}

However, upon closer examination, these paraphrase candidates do not always preserve the exact meaning of the original sentence. The antonyms used (e.g., "exclude" for "cover," "unhappy" for "honoured") are not always true complementary antonyms, which does not effectively negate the meaning. Moreover, the litotes construction ("does not fail to fuse") tends to add an emphatic nuance, rather than being a perfect semantic equivalent. Therefore, the boundary between paraphrasing and double negation is ambiguous, and their relationship requires more careful analysis. Given these issues, and because our primary focus is on standard negation, we ultimately decide to exclude double negation constructions as paraphrase candidates from our dataset.

\section{HoVer Dataset}
\label{appendix:hover}

\begin{table*}[htbp]
\begin{center}
\resizebox{\textwidth}{!}{%
\begin{tabular}{lll}
\hline
\textbf{Column} &
  \textbf{Detail} &
  \textbf{Example} \\ \hline \hline
id &
  Unique claim identifier &
  0 \\ \hline
uid &
  \begin{tabular}[c]{@{}l@{}}User/annotator identifier\end{tabular} &
  330ca632-e83f-4011-b11b-0d0158145036 \\ \hline 
claim &
  \begin{tabular}[c]{@{}l@{}}The statement to be verified, \\ often requiring multi-article evidence \end{tabular} &
  \begin{tabular}[c]{@{}l@{}}Skagen Painter Peder Severin Krøyer favored \\ naturalism along with Theodor Esbern Philipsen \\ and the artist Ossian Elgström studied \\ with in the early 1900s.\end{tabular} \\ \hline
supporting\_facts &
  \begin{tabular}[c]{@{}l@{}} List of Wikipedia article titles and sentence indices \\ providing evidence \end{tabular} &
  \begin{tabular}[c]{@{}l@{}}{[} \{ "key": "Kristian Zahrtmann", "value": 0 \}, \\ \{ "key": "Kristian Zahrtmann", "value": 1 \}, \\ \{ "key": "Peder Severin Krøyer", "value": 1 \}, \\ \{ "key": "Ossian Elgström", "value": 2 \} {]}\end{tabular} \\ \hline
label &
  \begin{tabular}[c]{@{}l@{}}Whether the claim is supported \end{tabular} &
  1: SUPPORTED or 0: NOT\_SUPPORTED\\ \hline
num\_hops &
  \begin{tabular}[c]{@{}l@{}}Number of articles required for verification \end{tabular} &
  2$\sim$4 \\ \hline
hpqa\_id &
  \begin{tabular}[c]{@{}l@{}}Reference to the original HotpotQA pair \end{tabular} &
  5ab7a86d5542995dae37e986 \\ \hline
\end{tabular}%
}
\caption{Details of HoVer dataset structure with examples.}
\label{tab:hover}
\end{center}
\end{table*}
The HoVer (\textbf{Ho}ppy \textbf{Ver}ification) dataset is developed for the tasks of multi-hop evidence retrieval and factual claim verification. In HoVer, each claim requires supporting evidence that spans multiple English Wikipedia articles to determine whether the claim is substantiated or not. The dataset is distributed under a CC BY-SA 4.0 License, and it can be accessed via its official homepage\footnote{\url{https://hover-nlp.github.io/}}. Table~\ref{tab:hover} offers an overview of the dataset’s structure. The data is split into training, validation, and test sets, containing 18,171, 4,000, and 4,000 examples respectively.

HoVer is constructed on top of the HotpotQA dataset, which is designed to evaluate multi-hop reasoning in question answering. HotpotQA itself is a large-scale collection of Wikipedia-based QA pairs created to address the limitations of prior QA datasets, which often fail to require complex reasoning or explanatory answers~\citep{yang2018hotpotqa}. The construction of HoVer involves rewriting HotpotQA question-answer pairs into claim statements, which are then validated and labeled by annotators. Claims are extended to require multi-hop evidence from up to four Wikipedia articles and are systematically modified to increase complexity. Final labels are assigned as \texttt{\small SUPPORTED} or \texttt{\small NOT-SUPPORTED}~\citep{jiang2020hover}.

\section{Wikipedia Summary Dataset}
\label{appendix:wiki}

\begin{table*}[htbp]
\begin{center}
\resizebox{\textwidth}{!}{%
\begin{tabular}{lll}
\hline
\textbf{Column}         & \textbf{Detail}                                              & \textbf{Example}     \\ \hline \hline
title                   & Article title from Wikipedia.                                     & Alain Connes         \\ \hline
description             &  \begin{tabular}[c]{@{}l@{}} A brief description or category for the article \\ (when available).\end{tabular} & French mathematician \\ \hline
summary &
  \begin{tabular}[c]{@{}l@{}}The extracted summary or introduction section \\ of the article, typically more concise than the full text.\end{tabular} &
  \begin{tabular}[c]{@{}l@{}}Alain Connes (; born 1 April 1947) \\is a French mathematician...\end{tabular} \\ \hline
full\_text &
  \begin{tabular}[c]{@{}l@{}}The complete article text (when included), \\ encompassing the full body of the Wikipedia page.\end{tabular} &
  \begin{tabular}[c]{@{}l@{}}Alain Connes (; born 1 April 1947) \\is a French mathematician...\end{tabular} \\ \hline
\_\_index\_level\_0\_\_ & Index number for each entry in the dataset.                       & 3 \\ \hline                  
\end{tabular}%
}
\caption{Details of Wikipedia Summary dataset structure with examples.}
\label{tab:wiki_summary}
\end{center}
\end{table*}

The Wikipedia Summary Dataset contains the titles and introductory summaries of English Wikipedia articles, extracted in September 2017. A summary or introduction in this context refers to the content from the article title up to the content outline (i.e., before the first section heading). The dataset was originally released via GitHub\footnote{\url{https://github.com/tscheepers/Wikipedia-Summary-Dataset}}, but is now accessible through the Hugging Face Hub\footnote{\url{https://huggingface.co/datasets/jordiclive/wikipedia-summary-dataset}}. The dataset license is not explicitly mentioned, but as the original Wikipedia data is distributed under the CC BY-SA 4.0, it is assumed that the dataset would be distributed under the same license. For licensing details, refer to the Wikimedia Terms of Use \footnote{\url{https://foundation.wikimedia.org/wiki/Policy:Terms_of_Use}}. Table~\ref{tab:wiki_summary} offers an overview of the dataset’s structure. The dataset comprises approximately 430,000 articles, only providing the training set~\citep{scheepers2017compositionality}.

\section{Human Review Protocol}
\label{appendix:review_protocol}

To ensure high-quality data construction, we implement a rigorous quality control protocol that combines generation, independent review, and iterative consensus building. The process involves the following key steps:

\begin{itemize}[nosep]
    \item \textbf{Task allocation and independence.} Authors are assigned distinct portions of the dataset, but no author is permitted to review the data they have generated. This ensures that each instance is subject to at least one independent review.
    \item \textbf{Sequential authoring across choices.} For the multiple-choice dataset, construction proceeds in four stages: standard negation, local negation, contradiction, and paraphrase. At each stage, different authors are responsible for creating the new option, while reviewers who have not authored that option perform the verification.
    \item \textbf{Cross-checking and layered review.} Each newly created option is reviewed by at least one other author, and reviewers also revisit earlier options in the same instance. For example, when reviewing the paraphrased sentence, the reviewer also checks that standard negation, local negation, and contradiction sentences are correct. As a result, every instance undergoes multiple rounds of verification across stages, such that all authors ultimately examine data they have not created themselves.
    \item \textbf{Guideline refinement and retroactive correction.} Generation and reviewing guidelines are continuously updated based on discussion of ambiguous or problematic cases. Whenever the guidelines changes, all previously created data are revisited to ensure compliance, promoting consistency across the dataset.
    \item \textbf{Consensus and adjudication.} Disagreements are discussed in weekly meetings and, if necessary, adjudicated by a lead reviewer, ensuring that no instance remains unresolved.
\end{itemize}

Overall, this iterative and layered procedure ensures that every instance in the multiple-choice dataset is independently reviewed across multiple stages, leading to stable guidelines and a consistent dataset.

\paragraph{Edit-based reliability and quality control statistics.}
Because our review process is revision-based rather than label-based, standard inter-annotator agreement metrics (e.g., Cohen’s $\kappa$) are not directly applicable. Instead of assigning categorical labels, reviewers have directly edited or deleted problematic instances to enforce the dataset guidelines.

We therefore report edit-based quality control statistics as a proxy for annotation reliability, including (i) the number of instances escalated to adjudication meetings and (ii) the number of instances that required edits after cross-checking. All cases requiring edits have been discussed in meetings and resolved by consensus, ensuring consistency across the dataset.

\begin{table}[h!]
\centering
\resizebox{\columnwidth}{!}{%
\begin{tabular}{@{}ccrrrr@{}}
\toprule
\multicolumn{2}{c}{\textbf{Subset}} & \multicolumn{1}{c}{\textbf{\# items}} & \multicolumn{1}{c}{\textbf{\begin{tabular}[c]{@{}c@{}}\# Brought\\ to meeting\end{tabular}}} & \multicolumn{1}{c}{\textbf{\# Edited}} & \multicolumn{1}{c}{\textbf{\begin{tabular}[c]{@{}c@{}}Edit\\ rate\end{tabular}}} \\ \midrule
\textbf{\begin{tabular}[c]{@{}c@{}}Sentence-\\ Negation\end{tabular}} & \textbf{\begin{tabular}[c]{@{}c@{}}Standard\\ Negation\end{tabular}} & 3,992 & 469 & 72 & 1.80\% \\ \midrule
\multirow{3}{*}{\textbf{\begin{tabular}[c]{@{}c@{}}Multiple \\ Choice\end{tabular}}} & \textbf{\begin{tabular}[c]{@{}c@{}}Standard/\\ Local Negation\end{tabular}} & 1,102 & 126 & 34 & 3.09\% \\ \cmidrule(l){2-6} 
 & \textbf{\begin{tabular}[c]{@{}c@{}}Contradiction/\\ Paraphrase\end{tabular}} & 1,102 & 409 & 117 & 10.62\% \\ \cmidrule(l){2-6} 
 & \textbf{\begin{tabular}[c]{@{}c@{}}Implication\\ (add-on)\end{tabular}} & 201 & 49 & 16 & 7.96\% \\ \bottomrule
\end{tabular}%
}
\caption{Edits include rewriting or deleting instances that violate guidelines.}
\end{table}

Overall, the relatively low edit rates after cross-checking indicate that most instances satisfied the guidelines upon independent review, while meeting-based adjudication played a critical role in resolving ambiguous or complex cases.

\section{Detailed Principles and Examples of the \bench}
\label{appendix:guidelines}
While the main text already defines the core notions of standard and local negation (Section~\ref{sec:3_negation}) and explains how they are applied throughout dataset construction (Section~\ref{sec:4_benchmark}), here we provide more detailed illustrations.

\paragraph{Paraphrasing before Negation.} Before negating, the main verb or other components may be paraphrased with synonyms, provided that the sentence's tense, structure, and meaning remain strictly equivalent before applying standard negation. Authors refer to the Merriam-Webster Thesaurus~\footnote{\url{https://www.merriam-webster.com/}}.
For example,
\begin{itemize}
    \item \textbf{Original Sentence}: Toumour is a village and rural commune in Niger \textbf{located near} the Niger–Nigeria \textbf{border}. 
    \begin{itemize}
        \item \textbf{Paraphrased Sentence}: Toumour is a village and rural commune in Niger\textbf{ that is found close to }the Niger–Nigeria \textbf{boundary}. \\
        $\rightarrow$  \textbf{Standard Negation after Paraphrase}: Toumour \textbf{isn't} a village and rural commune in Niger that is found close to the Niger–Nigeria boundary.
        \item \textbf{Explanation}: In this example, the participle clause "located near the Niger–Nigeria border" is rephrased as a relative clause "that is found close to the Niger–Nigeria boundary." Since both constructions serve as modifiers and preserve the same semantic role, we treat them as equivalent in meaning for the purpose of standard negation.
    \end{itemize}

    \item \textbf{Original Sentence}: The armed forces \textbf{said} Boko Haram \textbf{attacked} their military post on March 15, 2020, which they responded to by repelling the attack, killing 50 insurgents. 
    \begin{itemize}
        \item \textbf{Paraphrased Sentence}: The armed forces \textbf{stated} that Boko Haram \textbf{assaulted} their military post on March 15, 2020, which they responded to by repelling the attack, killing 50 insurgents. \\
        $\rightarrow$ \textbf{Standard Negation after Paraphrase}: The armed forces \textbf{didn't state} that Boko Haram assaulted their military post on March 15, 2020, which they responded to by repelling the attack, killing 50 insurgents.
        \item \textbf{Explanation}: In this example, the reporting verb “said” is paraphrased as “stated,” and the verb “attacked” is replaced with the synonym “assaulted.” These substitutions preserve the original tense and meaning, allowing standard negation to be applied without altering the semantic content of the sentence.
    \end{itemize}
\end{itemize}

\paragraph{Negation of Simple Sentences.} For simple, declarative sentences, standard negation is achieved by inserting "not" after the auxiliary or main verb, or by replacing the predicate with its complementary antonym. For example, "She is happy." $\rightarrow$ "She is not happy."; "The room is occupied." $\rightarrow$ "The room is unoccupied."
    
\paragraph{Negation in Compound Sentences.} When multiple clauses or propositions are coordinated (e.g., with "and", "or", "but"), standard negation is logically applied, governed by De Morgan's laws. Here, "but" is treated as a coordinating conjunction equivalent to "and" in terms of logical structure, so its negation follows the same principle.
\begin{itemize}
    \item Conjunction "$P$ and/but $Q$": the negation is "$\text{Neg}(P) \;\text{or}\; \text{Neg}(Q)$".
    \item Disjunction "$P$ or $Q$": the negation is "$\text{Neg}(P) \;\text{and}\; \text{Neg}(Q)$".
\end{itemize}
For example, "He passed the test and received an award." is negated as "He did not pass the test or did not receive an award."
    
When application of logical negation produces unnatural language, sentences may be split or slightly rephrased for fluency, provided logical meaning is preserved. For example, 
\begin{itemize}
    \item \textbf{Original:} "He finished the report and submitted the assignment."
    \item \textbf{Standard Negation:} "He did not finish the report or did not submit the assignment."
    \item \textbf{Standard Negation, but Splitted:} "He did not finish the report. Or, he did not submit the assignment."
\end{itemize}
    
\paragraph{Coordinated Elements in the Sentence.} When a sentence contains coordinated elements (such as subjects, objects, or predicates connected by "and" or "or"), standard negation typically follows logical principles derived from De Morgan's Laws. However, whether logical negation applies to each individual component or to the entire predicate as a whole depends on whether the coordination expresses multiple independent propositions or a single collective event.

\begin{itemize}
    \item If the coordination introduces semantically distinct propositions, that is, each conjunct could form a complete sentence on its own, negation must be applied to each proposition individually. For example, "My sister and I studied hard."\\
    This sentence can be interpreted as: "My sister studied hard and I studied hard."\\
    Therefore, the correct standard negation is: "My sister did not study hard, or I did not study hard."
    \item Conversely, if the coordination connects elements that jointly participate in a single action or state (e.g., a shared subject or a collective predicate), then the sentence is treated as a simple clause, and the predicate as a whole is negated. Logical decomposition is not appropriate. For example, "My sister and I share clothes."\\
    This expresses a single collective action involving both participants.\\
    Therefore, the correct standard negation is: "My sister and I do not share clothes."\\
    (NOT: "My sister does not share clothes, or I do not share clothes.")

    \item This distinction is crucial: even if two noun phrases are coordinated, if the sentence semantically decomposes into separate atomic propositions, standard negation must apply to each atomic proposition. Otherwise, it applies to the whole predicate as one unit.

    \item Other examples of semantically collective predicates where logical splitting is not appropriate include: "be the same", "have in common", "do something together", "combine", "unite", etc. These describe inherently joint or relational properties, not independent propositions. For example, "Clarence Brown and Peter Glenville are from the same country." should be negated as "Clarence Brown and Peter Glenville are not from the same country."
\end{itemize}
        
\paragraph{Use of Antonyms.} When replacing predicates with antonyms in standard negation, only complementary antonyms are appropriate, as they provide a clear binary opposition, ensuring logical consistency of negation. Gradable and relational antonyms are unsuitable for standard negation because their antonyms do not represent the logical complement of the original predicate. In other words, replacing a predicate $p$ with its antonym does not produce $\neg p$ in a truth-conditional sense. 

Specifically, unlike complementary antonyms, which form mutually exclusive pairs (i.e., $p \cup \neg p = U$ and $p \cap \neg p = \emptyset$), gradable and relational antonyms do not partition the meaning space cleanly, and thus fail to reverse the truth value reliably.

\begin{itemize}
    \item \textbf{Complementary Antonyms}: Also called binary/contradictory antonyms. These antonyms represent mutually exclusive pairs with no intermediate states. The presence of one implies the absence of the other. Examples include:
    \begin{itemize}
        \item alive / dead
        \item true / false
        \item present / absent
        \item occupied / vacant
    \end{itemize}
    Using complementary antonyms in negation ensures a direct and unambiguous reversal of the original proposition's truth value.

    \item \textbf{Gradable Antonyms}: These antonyms exist on a continuum and allow for varying degrees between the two extremes. Negating one does not necessarily affirm the other. Examples include:
    \begin{itemize}
        \item hot / cold
        \item happy / sad
        \item tall / short
        \item young / old
    \end{itemize}
    Due to their scalar nature, gradable antonyms are inappropriate for standard negation, as they do not provide a definitive binary opposition.

    \item \textbf{Relational Antonyms}: Also known as converse antonyms, these pairs describe a reciprocal relationship where one implies the existence of the other. Examples include:
    \begin{itemize}
        \item parent / child
        \item teacher / student
        \item buy / sell
        \item employer / employee
    \end{itemize}
    Relational antonyms are context-dependent and do not represent direct opposites in a binary sense, making them unsuitable for standard negation purposes.
\end{itemize}

\paragraph{General Principles of Standard Negation.} 
\begin{itemize}
    \item The negated sentence must preserve all elements (subject, tense, objects, adjuncts, etc.) of the original, except for the truth value of the main predicate.
    \item When naturalness and logical negation conflict, logical correctness takes priority, but minimal rephrasing is allowed for fluency.
    \item If the negated clause creates a contradiction with other parts of the sentence, the contradictory clause must be removed. For example, the standard negation of the sentence
    "While the spatial size of the entire universe is unknown, it is possible to measure the size of the observable universe, which is approximately 93 billion light-years in diameter." \\
    will be "While the spatial size of the entire universe is unknown, it isn’t possible to measure the size of the observable universe." \\
    The relative clause must be removed because its content directly contradicts the negated main clause.
\end{itemize}

\paragraph{Common Negation Errors and Corrections.}
\begin{itemize} 
    \item \textbf{Original sentence}: His characteristic style fuses samba, funk, rock \textbf{and} bossa nova with lyrics that blend humor and satire with often esoteric subject matter.
    \begin{itemize}
        \item \textbf{Incorrect negation}: His distinctive style \textbf{doesn't fuse} samba, funk, rock \textbf{or} bossa nova with lyrics that blend humor and satire with often esoteric subject matter. 
        \item \textbf{Correct negation}: His distinctive style\textbf{ doesn't fuse} samba, funk, rock \textbf{and} bossa nova with lyrics that blend humor and satire with often esoteric subject matter.
        \item \textbf{Explanation}: The verb "fuse" implies a combination of all listed elements. "and" must be preserved.
    \end{itemize}

    \item \textbf{Original sentence}: The mascot of Avon Center School is the "Koalaty Kid," \textbf{while} the mascot at Prairieview \textbf{is} an eagle \textbf{and} the mascot at Woodview \textbf{is} an owl.
    \begin{itemize}
        \item \textbf{Incorrect negation}: Avon Center School's mascot is not the "Koalaty Kid," Prairieview's mascot \textbf{is not} an eagle, \textbf{or} Woodview's mascot \textbf{is not} an owl.
        \item \textbf{Correct negation}: Avon Center School's mascot is not the "Koalaty Kid," \textbf{while} the mascot at Prairieview \textbf{is} an eagle \textbf{and} the mascot at Woodview \textbf{is} an owl.
        \item \textbf{Explanation}: Two clauses connected by while are not coordinated propositions (as with \textit{and} or \textit{or}), but instead express contrastive information. Therefore, applying logical negation across both clauses is incorrect. Negation should apply only to the main clause (here, the first statement), while the contrasting clause remains affirmative.
    \end{itemize}
\end{itemize}

\section{Code for Data Construction}
\label{appendix:datacode}

\subsection{Sentence-Negation Pair Dataset}
\label{appendix:traindatacode}

To construct the sentence-negation pair dataset, we begin by randomly sampling sentences labeled as "supported facts" from the HoVer dataset. Since the original data often contains grammatical errors, we utilize OpenAI's API~\citep{OpenAI} to automatically correct these issues. In cases where the selected text consists of multiple sentences, we merge or split them as needed to ensure that each example is a single sentence, aligning with our sentence-level task objective.

We select different model versions depending on the complexity of each task. For sentence merging, which demands nuanced contextual understanding and complex syntax, we use \texttt{\small GPT-4}. For grammar correction, where edits are more straightforward, \texttt{\small GPT-3.5} is sufficient.

\begin{lstlisting}[language=Python, caption={Fixing Grammar with OpenAI API.}, label={lst:grammar_fix}]
def grammar_fix(claim):
    messages = [{"role": "system", "content": "Fix grammatical errors."},
    {"role": "user", "content": f"If there are errors, please fix the sentence: {claim} \n If there aren't, return the original sentence. Provide only the resulting sentence without any additional explanation or introduction."}]
    response = client.chat.completions.create(model="gpt-3.5-turbo", messages=messages)
    fixed_text = response.choices[0].message.content.strip()
    return fixed_text
\end{lstlisting}

\begin{lstlisting}[language=Python, caption={Merging Sentences with OpenAI API.}, label={lst:merging_sentences}]
def merge_sentences_with_gpt(claim):
    messages = [{"role": "system", "content": "Merge sentences into a single one."},
    {"role": "user", "content": f"Merge these sentences: {claim} \n Provide only the resulting sentence without any additional explanation or introduction."}]
    response = client.chat.completions.create(model="gpt-4-turbo-preview", messages=messages)
    merged_text = response.choices[0].message.content.strip()
    return merged_text
\end{lstlisting}

\subsection{Multiple Choice Dataset}
\label{appendix:testdatacode}

To construct the multiple-choice dataset, we first segment the "summary" column of the Wikipedia Summary dataset, which often contains multiple sentences in a single entry, into individual sentences. To focus on the challenges of negation in complex sentences, we filter out sentences that are too short. This process is done with Python code. 

Since conditional sentences (e.g., "If $P$, $Q$") are rarely present in the Wikipedia summary dataset, we adopt a two-step approach: (1) prompting the model to generate conditional variants from given sentences (using OpenAI API, \texttt{\small GPT-4o-mini}), and (2) manually filtering or lightly editing the results to obtain valid conditionals. 

Subsequently, we automatically generate contradictions and paraphrases for each sentence via the OpenAI API (\texttt{\small GPT-4o}) as well, followed by human review. 
The following scripts illustrate the procedures.

\begin{lstlisting}[language=Python, caption={Sentence extraction and preprocessing from Wikipedia summaries.}, label={lst:split_text}]
import pandas as pd
import re
from datasets import load_dataset
import random

df = pd.DataFrame(load_dataset("jordiclive/wikipedia-summary-dataset")['train'].shuffle(seed=42).select(range(10000))
df = df.drop(columns=['full_text'])

def split_into_sentences(text):
    sentences = re.split(r'(?<=[.!?]) +', text)
    return sentences

df['sentence'] = df['summary'].apply(split_into_sentences)
df = df.explode('sentence')
df = df[df['sentence'].apply(lambda x: len(x.split()) >= 30)]
df = df.reset_index(drop=True)
df.to_csv("file/wikipedia_summary_sentences.csv", index=False)
\end{lstlisting}

\newpage

\begin{lstlisting}[language=Python, caption={Conditionals sentence generation.}, label={lst:conditionals}]
def generate_conditionals(sentence):
    prompt = f"""
    Based on the sentence below, write a conditional sentence that uses the main topic of the sentence. 
    The conditional sentence should express a hypothetical situation or cause-effect relationship related to the topic. It can be slightly complex in structure. 
    For example:
    - If it rains tomorrow, I will stay home.
    Sentence:
    '{sentence}'
    """
    
    completion = client.chat.completions.create(
        model="gpt-4o-mini",
        messages=[
            {"role": "system", "content": "You are a helpful assistant that specializes in generating conditional sentences."},
            {"role": "user", "content": prompt}
        ]
    )

    return completion.choices[0].message.content
\end{lstlisting}

\begin{lstlisting}[language=Python, caption={Contradiction generation.}, label={lst:contradiction}]
def generate_contradiction(sentence):
    prompt = f"""
    You will be given a sentence. Generate a contradictory sentence that directly conflicts with the original sentence without using standard negation.

    Definitions:
    - Standard negation: Directly negating the main verb or using words like 'not', 'no', 'never', or negative contractions such as \"isn't\", \"doesn't\", or \"can't\".
    - Contradiction: A sentence that logically conflicts with the original statement. The contradiction must be such that both sentences cannot logically be true at the same time under any circumstances.

    Important:
    - Do not change the main verb from the original sentence.
    - Do not use 'never' or other negative words to form the contradiction.
    - Ensure the contradicted sentence logically excludes the possibility of the original sentence being true simultaneously.

    Examples:
    Original sentence: \"The tallest student won the award.\"
    Contradicted sentence: \"The shortest student won the award.\"

    Original sentence: \"The room was completely dark.\"
    Contradicted sentence: \"The room was brightly lit.\"

    Original sentence: \"The event took place in the morning.\"
    Contradicted sentence: \"The event took place in the evening.\"

    Original sentence: \"All people are dying.\"
    Contradicted sentence: \"Some people are dying.\"

    Now, generate a contradictory sentence without standard negation, without changing the main verb, and ensuring the two sentences are logically incompatible, for the following:

    Original sentence: \"{sentence}\"

    Contradicted sentence:
    """

    completion = client.chat.completions.create(
        model="gpt-4o",
        messages=[
            {"role": "system", "content": "You are a helpful assistant tasked with generating logical contradictions. Do not use negation to make contradiction."},
            {"role": "user", "content": prompt}
        ]
    )

    return completion.choices[0].message.content 
\end{lstlisting}

\begin{lstlisting}[language=Python, caption={Paraphrase generation.}, label={lst:paraphrase}]
def generate_paraphrase(sentence):
    prompt = f"""
        Paraphrase the following sentence using synonyms or slight structural variations without changing its meaning. 
        Do not add or remove any main verbs. Keep the original intent of the sentence intact.

        Original sentence: "{sentence}"

        Paraphrased sentence:
        """

    completion = client.chat.completions.create(
        model="gpt-4o",
        messages=[
            {"role": "system", "content": "You are a helpful assistant skilled at generating paraphrases while keeping the meaning of sentences unchanged."},
            {
                "role": "user",
                "content": prompt
            }
        ]
    )

    return completion.choices[0].message.content  
\end{lstlisting}

\begin{table*}[htbp]
\resizebox{\textwidth}{!}{%
\begin{tabular}{@{}llrr@{}}
\toprule
\multicolumn{1}{c}{\textbf{choice2\_type}} & \multicolumn{1}{c}{\textbf{Definition}} & \multicolumn{1}{c}{\textbf{\begin{tabular}[c]{@{}c@{}}Demonstration Set\end{tabular}}} & \multicolumn{1}{c}{\textbf{Test Set}} \\ \midrule\midrule
\textbf{relative\_part} & \begin{tabular}[c]{@{}l@{}}negation inside relative clauses \\ (e.g., "who did not attend…").\end{tabular} & 12 & 312 \\ \midrule
\textbf{pp\_part} & \begin{tabular}[c]{@{}l@{}}negation in participle clauses\\ (e.g., "not walking through the park…").\end{tabular} & 12 & 308 \\ \midrule
\textbf{adverb\_part} & \begin{tabular}[c]{@{}l@{}}negation in adverbial clauses\\ (e.g., "because it was not raining").\end{tabular} & 12 & 310 \\ \midrule
\textbf{compound\_part} & negation applied to one clause within a compound sentence. & 12 & 294 \\ \midrule
\textbf{non-applicable} & \begin{tabular}[c]{@{}l@{}}used when the sentence structure does not support\\ a valid local negation variant under our definition.\end{tabular} & 2 & 37 \\ \midrule \midrule
 & \textbf{Total} & 50 & 1,261 \\ \cmidrule(l){2-4} 
\end{tabular}%
}
\caption{Choice 2 Types and Distributions.}
\label{tab:choice2_type}
\end{table*}

\section{\bench Dataset Structure}
\bench consists of two subsets: a sentence-negation pair dataset for supervised fine-tuning and a multiple-choice dataset for evaluation. Both datasets are built on English text and reviewed by authors following strict guidelines.

\subsection{Sentence-Negation Pair Dataset}
This subset contains pairs of affirmative and corresponding standard negation sentences. It includes the following fields:

\begin{itemize}[nosep]
    \item \texttt{\small index}: the index of the data.
    \item \texttt{\small premise}: the original sentence.
    \item \texttt{\small hypothesis}: its logically negated form.
\end{itemize}

\subsection{Multiple-Choice Dataset}

This evaluation set presents each original sentence with four candidate transformations.
\begin{itemize}[nosep]
    \item \texttt{\small wikipedia\_index}: the original index of the Wikipedia Summary dataset.
    \item \texttt{\small index}: the index of the data.
    \item \texttt{\small sentence}: the original sentence.
    \item \texttt{\small choice1}: standard negation (correct answer).
    \item \texttt{\small choice2}: local negation (subclausal negation).
    \item \texttt{\small choice2\_type}: specifies the type of local negation.
    \item \texttt{\small choice2\_element}: a short description of the phrase or clause that was negated (e.g., "being built", "which crashed").
    \item \texttt{\small choice3}: contradiction (non-negated, semantically incompatible).
    \item \texttt{\small choice4}: paraphrase (semantically equivalent).
\end{itemize}

The details of \texttt{\small choice2\_type} and distribution on demonstration and test sets are described in Table~\ref{tab:choice2_type}. It follows the definition in Table~\ref{tab:localtype}.

In addition to the Wikipedia Summary dataset, we supplement the evaluation set with conditionals (e.g., If $P$, $Q$) by manually searching Wikipedia articles where such constructions are more likely to occur (e.g., Newton's laws of motion). Among the 100 conditional sentences included across demonstration and test sets, 20 are collected through manual search (marked with indices beginning with "S" in \texttt{\small wikipedia\_index}). Meanwhile, the remaining 80 are sampled from the Wikipedia Summary dataset and converted into conditional form using the script in~\autoref{lst:conditionals} (Appendix~\ref{appendix:datacode}).

\section{Prompt Templates Used in Evaluation}
\label{appendix:prompt_select}
We design and employ two instruction formats for in-context learning. These instructions differ in the level of guidance they provide, ranging from a task definition to a detailed procedural instruction. Specifically, we use the following two instruction styles. These instruction formats correspond to the \texttt{\small task instruction} component shown in Figure~\ref{fig:intro_example}.
\begin{enumerate}
    \item \textbf{Definition instruction}: 
    A concise definition of standard negation, which specifies the target operation, \\"\texttt{\small Standard negation is sentential negation that reverses the truth value of the sentence by negating the main predicate(s) of the main clause(s). Keep the rest of the sentence content unchanged.}"
    \item \textbf{Detailed instruction}:
    A step-by-step instruction that presents standard negation as an explicit procedure. This prompt describes how to identify the main clause and its main predicate, preserve other sentence elements, apply syntactic negation or complementary antonyms where appropriate, and negate complex sentences by reversing their logical structure. The full prompt format is shown in Listing~\ref{lst:detail-prompt}.
\end{enumerate}

\begin{lstlisting}[basicstyle=\ttfamily\scriptsize, breaklines=true, frame=single, caption={Detailed instruction format.}, label={lst:detail-prompt}, numbers=none, xleftmargin=0.05\textwidth]
Standard negation reverses the truth value of the main predicate in the main clause while keeping all other elements of the main clause unchanged.
Do not negate subordinate clauses or modify other parts of the sentence.

To do this:
1) Identify the main clause and its main verb (main predicate). Ignore subordinate clauses.
2) Preserve all other main-clause content.
3) Insert a negative particle such as "not" into the main verb, or replace it with a complementary antonym only if it forms an absolute binary (e.g., alive/dead, true/false, possible/impossible).
4) If the sentence contains multiple propositions connected by logical operators (e.g., and, or, conditional constructions), negate it in a way that reverses the entire proposition (e.g., A and B -> not A or not B; If A then B -> A and not B).
\end{lstlisting}

Below, we show the prompt formats used for completion-based and option-selection evaluations, including prompt–response structures for each evaluation setting.

\begin{lstlisting}[basicstyle=\ttfamily\scriptsize, breaklines=true, frame=single, caption={Completion-based Format.}, label={lst:completion_prompt}, numbers=none, xleftmargin=0.01\textwidth]
Prompt] "{instruction format}

Generate the standard negation of the given sentence.
Sentence: Chromosome 2 is the second-largest human chromosome, spanning more than 242 million base pairs and representing almost eight percent of the total DNA in human cells.
Negation:"

Response] "Chromosome 2 isn't the second-largest human chromosome, which measures more than 242 million base pairs and represents almost eight percent of the entire DNA in human cells."
\end{lstlisting}

\newpage

\begin{lstlisting}[basicstyle=\ttfamily\scriptsize, breaklines=true, frame=single, caption={Option-selection Format.}, label={lst:option_prompt}, numbers=none, xleftmargin=0.01\textwidth]
Prompt] "Given the following instruction and candidate answers, choose the single best answer.

{instruction format}

Generate the standard negation of the given sentence.
Sentence: Chromosome 2 is the second-largest human chromosome, spanning more than 242 million base pairs and representing almost eight percent of the total DNA in human cells.

A. Chromosome 2 isn't the second-largest human chromosome, which measures more than 242 million base pairs and represents almost eight percent of the entire DNA in human cells.
B. Chromosome 2 is the second-largest human chromosome, which doesn't span more than 242 million base pairs or represent nearly eight percent of the whole DNA in human cells.
C. Chromosome 2 is the smallest human chromosome, spanning fewer than 50 million base pairs and representing less than two percent of the total DNA in human cells.
D. Chromosome 2 is the second-biggest human chromosome, with over 242 million base pairs, making up nearly 8% of all DNA in human cells.

Your response should be one of A, B, C, D.
Only output the letter.
Answer:"

Response] "A"

\end{lstlisting}

\section{Evaluation Setup}
\label{appendix:details_eval}
We use 8 NVIDIA GeForce RTX 3090 GPUs with 24GB of memory and CUDA 12.4 for all fine-tuning and evaluation.

\subsection{Models}
We evaluate a diverse set of pretrained and instruction-tuned models across multiple model families and scales, ranging from about a billion to 14 billion parameters. Our experiments include models from Gemma~\citep{team2024gemma}, Qwen~\citep{yang2025qwen3}, LLaMA~\citep{grattafiori2024llama}, and Mistral families~\citep{jiang2023mistral7b}, as well as API-based models from OpenAI~\citep{achiam2023gpt, hurst2024gpt} and Anthropic~\citep{TheC3}. API models are evaluated only in the option-selection setting, as they do not provide access to token-level log probabilities required for completion-based evaluation.

\begin{itemize}[nosep]
    \item \textbf{Gemma models:}
    \begin{itemize}[nosep]
        \item gemma-2-2b, gemma-2-2b-it
        \item gemma-2-9b, gemma-2-9b-it
    \end{itemize}

    \item \textbf{Qwen models:}
    \begin{itemize}[nosep]
        \item Qwen3-0.6B-Base, Qwen3-0.6B
        \item Qwen3-1.7B-Base, Qwen3-1.7B
        \item Qwen3-4B-Base, Qwen3-4B
        \item Qwen3-8B-Base, Qwen3-8B
        \item Qwen3-14B-Base, Qwen3-14B
        \item Qwen3-32B
    \end{itemize}

    \item \textbf{LLaMA models:}
    \begin{itemize}[nosep]
        \item Llama-3.1-8B, Llama-3.1-8B-Instruct
        \item Llama-3.2-1B, Llama-3.2-1B-Instruct
        \item Llama-3.2-3B, Llama-3.2-3B-Instruct
    \end{itemize}

    \item \textbf{Mistral models:}
    \begin{itemize}[nosep]
        \item Mistral-7B-v0.3, Mistral-7B-Instruct-v0.3
        \item Mistral-Nemo-Base-2407 (12B size), Mistral-Nemo-Instruct-2407 (12B size)
        \item Mistral-Small-24B-Base-2501, Mistral-Small-24B-Instruct-2501
    \end{itemize}

    \item \textbf{API models:}
    \begin{itemize}[nosep]
        \item GPT: GPT-4o mini, GPT-4o, GPT-4.1 mini, GPT-4.1
        \item Claude: Haiku 4.5, Sonnet 4.5
    \end{itemize}
\end{itemize}

\subsection{Zero-shot and few-shot evaluation}
For each model, we evaluate performance in both zero-shot and few-shot settings using the Language Model Evaluation Harness~\citep{eval-harness}. In the few-shot scenario, we use examples from the demonstration set as in-context demonstrations. Results are averaged over five random seeds (42, 1234, 3000, 5000, and 7000) and are reported for one, five, and ten examples from the demonstration set (1-shot, 5-shot, and 10-shot). We present the performance results on the test set for each model and prompt configuration.

\subsection{Supervised fine-tuning}
We conduct Supervised Fine-Tuning (SFT) using the LLaMA-Factory framework~\citep{zheng2024llamafactory} on the Sentence-Negation Pair dataset from \bench. The dataset is formatted in the Alpaca instruction style~\citep{alpaca}. To achieve parameter-efficient training, we apply Low-Rank Adaptation (LoRA)~\citep{hu2022lora} with a rank of 8, targeting all linear layers. The fine-tuning process is carried out for three epochs, using a batch size of 1, a gradient accumulation step of 8, cosine learning rate scheduling, and bfloat16 precision. We apply supervised fine-tuning only to models with fewer than 10B parameters, in order to focus on analyzing the effects of SFT on relatively smaller models, where instruction tuning is expected to have a more pronounced impact. After the SFT, we evaluate the model's zero-shot performance to directly assess its ability to generalize from instruction tuning without being influenced by in-context examples. It is important to note that API models are not fine-tuned, as they are not compatible with the LLaMA-Factory framework.

\section{Total Model Performance on \bench}
\label{appendix:result_bench}

This section provides the full set of results on \bench for all evaluated models and training settings. Following the prompt templates in Appendix~\ref{appendix:prompt_select}, we report results separately for the definition instruction and detailed instruction styles. 

For each instruction style, models are evaluated under three conditions: (1) zero-shot, (2) few-shot with 1, 5, and 10 in-context demonstrations (averaged across random seeds), and (3) supervised fine-tuning (SFT). For SFT, models are evaluated in the zero-shot setting to isolate the effect
of task-specific training without the influence of in-context examples.

Tables~\ref{tab:total_def} and~\ref{tab:total_detail} report the results of open-weight models under the definition and detailed instruction styles, respectively.
Tables~\ref{tab:total_def_api} and~\ref{tab:total_detail_api} present the corresponding results for API models under the same instruction settings.

\section{General Benchmark Performance after SFT}
\label{appendix:general_sft}
To assess whether supervised fine-tuning (SFT) on \bench affects performance on broader natural language understanding tasks, we evaluate models on six widely used benchmarks: ARC-Challenge, ARC-Easy, GSM8K, HellaSwag, OpenBookQA, and WinoGrande~\citep{clark2018think, cobbe2021training, zellers2019hellaswag, mihaylov2018can, sakaguchi2021winogrande}. These datasets cover a diverse range of domains, including commonsense reasoning, scientific knowledge, mathematics, and reading comprehension.

Table~\ref{tab:general_sft_def} and Table~\ref{tab:general_sft_detail} summarize the results. We find that performance on general benchmarks remains broadly stable after SFT, indicating that fine-tuning on \bench does not substantially harm general capabilities.

\section{Total Error Analysis of Model Predictions on the \bench}
\label{appendix:result_analysis}

This section extends the error analysis presented in Section~\ref{sec:5_3_analysis}, providing the complete results. In particular, we examine (i) incorrect choice distributions and (ii) confusion rates for local negation categories, comparing models of different sizes and training paradigms (pretrained vs. instruction-tuned) under both zero-shot and few-shot conditions. 

\begin{table}[h!]
\centering
\resizebox{\columnwidth}{!}{%
\begin{tabular}{@{}lcccc@{}}
\toprule
\multicolumn{1}{c}{\multirow{2}{*}{\textbf{}}} & \multicolumn{2}{c}{\textbf{definition}} & \multicolumn{2}{c}{\textbf{detailed}} \\ \cmidrule(l){2-5} 
\multicolumn{1}{c}{} & \textbf{completion} & \textbf{option} & \textbf{completion} & \textbf{option} \\ \midrule
\textbf{Gemma2} & Table~\ref{tab:neg_gemma_def_completion} & Table~\ref{tab:neg_gemma_def_option} & Table~\ref{tab:neg_gemma_detail_completion} & Table~\ref{tab:neg_gemma_detail_option} \\
\textbf{Qwen3} & Table~\ref{tab:neg_qwen_def_completion} & Table~\ref{tab:neg_qwen_def_option} & Table~\ref{tab:neg_qwen_detail_completion} & Table~\ref{tab:neg_qwen_detail_option} \\
\textbf{Llama3} & Table~\ref{tab:neg_llama_def_completion} & Table~\ref{tab:neg_llama_def_option} & Table~\ref{tab:neg_llama_detail_completion} & Table~\ref{tab:neg_llama_detail_option} \\
\textbf{Mistral} & Table~\ref{tab:neg_mistral_def_completion} & Table~\ref{tab:neg_mistral_def_option} & Table~\ref{tab:neg_mistral_detail_completion} & Table~\ref{tab:neg_mistral_detail_option} \\
\textbf{API} &  & Table~\ref{tab:neg_api_def} &  & Table~\ref{tab:neg_api_detail} \\ \bottomrule
\end{tabular}%
}
\caption{Complete prediction analysis by model family.}
\end{table}

We report few-shot results using a fixed random seed (1234). Averaging over multiple seeds was avoided, as it could obscure specific error patterns and make confusion analysis less interpretable. Each table follows the same instruction format, reporting error rates, incorrect choice distributions, and confusion rates across local negation subcategories under zero-shot, few-shot, and SFT conditions.

For API models, we also track cases labeled as “Answer Format Wrong.” This category captures instances where the model’s output does not follow the required answer format (selecting strictly one of A, B, C, or D). Because such responses cannot be mapped to a specific incorrect option, they are not included in the incorrect choice distribution or confusion rate. Instead, we report their raw counts alongside the other error statistics. This also serves as an indicator of the model’s ability to follow output-format instructions.

\section{Human Evaluation}
\label{appendix:human_eval}

The human evaluation was approved by the Institutional Review Board (IRB No. 2512/004-015). All participants provided informed consent through an official IRB-approved consent form prior to participation and were fully informed of the study’s purpose, procedures, potential risks and benefits, and their right to withdraw at any time without penalty. Participants were also notified that only aggregate statistics of their evaluation scores would be reported.

All responses were collected anonymously, and no personally identifiable information was included in the dataset. Participants were compensated KRW 50,000 for a two-hour session, which exceeds the minimum hourly wage (KRW 10,320 per hour) and constitutes adequate compensation. To minimize fatigue and ensure reliable responses, participants were given sufficient rest during the evaluation, and the total duration was limited to approximately two hours per session. Participants were required to be native or highly proficient English users (CEFR\footnote{Common European Framework of Reference for languages.} C-level or above), as verified during recruitment. A total of 11 adult participants aged in their 20s to 30s took part in the study, including 6 highly proficient English users at the CEFR C level and 5 native or dominant English speakers.

A total of 50 items were randomly sampled from the multiple-choice evaluation split of \bench, with stratified sampling to preserve the distribution of distractor types. Specifically, the sample included 12 items each for relative clause, participle clause, compound sentence, and adverbial clause (local negation type), and 2 items without local negation. All items were presented using the detailed instruction format. Participants were asked to select the option corresponding to standard negation for each item. No example items were provided during the evaluation.

\begin{table}[h!]
\centering
\resizebox{0.7\columnwidth}{!}{%
\begin{tabular}{@{}ccccc@{}}
\toprule
\textbf{Mean} & \textbf{Median} & \textbf{Min} & \textbf{Max} & \textbf{SD} \\ \midrule
0.78 & 0.82 & 0.58 & 0.96 & 0.139 \\ \bottomrule
\end{tabular}%
}
\caption{Human evaluation results.}
\label{tab:human_eval}
\end{table}

Table~\ref{tab:human_eval} summarizes the results of the human evaluation conducted with 11 participants. Human accuracy ranges from 0.58 to 0.96, with a mean of 0.78. This range indicates that distinguishing standard negation from closely related alternatives can be challenging even for human readers, consistent with prior observations that negation processing requires careful semantic and syntactic reasoning. The average human accuracy is higher than the zero-shot performance of most models reported in this paper, suggesting that humans generally handle sentence-level negation more reliably without task-specific training.

\begin{table*}[htbp]
\centering
\resizebox{\textwidth}{!}{%
\begin{tabular}{@{}lrrrrrrrrrr@{}}
\toprule
 &
  \multicolumn{2}{c}{\textbf{zero-shot}} &
  \multicolumn{2}{c}{\textbf{1-shot}} &
  \multicolumn{2}{c}{\textbf{5-shot}} &
  \multicolumn{2}{c}{\textbf{10-shot}} &
  \multicolumn{2}{c}{\textbf{after SFT}} \\ \cmidrule(l){2-11} 
\textbf{evaluation setting} &
  \multicolumn{1}{c}{\textbf{completion}} &
  \multicolumn{1}{c}{\textbf{option}} &
  \multicolumn{1}{c}{\textbf{\begin{tabular}[c]{@{}c@{}}completion\\ (±SD)\end{tabular}}} &
  \multicolumn{1}{c}{\textbf{\begin{tabular}[c]{@{}c@{}}option\\ (±SD)\end{tabular}}} &
  \multicolumn{1}{c}{\textbf{\begin{tabular}[c]{@{}c@{}}completion\\ (±SD)\end{tabular}}} &
  \multicolumn{1}{c}{\textbf{\begin{tabular}[c]{@{}c@{}}option\\ (±SD)\end{tabular}}} &
  \multicolumn{1}{c}{\textbf{\begin{tabular}[c]{@{}c@{}}completion\\ (±SD)\end{tabular}}} &
  \multicolumn{1}{c}{\textbf{\begin{tabular}[c]{@{}c@{}}option\\ (±SD)\end{tabular}}} &
  \multicolumn{1}{c}{\textbf{completion}} &
  \multicolumn{1}{c}{\textbf{option}} \\ \midrule\midrule
\textbf{gemma-2-2b} &
  0.436 &
  0.236 &
  \begin{tabular}[c]{@{}r@{}}0.480\\ (±0.006)\end{tabular} &
  \begin{tabular}[c]{@{}r@{}}0.328\\ (±0.004)\end{tabular} &
  \begin{tabular}[c]{@{}r@{}}0.565\\ (±0.010)\end{tabular} &
  \begin{tabular}[c]{@{}r@{}}0.373\\ (±0.009)\end{tabular} &
  \begin{tabular}[c]{@{}r@{}}0.599\\ (±0.004)\end{tabular} &
  \begin{tabular}[c]{@{}r@{}}0.398\\ (±0.011)\end{tabular} &
  0.781 &
  0.260 \\
\textbf{gemma-2-2b-it} &
  0.516 &
  0.504 &
  \begin{tabular}[c]{@{}r@{}}0.514\\ (±0.008)\end{tabular} &
  \begin{tabular}[c]{@{}r@{}}0.501\\ (±0.008)\end{tabular} &
  \begin{tabular}[c]{@{}r@{}}0.569\\ (±0.007)\end{tabular} &
  \begin{tabular}[c]{@{}r@{}}0.510\\ (±0.012)\end{tabular} &
  \begin{tabular}[c]{@{}r@{}}0.594\\ (±0.003)\end{tabular} &
  \begin{tabular}[c]{@{}r@{}}0.540\\ (±0.007)\end{tabular} &
  0.741 &
  0.543 \\
\textbf{gemma-2-9b} &
  0.470 &
  0.468 &
  \begin{tabular}[c]{@{}r@{}}0.499\\ (±0.005)\end{tabular} &
  \begin{tabular}[c]{@{}r@{}}0.530\\ (±0.009)\end{tabular} &
  \begin{tabular}[c]{@{}r@{}}0.565\\ (±0.007)\end{tabular} &
  \begin{tabular}[c]{@{}r@{}}0.588\\ (±0.006)\end{tabular} &
  \begin{tabular}[c]{@{}r@{}}0.597\\ (±0.005)\end{tabular} &
  \begin{tabular}[c]{@{}r@{}}0.626\\ (±0.005)\end{tabular} &
  0.771 &
  0.751 \\
\textbf{gemma-2-9b-it} &
  0.508 &
  0.554 &
  \begin{tabular}[c]{@{}r@{}}0.562\\ (±0.007)\end{tabular} &
  \begin{tabular}[c]{@{}r@{}}0.631\\ (±0.005)\end{tabular} &
  \begin{tabular}[c]{@{}r@{}}0.653\\ (±0.007)\end{tabular} &
  \begin{tabular}[c]{@{}r@{}}0.705\\ (±0.007)\end{tabular} &
  \begin{tabular}[c]{@{}r@{}}0.689\\ (±0.005)\end{tabular} &
  \begin{tabular}[c]{@{}r@{}}0.724\\ (±0.011)\end{tabular} &
  0.767 &
  {\color{red} 0.814} \\ \midrule
\textbf{Qwen3-0.6B-Base} &
  0.453 &
  0.349 &
  \begin{tabular}[c]{@{}r@{}}0.505\\ (±0.008)\end{tabular} &
  \begin{tabular}[c]{@{}r@{}}0.406\\ (±0.007)\end{tabular} &
  \begin{tabular}[c]{@{}r@{}}0.581\\ (±0.005)\end{tabular} &
  \begin{tabular}[c]{@{}r@{}}0.428\\ (±0.004)\end{tabular} &
  \begin{tabular}[c]{@{}r@{}}0.616\\ (±0.005)\end{tabular} &
  \begin{tabular}[c]{@{}r@{}}0.465\\ (±0.010)\end{tabular} &
  0.708 &
  0.479 \\
\textbf{Qwen3-0.6B} &
  0.416 &
  0.438 &
  \begin{tabular}[c]{@{}r@{}}0.420\\ (±0.006)\end{tabular} &
  \begin{tabular}[c]{@{}r@{}}0.466\\ (±0.004)\end{tabular} &
  \begin{tabular}[c]{@{}r@{}}0.487\\ (±0.003)\end{tabular} &
  \begin{tabular}[c]{@{}r@{}}0.502\\ (±0.006)\end{tabular} &
  \begin{tabular}[c]{@{}r@{}}0.530\\ (±0.006)\end{tabular} &
  \begin{tabular}[c]{@{}r@{}}0.528\\ (±0.007)\end{tabular} &
  0.645 &
  0.555 \\
\textbf{Qwen3-1.7B-Base} &
  0.481 &
  0.526 &
  \begin{tabular}[c]{@{}r@{}}0.525\\ (±0.007)\end{tabular} &
  \begin{tabular}[c]{@{}r@{}}0.559\\ (±0.007)\end{tabular} &
  \begin{tabular}[c]{@{}r@{}}0.577\\ (±0.004)\end{tabular} &
  \begin{tabular}[c]{@{}r@{}}0.641\\ (±0.004)\end{tabular} &
  \begin{tabular}[c]{@{}r@{}}0.604\\ (±0.001)\end{tabular} &
  \begin{tabular}[c]{@{}r@{}}0.676\\ (±0.007)\end{tabular} &
  0.699 &
  0.493 \\
\textbf{Qwen3-1.7B} &
  0.398 &
  0.532 &
  \begin{tabular}[c]{@{}r@{}}0.406\\ (±0.006)\end{tabular} &
  \begin{tabular}[c]{@{}r@{}}0.547\\ (±0.009)\end{tabular} &
  \begin{tabular}[c]{@{}r@{}}0.476\\ (±0.007)\end{tabular} &
  \begin{tabular}[c]{@{}r@{}}0.574\\ (±0.008)\end{tabular} &
  \begin{tabular}[c]{@{}r@{}}0.514\\ (±0.005)\end{tabular} &
  \begin{tabular}[c]{@{}r@{}}0.578\\ (±0.007)\end{tabular} &
  0.668 &
  0.608 \\
\textbf{Qwen3-4B-Base} &
  0.454 &
  0.602 &
  \begin{tabular}[c]{@{}r@{}}0.489\\ (±0.002)\end{tabular} &
  \begin{tabular}[c]{@{}r@{}}0.611\\ (±0.006)\end{tabular} &
  \begin{tabular}[c]{@{}r@{}}0.535\\ (±0.004)\end{tabular} &
  \begin{tabular}[c]{@{}r@{}}0.653\\ (±0.003)\end{tabular} &
  \begin{tabular}[c]{@{}r@{}}0.560\\ (±0.007)\end{tabular} &
  \begin{tabular}[c]{@{}r@{}}0.659\\ (±0.009)\end{tabular} &
  0.733 &
  0.709 \\
\textbf{Qwen3-4B} &
  0.410 &
  0.562 &
  \begin{tabular}[c]{@{}r@{}}0.461\\ (±0.008)\end{tabular} &
  \begin{tabular}[c]{@{}r@{}}0.619\\ (±0.005)\end{tabular} &
  \begin{tabular}[c]{@{}r@{}}0.548\\ (±0.010)\end{tabular} &
  \begin{tabular}[c]{@{}r@{}}0.676\\ (±0.005)\end{tabular} &
  \begin{tabular}[c]{@{}r@{}}0.597\\ (±0.007)\end{tabular} &
  \begin{tabular}[c]{@{}r@{}}0.708\\ (±0.006)\end{tabular} &
  0.715 &
  0.722 \\
\textbf{Qwen3-8B-Base} &
  0.474 &
  0.690 &
  \begin{tabular}[c]{@{}r@{}}0.505\\ (±0.004)\end{tabular} &
  \begin{tabular}[c]{@{}r@{}}0.692\\ (±0.003)\end{tabular} &
  \begin{tabular}[c]{@{}r@{}}0.566\\ (±0.009)\end{tabular} &
  \begin{tabular}[c]{@{}r@{}}0.741\\ (±0.008)\end{tabular} &
  \begin{tabular}[c]{@{}r@{}}0.603\\ (±0.009)\end{tabular} &
  \begin{tabular}[c]{@{}r@{}}0.758\\ (±0.007)\end{tabular} &
  0.703 &
  0.768 \\
\textbf{Qwen3-8B} &
  0.442 &
  0.589 &
  \begin{tabular}[c]{@{}r@{}}0.487\\ (±0.011)\end{tabular} &
  \begin{tabular}[c]{@{}r@{}}0.664\\ (±0.009)\end{tabular} &
  \begin{tabular}[c]{@{}r@{}}0.556\\ (±0.008)\end{tabular} &
  \begin{tabular}[c]{@{}r@{}}0.712\\ (±0.006)\end{tabular} &
  \begin{tabular}[c]{@{}r@{}}0.597\\ (±0.006)\end{tabular} &
  \begin{tabular}[c]{@{}r@{}}0.740\\ (±0.005)\end{tabular} &
  0.722 &
  0.784 \\
\textbf{Qwen3-14B-Base} &
  0.489 &
  0.708 &
  \begin{tabular}[c]{@{}r@{}}0.547\\ (±0.006)\end{tabular} &
  \begin{tabular}[c]{@{}r@{}}0.766\\ (±0.013)\end{tabular} &
  \begin{tabular}[c]{@{}r@{}}0.613\\ (±0.005)\end{tabular} &
  {\begin{tabular}[c]{@{}r@{}}\color{red}{0.806}\\ (±0.005)\end{tabular}} &
  \begin{tabular}[c]{@{}r@{}}0.653\\ (±0.007)\end{tabular} &
  {\begin{tabular}[c]{@{}r@{}}\color{red}{0.821}\\ (±0.004)\end{tabular}} &
  - &
  - \\
\textbf{Qwen3-14B} &
  0.440 &
  0.661 &
  \begin{tabular}[c]{@{}r@{}}0.492\\ (±0.008)\end{tabular} &
  \begin{tabular}[c]{@{}r@{}}0.677\\ (±0.009)\end{tabular} &
  \begin{tabular}[c]{@{}r@{}}0.569\\ (±0.004)\end{tabular} &
  \begin{tabular}[c]{@{}r@{}}0.732\\ (±0.007)\end{tabular} &
  \begin{tabular}[c]{@{}r@{}}0.622\\ (±0.006)\end{tabular} &
  \begin{tabular}[c]{@{}r@{}}0.751\\ (±0.003)\end{tabular} &
  - &
  - \\
\textbf{Qwen3-32B} &
  0.470 &
  {\color{red} 0.730} &
  \begin{tabular}[c]{@{}r@{}}0.534\\ (±0.011)\end{tabular} &
  {\begin{tabular}[c]{@{}r@{}}\color{red}{0.768}\\ (±0.010)\end{tabular}} &
  \begin{tabular}[c]{@{}r@{}}0.650\\ (±0.009)\end{tabular} &
  \begin{tabular}[c]{@{}r@{}}0.792\\ (±0.009)\end{tabular} &
  \begin{tabular}[c]{@{}r@{}}0.690\\ (±0.006)\end{tabular} &
  \begin{tabular}[c]{@{}r@{}}0.817\\ (±0.003)\end{tabular} &
  - &
  - \\ \midrule
\textbf{Llama-3.1-8B} &
  0.464 &
  0.474 &
  \begin{tabular}[c]{@{}r@{}}0.495\\ (±0.006)\end{tabular} &
  \begin{tabular}[c]{@{}r@{}}0.511\\ (±0.005)\end{tabular} &
  \begin{tabular}[c]{@{}r@{}}0.593\\ (±0.010)\end{tabular} &
  \begin{tabular}[c]{@{}r@{}}0.580\\ (±0.011)\end{tabular} &
  \begin{tabular}[c]{@{}r@{}}0.643\\ (±0.008)\end{tabular} &
  \begin{tabular}[c]{@{}r@{}}0.629\\ (±0.010)\end{tabular} &
  0.774 &
  0.559 \\
\textbf{\begin{tabular}[c]{@{}l@{}}Llama-3.1-8B-\\ Instruct\end{tabular}} &
  0.464 &
  0.538 &
  \begin{tabular}[c]{@{}r@{}}0.540\\ (±0.005)\end{tabular} &
  \begin{tabular}[c]{@{}r@{}}0.595\\ (±0.012)\end{tabular} &
  \begin{tabular}[c]{@{}r@{}}0.628\\ (±0.006)\end{tabular} &
  \begin{tabular}[c]{@{}r@{}}0.652\\ (±0.009)\end{tabular} &
  \begin{tabular}[c]{@{}r@{}}0.668\\ (±0.006)\end{tabular} &
  \begin{tabular}[c]{@{}r@{}}0.673\\ (±0.005)\end{tabular} &
  0.771 &
  0.644 \\
\textbf{Llama-3.2-1B} &
  0.409 &
  0.259 &
  \begin{tabular}[c]{@{}r@{}}0.431\\ (±0.007)\end{tabular} &
  \begin{tabular}[c]{@{}r@{}}0.263\\ (±0.011)\end{tabular} &
  \begin{tabular}[c]{@{}r@{}}0.492\\ (±0.006)\end{tabular} &
  \begin{tabular}[c]{@{}r@{}}0.250\\ (±0.010)\end{tabular} &
  \begin{tabular}[c]{@{}r@{}}0.526\\ (±0.004)\end{tabular} &
  \begin{tabular}[c]{@{}r@{}}0.262\\ (±0.009)\end{tabular} &
  0.750 &
  0.259 \\
\textbf{\begin{tabular}[c]{@{}l@{}}Llama-3.2-1B-\\ Instruct\end{tabular}} &
  0.431 &
  0.301 &
  \begin{tabular}[c]{@{}r@{}}0.468\\ (±0.006)\end{tabular} &
  \begin{tabular}[c]{@{}r@{}}0.376\\ (±0.013)\end{tabular} &
  \begin{tabular}[c]{@{}r@{}}0.526\\ (±0.007)\end{tabular} &
  \begin{tabular}[c]{@{}r@{}}0.351\\ (±0.009)\end{tabular} &
  \begin{tabular}[c]{@{}r@{}}0.548\\ (±0.009)\end{tabular} &
  \begin{tabular}[c]{@{}r@{}}0.336\\ (±0.012)\end{tabular} &
  0.761 &
  0.286 \\
\textbf{Llama-3.2-3B} &
  0.435 &
  0.362 &
  \begin{tabular}[c]{@{}r@{}}0.469\\ (±0.005)\end{tabular} &
  \begin{tabular}[c]{@{}r@{}}0.443\\ (±0.004)\end{tabular} &
  \begin{tabular}[c]{@{}r@{}}0.563\\ (±0.006)\end{tabular} &
  \begin{tabular}[c]{@{}r@{}}0.472\\ (±0.003)\end{tabular} &
  \begin{tabular}[c]{@{}r@{}}0.588\\ (±0.006)\end{tabular} &
  \begin{tabular}[c]{@{}r@{}}0.483\\ (±0.004)\end{tabular} &
  0.757 &
  0.385 \\
\textbf{\begin{tabular}[c]{@{}l@{}}Llama-3.2-3B-\\ Instruct\end{tabular}} &
  0.452 &
  0.512 &
  \begin{tabular}[c]{@{}r@{}}0.456\\ (±0.007)\end{tabular} &
  \begin{tabular}[c]{@{}r@{}}0.539\\ (±0.007)\end{tabular} &
  \begin{tabular}[c]{@{}r@{}}0.548\\ (±0.005)\end{tabular} &
  \begin{tabular}[c]{@{}r@{}}0.499\\ (±0.002)\end{tabular} &
  \begin{tabular}[c]{@{}r@{}}0.590\\ (±0.004)\end{tabular} &
  \begin{tabular}[c]{@{}r@{}}0.493\\ (±0.005)\end{tabular} &
  {\color{red} 0.795} &
  0.651 \\ \midrule
\textbf{Mistral-7B-v0.3} &
  0.467 &
  0.485 &
  \begin{tabular}[c]{@{}r@{}}0.464\\ (±0.006)\end{tabular} &
  \begin{tabular}[c]{@{}r@{}}0.494\\ (±0.015)\end{tabular} &
  \begin{tabular}[c]{@{}r@{}}0.557\\ (±0.014)\end{tabular} &
  \begin{tabular}[c]{@{}r@{}}0.631\\ (±0.007)\end{tabular} &
  \begin{tabular}[c]{@{}r@{}}0.599\\ (±0.005)\end{tabular} &
  \begin{tabular}[c]{@{}r@{}}0.653\\ (±0.007)\end{tabular} &
  0.784 &
  0.480 \\
\textbf{\begin{tabular}[c]{@{}l@{}}Mistral-7B-Instruct-\\ v0.3\end{tabular}} &
  {\color{red} 0.611} &
  0.657 &
  {\begin{tabular}[c]{@{}r@{}}\color{red}{0.643}\\ (±0.006)\end{tabular}} &
  \begin{tabular}[c]{@{}r@{}}0.643\\ (±0.003)\end{tabular} &
  {\begin{tabular}[c]{@{}r@{}}\color{red}{0.694}\\ (±0.005)\end{tabular}} &
  \begin{tabular}[c]{@{}r@{}}0.638\\ (±0.009)\end{tabular} &
  {\begin{tabular}[c]{@{}r@{}}\color{red}{0.715}\\ (±0.008)\end{tabular}} &
  \begin{tabular}[c]{@{}r@{}}0.650\\ (±0.008)\end{tabular} &
  0.793 &
  0.692 \\
\textbf{\begin{tabular}[c]{@{}l@{}}Mistral-Nemo-\\ Base-2407 (12B)\end{tabular}} &
  0.460 &
  0.412 &
  \begin{tabular}[c]{@{}r@{}}0.490\\ (±0.010)\end{tabular} &
  \begin{tabular}[c]{@{}r@{}}0.520\\ (±0.007)\end{tabular} &
  \begin{tabular}[c]{@{}r@{}}0.600\\ (±0.003)\end{tabular} &
  \begin{tabular}[c]{@{}r@{}}0.568\\ (±0.008)\end{tabular} &
  \begin{tabular}[c]{@{}r@{}}0.649\\ (±0.007)\end{tabular} &
  \begin{tabular}[c]{@{}r@{}}0.599\\ (±0.009)\end{tabular} &
  - &
  - \\
\textbf{\begin{tabular}[c]{@{}l@{}}Mistral-Nemo-\\ Instruct-2407 (12B)\end{tabular}} &
  0.487 &
  0.604 &
  \begin{tabular}[c]{@{}r@{}}0.528\\ (±0.004)\end{tabular} &
  \begin{tabular}[c]{@{}r@{}}0.630\\ (±0.007)\end{tabular} &
  \begin{tabular}[c]{@{}r@{}}0.633\\ (±0.009)\end{tabular} &
  \begin{tabular}[c]{@{}r@{}}0.660\\ (±0.006)\end{tabular} &
  \begin{tabular}[c]{@{}r@{}}0.677\\ (±0.007)\end{tabular} &
  \begin{tabular}[c]{@{}r@{}}0.655\\ (±0.006)\end{tabular} &
  - &
  - \\
\textbf{\begin{tabular}[c]{@{}l@{}}Mistral-Small-24B-\\ Base-2501\end{tabular}} &
  0.474 &
  0.568 &
  \begin{tabular}[c]{@{}r@{}}0.500\\ (±0.003)\end{tabular} &
  \begin{tabular}[c]{@{}r@{}}0.618\\ (±0.005)\end{tabular} &
  \begin{tabular}[c]{@{}r@{}}0.597\\ (±0.006)\end{tabular} &
  \begin{tabular}[c]{@{}r@{}}0.721\\ (±0.005)\end{tabular} &
  \begin{tabular}[c]{@{}r@{}}0.647\\ (±0.007)\end{tabular} &
  \begin{tabular}[c]{@{}r@{}}0.783\\ (±0.003)\end{tabular} &
  - &
  - \\
\textbf{\begin{tabular}[c]{@{}l@{}}Mistral-Small-24B-\\ Instruct-2501\end{tabular}} &
  0.526 &
  0.686 &
  \begin{tabular}[c]{@{}r@{}}0.569\\ (±0.006)\end{tabular} &
  \begin{tabular}[c]{@{}r@{}}0.705\\ (±0.005)\end{tabular} &
  \begin{tabular}[c]{@{}r@{}}0.637\\ (±0.009)\end{tabular} &
  \begin{tabular}[c]{@{}r@{}}0.731\\ (±0.011)\end{tabular} &
  \begin{tabular}[c]{@{}r@{}}0.688\\ (±0.006)\end{tabular} &
  \begin{tabular}[c]{@{}r@{}}0.780\\ (±0.003)\end{tabular} &
  - &
  - \\ \midrule\midrule
\textbf{total average} &
  \textbf{0.464} &
  \textbf{0.519} &
  \textbf{0.499} &
  \textbf{0.559} &
  \textbf{0.577} &
  \textbf{0.599} &
  \textbf{0.615} &
  \textbf{0.622} &
  \textbf{0.740} &
  \textbf{0.554} \\ \bottomrule
\end{tabular}%
}
\caption{Zero-shot, few-shot, and SFT evaluation results on \bench with the \textbf{definition instruction}.
SD denotes standard deviation across random seeds or runs. 
Few-shot results are averaged over five random seeds (42, 1234, 3000, 5000, and 7000) and one, five, and ten demonstration examples (1-, 5-, 10-shot). Red text indicates the model with the highest performance in each column.}
\label{tab:total_def}
\end{table*}

\begin{table*}[htbp]
\centering
\resizebox{\textwidth}{!}{%
\begin{tabular}{@{}lrrrrrrrrrr@{}}
\cmidrule(l){2-11}
\multicolumn{1}{c}{\textbf{}} &
  \multicolumn{2}{c}{\textbf{zero-shot}} &
  \multicolumn{2}{c}{\textbf{1-shot}} &
  \multicolumn{2}{c}{\textbf{5-shot}} &
  \multicolumn{2}{c}{\textbf{10-shot}} &
  \multicolumn{2}{c}{\textbf{after SFT}} \\ \midrule
\multicolumn{1}{c}{\textbf{evaluation setting}} &
  \multicolumn{1}{c}{\textbf{completion}} &
  \multicolumn{1}{c}{\textbf{option}} &
  \multicolumn{1}{c}{\textbf{\begin{tabular}[c]{@{}c@{}}completion\\ (±SD)\end{tabular}}} &
  \multicolumn{1}{c}{\textbf{\begin{tabular}[c]{@{}c@{}}option\\ (±SD)\end{tabular}}} &
  \multicolumn{1}{c}{\textbf{\begin{tabular}[c]{@{}c@{}}completion\\ (±SD)\end{tabular}}} &
  \multicolumn{1}{c}{\textbf{\begin{tabular}[c]{@{}c@{}}option\\ (±SD)\end{tabular}}} &
  \multicolumn{1}{c}{\textbf{\begin{tabular}[c]{@{}c@{}}completion\\ (±SD)\end{tabular}}} &
  \multicolumn{1}{c}{\textbf{\begin{tabular}[c]{@{}c@{}}option\\ (±SD)\end{tabular}}} &
  \multicolumn{1}{c}{\textbf{completion}} &
  \multicolumn{1}{c}{\textbf{option}} \\ \midrule \midrule
\textbf{gemma-2-2b} &
  0.438 &
  0.276 &
  \begin{tabular}[c]{@{}r@{}}0.494\\ (±0.005)\end{tabular} &
  \begin{tabular}[c]{@{}r@{}}0.343\\ (±0.005)\end{tabular} &
  \begin{tabular}[c]{@{}r@{}}0.570\\ (±0.004)\end{tabular} &
  \begin{tabular}[c]{@{}r@{}}0.360\\ (±0.018)\end{tabular} &
  \begin{tabular}[c]{@{}r@{}}0.604\\ (±0.004)\end{tabular} &
  \begin{tabular}[c]{@{}r@{}}0.378\\ (±0.010)\end{tabular} &
  0.768 &
  0.259 \\
\textbf{gemma-2-2b-it} &
  0.505 &
  0.470 &
  \begin{tabular}[c]{@{}r@{}}0.488\\ (±0.013)\end{tabular} &
  \begin{tabular}[c]{@{}r@{}}0.474\\ (±0.009)\end{tabular} &
  \begin{tabular}[c]{@{}r@{}}0.545\\ (±0.007)\end{tabular} &
  \begin{tabular}[c]{@{}r@{}}0.487\\ (±0.011)\end{tabular} &
  \begin{tabular}[c]{@{}r@{}}0.580\\ (±0.002)\end{tabular} &
  \begin{tabular}[c]{@{}r@{}}0.500\\ (±0.009)\end{tabular} &
  0.733 &
  0.663 \\
\textbf{gemma-2-9b} &
  0.469 &
  0.466 &
  \begin{tabular}[c]{@{}r@{}}0.513\\ (±0.005)\end{tabular} &
  \begin{tabular}[c]{@{}r@{}}0.530\\ (±0.009)\end{tabular} &
  \begin{tabular}[c]{@{}r@{}}0.570\\ (±0.006)\end{tabular} &
  \begin{tabular}[c]{@{}r@{}}0.568\\ (±0.006)\end{tabular} &
  \begin{tabular}[c]{@{}r@{}}0.604\\ (±0.005)\end{tabular} &
  \begin{tabular}[c]{@{}r@{}}0.606\\ (±0.007)\end{tabular} &
  0.772 &
  0.797 \\
\textbf{gemma-2-9b-it} &
  0.542 &
  0.569 &
  \begin{tabular}[c]{@{}r@{}}0.597\\ (±0.005)\end{tabular} &
  \begin{tabular}[c]{@{}r@{}}0.635\\ (±0.005)\end{tabular} &
  \begin{tabular}[c]{@{}r@{}}0.653\\ (±0.006)\end{tabular} &
  \begin{tabular}[c]{@{}r@{}}0.695\\ (±0.010)\end{tabular} &
  \begin{tabular}[c]{@{}r@{}}0.690\\ (±0.003)\end{tabular} &
  \begin{tabular}[c]{@{}r@{}}0.715\\ (±0.009)\end{tabular} &
  0.764 &
  0.805 \\ \midrule
\textbf{Qwen3-0.6B-Base} &
  0.424 &
  0.342 &
  \begin{tabular}[c]{@{}r@{}}0.487\\ (±0.008)\end{tabular} &
  \begin{tabular}[c]{@{}r@{}}0.397\\ (±0.004)\end{tabular} &
  \begin{tabular}[c]{@{}r@{}}0.564\\ (±0.006)\end{tabular} &
  \begin{tabular}[c]{@{}r@{}}0.416\\ (±0.008)\end{tabular} &
  \begin{tabular}[c]{@{}r@{}}0.607\\ (±0.006)\end{tabular} &
  \begin{tabular}[c]{@{}r@{}}0.461\\ (±0.010)\end{tabular} &
  0.723 &
  0.325 \\
\textbf{Qwen3-0.6B} &
  0.413 &
  0.459 &
  \begin{tabular}[c]{@{}r@{}}0.411\\ (±0.004)\end{tabular} &
  \begin{tabular}[c]{@{}r@{}}0.414\\ (±0.009)\end{tabular} &
  \begin{tabular}[c]{@{}r@{}}0.487\\ (±0.003)\end{tabular} &
  \begin{tabular}[c]{@{}r@{}}0.468\\ (±0.003)\end{tabular} &
  \begin{tabular}[c]{@{}r@{}}0.535\\ (±0.007)\end{tabular} &
  \begin{tabular}[c]{@{}r@{}}0.495\\ (±0.008)\end{tabular} &
  0.679 &
  0.526 \\
\textbf{Qwen3-1.7B-Base} &
  0.462 &
  0.483 &
  \begin{tabular}[c]{@{}r@{}}0.511\\ (±0.007)\end{tabular} &
  \begin{tabular}[c]{@{}r@{}}0.542\\ (±0.004)\end{tabular} &
  \begin{tabular}[c]{@{}r@{}}0.574\\ (±0.005)\end{tabular} &
  \begin{tabular}[c]{@{}r@{}}0.620\\ (±0.002)\end{tabular} &
  \begin{tabular}[c]{@{}r@{}}0.597\\ (±0.005)\end{tabular} &
  \begin{tabular}[c]{@{}r@{}}0.653\\ (±0.003)\end{tabular} &
  0.724 &
  0.609 \\
\textbf{Qwen3-1.7B} &
  0.420 &
  0.447 &
  \begin{tabular}[c]{@{}r@{}}0.442\\ (±0.008)\end{tabular} &
  \begin{tabular}[c]{@{}r@{}}0.502\\ (±0.013)\end{tabular} &
  \begin{tabular}[c]{@{}r@{}}0.490\\ (±0.006)\end{tabular} &
  \begin{tabular}[c]{@{}r@{}}0.547\\ (±0.007)\end{tabular} &
  \begin{tabular}[c]{@{}r@{}}0.522\\ (±0.006)\end{tabular} &
  \begin{tabular}[c]{@{}r@{}}0.556\\ (±0.010)\end{tabular} &
  0.642 &
  0.573 \\
\textbf{Qwen3-4B-Base} &
  0.485 &
  0.585 &
  \begin{tabular}[c]{@{}r@{}}0.503\\ (±0.005)\end{tabular} &
  \begin{tabular}[c]{@{}r@{}}0.581\\ (±0.006)\end{tabular} &
  \begin{tabular}[c]{@{}r@{}}0.546\\ (±0.004)\end{tabular} &
  \begin{tabular}[c]{@{}r@{}}0.636\\ (±0.006)\end{tabular} &
  \begin{tabular}[c]{@{}r@{}}0.566\\ (±0.006)\end{tabular} &
  \begin{tabular}[c]{@{}r@{}}0.649\\ (±0.009)\end{tabular} &
  0.729 &
  0.711 \\
\textbf{Qwen3-4B} &
  0.439 &
  0.577 &
  \begin{tabular}[c]{@{}r@{}}0.489\\ (±0.010)\end{tabular} &
  \begin{tabular}[c]{@{}r@{}}0.635\\ (±0.003)\end{tabular} &
  \begin{tabular}[c]{@{}r@{}}0.559\\ (±0.008)\end{tabular} &
  \begin{tabular}[c]{@{}r@{}}0.652\\ (±0.004)\end{tabular} &
  \begin{tabular}[c]{@{}r@{}}0.600\\ (±0.005)\end{tabular} &
  \begin{tabular}[c]{@{}r@{}}0.685\\ (±0.003)\end{tabular} &
  0.688 &
  0.711 \\
\textbf{Qwen3-8B-Base} &
  0.455 &
  0.668 &
  \begin{tabular}[c]{@{}r@{}}0.513\\ (±0.004)\end{tabular} &
  \begin{tabular}[c]{@{}r@{}}0.700\\ (±0.007)\end{tabular} &
  \begin{tabular}[c]{@{}r@{}}0.557\\ (±0.005)\end{tabular} &
  \begin{tabular}[c]{@{}r@{}}0.737\\ (±0.004)\end{tabular} &
  \begin{tabular}[c]{@{}r@{}}0.591\\ (±0.009)\end{tabular} &
  \begin{tabular}[c]{@{}r@{}}0.755\\ (±0.009)\end{tabular} &
  0.714 &
  \color{red}{0.823} \\
\textbf{Qwen3-8B} &
  0.473 &
  0.619 &
  \begin{tabular}[c]{@{}r@{}}0.498\\ (±0.014)\end{tabular} &
  \begin{tabular}[c]{@{}r@{}}0.695\\ (±0.006)\end{tabular} &
  \begin{tabular}[c]{@{}r@{}}0.548\\ (±0.007)\end{tabular} &
  \begin{tabular}[c]{@{}r@{}}0.693\\ (±0.007)\end{tabular} &
  \begin{tabular}[c]{@{}r@{}}0.586\\ (±0.007)\end{tabular} &
  \begin{tabular}[c]{@{}r@{}}0.723\\ (±0.007)\end{tabular} &
  0.740 &
  0.760 \\
\textbf{Qwen3-14B-Base} &
  0.508 &
  0.736 &
  \begin{tabular}[c]{@{}r@{}}0.553\\ (±0.008)\end{tabular} &
  \begin{tabular}[c]{@{}r@{}}0.771\\ (±0.010)\end{tabular} &
  \begin{tabular}[c]{@{}r@{}}0.628\\ (±0.004)\end{tabular} &
  \begin{tabular}[c]{@{}r@{}}0.797\\ (±0.007)\end{tabular} &
  \begin{tabular}[c]{@{}r@{}}0.660\\ (±0.007)\end{tabular} &
  \begin{tabular}[c]{@{}r@{}}0.816\\ (±0.006)\end{tabular} &
  - &
  - \\
\textbf{Qwen3-14B} &
  0.488 &
  0.720 &
  \begin{tabular}[c]{@{}r@{}}0.522\\ (±0.007)\end{tabular} &
  \begin{tabular}[c]{@{}r@{}}0.718\\ (±0.011)\end{tabular} &
  \begin{tabular}[c]{@{}r@{}}0.601\\ (±0.008)\end{tabular} &
  \begin{tabular}[c]{@{}r@{}}0.762\\ (±0.009)\end{tabular} &
  \begin{tabular}[c]{@{}r@{}}0.636\\ (±0.005)\end{tabular} &
  \begin{tabular}[c]{@{}r@{}}0.766\\ (±0.005)\end{tabular} &
  - &
  - \\
\textbf{Qwen3-32B} &
  0.514 &
  \color{red}{0.761} &
  \begin{tabular}[c]{@{}r@{}}0.570\\ (±0.011)\end{tabular} &
  \begin{tabular}[c]{@{}r@{}}\color{red}{0.814}\\ (±0.004)\end{tabular} &
  \begin{tabular}[c]{@{}r@{}}0.657\\ (±0.003)\end{tabular} &
  \begin{tabular}[c]{@{}r@{}}\color{red}{0.808}\\ (±0.001)\end{tabular} &
  \begin{tabular}[c]{@{}r@{}}0.695\\ (±0.004)\end{tabular} &
  \begin{tabular}[c]{@{}r@{}}\color{red}{0.821}\\ (±0.005)\end{tabular} &
  - &
  - \\ \midrule
\textbf{Llama-3.1-8B} &
  0.451 &
  0.431 &
  \begin{tabular}[c]{@{}r@{}}0.486\\ (±0.006)\end{tabular} &
  \begin{tabular}[c]{@{}r@{}}0.484\\ (±0.003)\end{tabular} &
  \begin{tabular}[c]{@{}r@{}}0.564\\ (±0.008)\end{tabular} &
  \begin{tabular}[c]{@{}r@{}}0.560\\ (±0.010)\end{tabular} &
  \begin{tabular}[c]{@{}r@{}}0.620\\ (±0.004)\end{tabular} &
  \begin{tabular}[c]{@{}r@{}}0.611\\ (±0.010)\end{tabular} &
  \color{red}{0.802} &
  0.672 \\
\textbf{\begin{tabular}[c]{@{}l@{}}Llama-3.1-8B-\\ Instruct\end{tabular}} &
  0.532 &
  0.514 &
  \begin{tabular}[c]{@{}r@{}}0.589\\ (±0.004)\end{tabular} &
  \begin{tabular}[c]{@{}r@{}}0.588\\ (±0.010)\end{tabular} &
  \begin{tabular}[c]{@{}r@{}}0.643\\ (±0.007)\end{tabular} &
  \begin{tabular}[c]{@{}r@{}}0.626\\ (±0.013)\end{tabular} &
  \begin{tabular}[c]{@{}r@{}}0.679\\ (±0.007)\end{tabular} &
  \begin{tabular}[c]{@{}r@{}}0.648\\ (±0.004)\end{tabular} &
  0.789 &
  0.730 \\
\textbf{Llama-3.2-1B} &
  0.403 &
  0.259 &
  \begin{tabular}[c]{@{}r@{}}0.439\\ (±0.009)\end{tabular} &
  \begin{tabular}[c]{@{}r@{}}0.263\\ (±0.007)\end{tabular} &
  \begin{tabular}[c]{@{}r@{}}0.497\\ (±0.007)\end{tabular} &
  \begin{tabular}[c]{@{}r@{}}0.249\\ (±0.009)\end{tabular} &
  \begin{tabular}[c]{@{}r@{}}0.527\\ (±0.005)\end{tabular} &
  \begin{tabular}[c]{@{}r@{}}0.264\\ (±0.008)\end{tabular} &
  0.747 &
  0.259 \\
\textbf{\begin{tabular}[c]{@{}l@{}}Llama-3.2-1B-\\ Instruct\end{tabular}} &
  0.416 &
  0.300 &
  \begin{tabular}[c]{@{}r@{}}0.468\\ (±0.003)\end{tabular} &
  \begin{tabular}[c]{@{}r@{}}0.347\\ (±0.009)\end{tabular} &
  \begin{tabular}[c]{@{}r@{}}0.523\\ (±0.008)\end{tabular} &
  \begin{tabular}[c]{@{}r@{}}0.339\\ (±0.010)\end{tabular} &
  \begin{tabular}[c]{@{}r@{}}0.542\\ (±0.009)\end{tabular} &
  \begin{tabular}[c]{@{}r@{}}0.330\\ (±0.009)\end{tabular} &
  0.703 &
  0.263 \\
\textbf{Llama-3.2-3B} &
  0.447 &
  0.374 &
  \begin{tabular}[c]{@{}r@{}}0.478\\ (±0.003)\end{tabular} &
  \begin{tabular}[c]{@{}r@{}}0.419\\ (±0.006)\end{tabular} &
  \begin{tabular}[c]{@{}r@{}}0.568\\ (±0.006)\end{tabular} &
  \begin{tabular}[c]{@{}r@{}}0.470\\ (±0.006)\end{tabular} &
  \begin{tabular}[c]{@{}r@{}}0.590\\ (±0.004)\end{tabular} &
  \begin{tabular}[c]{@{}r@{}}0.479\\ (±0.008)\end{tabular} &
  0.761 &
  0.364 \\
\textbf{\begin{tabular}[c]{@{}l@{}}Llama-3.2-3B-\\ Instruct\end{tabular}} &
  0.488 &
  0.488 &
  \begin{tabular}[c]{@{}r@{}}0.464\\ (±0.010)\end{tabular} &
  \begin{tabular}[c]{@{}r@{}}0.518\\ (±0.014)\end{tabular} &
  \begin{tabular}[c]{@{}r@{}}0.540\\ (±0.007)\end{tabular} &
  \begin{tabular}[c]{@{}r@{}}0.492\\ (±0.005)\end{tabular} &
  \begin{tabular}[c]{@{}r@{}}0.583\\ (±0.011)\end{tabular} &
  \begin{tabular}[c]{@{}r@{}}0.481\\ (±0.006)\end{tabular} &
  0.791 &
  0.582 \\ \midrule 
\textbf{Mistral-7B-v0.3} &
  0.479 &
  0.502 &
  \begin{tabular}[c]{@{}r@{}}0.469\\ (±0.005)\end{tabular} &
  \begin{tabular}[c]{@{}r@{}}0.472\\ (±0.016)\end{tabular} &
  \begin{tabular}[c]{@{}r@{}}0.555\\ (±0.012)\end{tabular} &
  \begin{tabular}[c]{@{}r@{}}0.627\\ (±0.009)\end{tabular} &
  \begin{tabular}[c]{@{}r@{}}0.597\\ (±0.006)\end{tabular} &
  \begin{tabular}[c]{@{}r@{}}0.652\\ (±0.006)\end{tabular} &
  0.792 &
  0.477 \\
\textbf{\begin{tabular}[c]{@{}l@{}}Mistral-7B-Instruct-\\ v0.3\end{tabular}} &
  \color{red}{0.662} &
  0.634 &
  \begin{tabular}[c]{@{}r@{}}\color{red}{0.661}\\ (±0.009)\end{tabular} &
  \begin{tabular}[c]{@{}r@{}}0.643\\ (±0.009)\end{tabular} &
  \begin{tabular}[c]{@{}r@{}}\color{red}{0.699}\\ (±0.004)\end{tabular} &
  \begin{tabular}[c]{@{}r@{}}0.635\\ (±0.008)\end{tabular} &
  \begin{tabular}[c]{@{}r@{}}\color{red}{0.712}\\ (±0.009)\end{tabular} &
  \begin{tabular}[c]{@{}r@{}}0.643\\ (±0.008)\end{tabular} &
  0.795 &
  0.711 \\
\textbf{\begin{tabular}[c]{@{}l@{}}Mistral-Nemo-\\ Base-2407 (12B)\end{tabular}} &
  0.466 &
  0.422 &
  \begin{tabular}[c]{@{}r@{}}0.498\\ (±0.008)\end{tabular} &
  \begin{tabular}[c]{@{}r@{}}0.522\\ (±0.015)\end{tabular} &
  \begin{tabular}[c]{@{}r@{}}0.595\\ (±0.005)\end{tabular} &
  \begin{tabular}[c]{@{}r@{}}0.557\\ (±0.006)\end{tabular} &
  \begin{tabular}[c]{@{}r@{}}0.642\\ (±0.005)\end{tabular} &
  \begin{tabular}[c]{@{}r@{}}0.593\\ (±0.007)\end{tabular} &
  - &
  - \\
\textbf{\begin{tabular}[c]{@{}l@{}}Mistral-Nemo-\\ Instruct-2407 (12B)\end{tabular}} &
  0.511 &
  0.627 &
  \begin{tabular}[c]{@{}r@{}}0.546\\ (±0.006)\end{tabular} &
  \begin{tabular}[c]{@{}r@{}}0.643\\ (±0.005)\end{tabular} &
  \begin{tabular}[c]{@{}r@{}}0.636\\ (±0.008)\end{tabular} &
  \begin{tabular}[c]{@{}r@{}}0.658\\ (±0.005)\end{tabular} &
  \begin{tabular}[c]{@{}r@{}}0.676\\ (±0.005)\end{tabular} &
  \begin{tabular}[c]{@{}r@{}}0.654\\ (±0.006)\end{tabular} &
  - &
  - \\
\textbf{\begin{tabular}[c]{@{}l@{}}Mistral-Small-24B-\\ Base-2501\end{tabular}} &
  0.516 &
  0.575 &
  \begin{tabular}[c]{@{}r@{}}0.529\\ (±0.006)\end{tabular} &
  \begin{tabular}[c]{@{}r@{}}0.633\\ (±0.007)\end{tabular} &
  \begin{tabular}[c]{@{}r@{}}0.610\\ (±0.008)\end{tabular} &
  \begin{tabular}[c]{@{}r@{}}0.731\\ (±0.004)\end{tabular} &
  \begin{tabular}[c]{@{}r@{}}0.656\\ (±0.005)\end{tabular} &
  \begin{tabular}[c]{@{}r@{}}0.786\\ (±0.005)\end{tabular} &
  - &
  - \\
\textbf{\begin{tabular}[c]{@{}l@{}}Mistral-Small-24B-\\ Instruct-2501\end{tabular}} &
  0.581 &
  0.729 &
  \begin{tabular}[c]{@{}r@{}}0.638\\ (±0.012)\end{tabular} &
  \begin{tabular}[c]{@{}r@{}}0.758\\ (±0.007)\end{tabular} &
  \begin{tabular}[c]{@{}r@{}}0.668\\ (±0.006)\end{tabular} &
  \begin{tabular}[c]{@{}r@{}}0.757\\ (±0.007)\end{tabular} &
  \begin{tabular}[c]{@{}r@{}}0.702\\ (±0.005)\end{tabular} &
  \begin{tabular}[c]{@{}r@{}}0.804\\ (±0.005)\end{tabular} &
  - &
  - \\ \midrule\midrule
\textbf{total average} &
  \textbf{0.481} &
  \textbf{0.520} &
  \textbf{0.513} &
  \textbf{0.557} &
  \textbf{0.580} &
  \textbf{0.591} &
  \textbf{0.615} &
  \textbf{0.612} &
  \textbf{0.740} &
  \textbf{0.564} \\ \bottomrule
\end{tabular}%
}
\caption{Zero-shot, few-shot, and SFT evaluation results on \bench with the \textbf{detailed instruction}.
SD denotes standard deviation across random seeds or runs. 
Few-shot results are averaged over five random seeds (42, 1234, 3000, 5000, and 7000) and one, five, and ten demonstration examples (1-, 5-, 10-shot). Red text indicates the model with the highest performance in each column.}
\label{tab:total_detail}
\end{table*}

\begin{table*}[htbp]
\centering
\resizebox{0.65\textwidth}{!}{%
\begin{tabular}{@{}lrrrr@{}}
\toprule
\multicolumn{1}{c}{\textbf{models}} &
  \multicolumn{1}{c}{\textbf{zero-shot}} &
  \multicolumn{1}{c}{\textbf{1-shot (±SD)}} &
  \multicolumn{1}{c}{\textbf{5-shot (±SD)}} &
  \multicolumn{1}{c}{\textbf{10-shot (±SD)}} \\ \midrule\midrule
\textbf{GPT-4o mini}  & 0.713 & 0.737 (±0.011) & 0.773 (±0.005) & 0.795 (±0.011) \\
\textbf{GPT-4o}       & 0.782 & 0.785 (±0.005) & 0.800 (±0.012) & 0.824 (±0.006) \\
\textbf{GPT-4.1 mini} & 0.757 & 0.800 (±0.006) & 0.840 (±0.007) & 0.851 (±0.007) \\
\textbf{GPT-4.1}      & \color{red}{0.881} & 0.863 (±0.003) & 0.874 (±0.003) & 0.891 (±0.001) \\ \midrule
\textbf{Haiku 4.5}    & 0.772 & 0.775 (±0.005) & 0.816 (±0.008) & 0.837 (±0.006) \\
\textbf{Sonnet 4.5}   & 0.876 & {\color{red}{0.880}} (±0.006) & {\color{red}{0.875}} (±0.002) & {\color{red}{0.893}} (±0.009) \\ \midrule\midrule
\textbf{total average} &
  \textbf{0.797} &
  \textbf{0.807} &
  \textbf{0.830} &
  \textbf{0.849} \\ \bottomrule
\end{tabular}%
}
\caption{Zero-shot and few-shot results of API models on \bench with the \textbf{definition instruction}.
SD denotes standard deviation across random seeds or runs. 
Few-shot results are averaged over five random seeds (42, 1234, 3000, 5000, and 7000) and one, five, and ten demonstration examples (1-, 5-, 10-shot). Red text indicates the model with the highest performance in each column.}
\label{tab:total_def_api}
\end{table*}

\begin{table*}[htbp]
\centering
\resizebox{0.65\textwidth}{!}{%
\begin{tabular}{@{}lrrrr@{}}
\toprule
\multicolumn{1}{c}{\textbf{models}} &
  \multicolumn{1}{c}{\textbf{zero-shot}} &
  \multicolumn{1}{c}{\textbf{1-shot(±SD)}} &
  \multicolumn{1}{c}{\textbf{5-shot (±SD)}} &
  \multicolumn{1}{c}{\textbf{10-shot (±SD)}} \\ \midrule \midrule
\textbf{GPT-4o mini}   & 0.756          & 0.768 (±0.007)  & 0.774 (±0.005)  & 0.790 (±0.011)  \\
\textbf{GPT-4o}        & 0.863          & 0.847 (±0.006)  & 0.825 (±0.012)  & 0.826 (±0.006)  \\
\textbf{GPT-4.1 mini}  & 0.902          & 0.873 (±0.004)  & 0.870 (±0.007)  & 0.870 (±0.007)  \\
\textbf{GPT-4.1}       & \color{red}{0.936}          & {\color{red}{0.915}} (±0.004)  & {\color{red}{0.901}} (±0.003)  & {\color{red}{0.899}} (±0.001)  \\ \midrule
\textbf{Haiku 4.5}     & 0.865          & 0.870 (±0.004)  & 0.870 (±0.008)  & 0.879 (±0.006)  \\
\textbf{Sonnet 4.5}    & 0.915          & 0.914 (±0.004)  & 0.882 (±0.002)  & 0.825 (±0.009)  \\ \midrule \midrule
\textbf{total average} & \textbf{0.873} & \textbf{0.865} & \textbf{0.854} & \textbf{0.848} \\ \bottomrule
\end{tabular}%
}
\caption{Zero-shot and few-shot results of API models on \bench with the \textbf{detailed instruction}.
SD denotes standard deviation across random seeds or runs. 
Few-shot results are averaged over five random seeds (42, 1234, 3000, 5000, and 7000) and one, five, and ten demonstration examples (1-, 5-, 10-shot). Red text indicates the model with the highest performance in each column.}
\label{tab:total_detail_api}
\end{table*}

\begin{table*}[htbp]
\centering
\resizebox{\textwidth}{!}{%
\begin{tabular}{@{}lrrrrrrr@{}}
\toprule
\multicolumn{1}{c}{\textbf{tasks}} & \multicolumn{1}{c}{\textbf{ARC-Challenge}} & \multicolumn{1}{c}{\textbf{ARC-Easy}} & \multicolumn{1}{c}{\textbf{GSM8K}} & \multicolumn{1}{c}{\textbf{HellaSwag}} & \multicolumn{1}{c}{\textbf{OpenBookQA}} & \multicolumn{1}{c}{\textbf{WinoGrande}} & \multicolumn{1}{c}{\multirow{2}{*}{\textbf{average}}} \\ \cmidrule(r){1-7}
\multicolumn{1}{c}{\textbf{metric}} & \multicolumn{1}{c}{\textbf{acc}} & \multicolumn{1}{c}{\textbf{acc}} & \multicolumn{1}{c}{\textbf{exact\_match}} & \multicolumn{1}{c}{\textbf{acc\_norm}} & \multicolumn{1}{c}{\textbf{acc\_norm}} & \multicolumn{1}{c}{\textbf{acc}} & \multicolumn{1}{c}{} \\ \midrule \midrule
\textbf{gemma-2-2b} & 0.468 (+0.001) & 0.797 (- 0.002) & 0.259 (+0.002) & 0.740 (- 0.001) & 0.422 (- 0.002) & 0.680 (+0.002) & 0.561 (+0.000) \\
\textbf{gemma-2-2b-it} & 0.498 (+0.003) & 0.809 (+0.002) & 0.438 (- 0.014) & 0.716 (- 0.001) & 0.434 (- 0.002) & 0.680 (+0.007) & 0.596 (- 0.001) \\
\textbf{gemma-2-9b} & 0.616 (+0.005) & 0.874 (+0.002) & 0.672 (+0.035) & 0.800 (+0.012) & 0.472 (- 0.006) & 0.741 (+0.009) & 0.696 (+0.010) \\
\textbf{gemma-2-9b-it} & 0.629 (+0.008) & 0.858 (+0.010) & 0.820 (+0.013) & 0.800 (+0.000) & 0.502 (- 0.008) & 0.762 (- 0.010) & 0.729 (+0.002) \\ \midrule
\textbf{Qwen3-0.6B-Base} & 0.326 (- 0.006) & 0.668 (- 0.007) & 0.510 (- 0.005) & 0.517 (+0.002) & 0.340 (+0.000) & 0.594 (- 0.009) & 0.493 (- 0.004) \\
\textbf{Qwen3-0.6B} & 0.317 (- 0.010) & 0.607 (- 0.011) & 0.409 (+0.019) & 0.473 (- 0.009) & 0.318 (- 0.006) & 0.560 (- 0.006) & 0.447 (- 0.004) \\
\textbf{Qwen3-1.7B-Base} & 0.416 (- 0.003) & 0.733 (+0.013) & 0.723 (- 0.034) & 0.665 (+0.000) & 0.392 (- 0.002) & 0.639 (+0.006) & 0.595 (- 0.003) \\
\textbf{Qwen3-1.7B} & 0.403 (+0.009) & 0.726 (+0.017) & 0.692 (- 0.012) & 0.604 (+0.004) & 0.366 (- 0.004) & 0.608 (+0.010) & 0.567 (+0.004) \\
\textbf{Qwen3-4B-Base} & 0.477 (+0.016) & 0.789 (+0.006) & 0.842 (- 0.008) & 0.737 (+0.004) & 0.408 (- 0.010) & 0.706 (- 0.002) & 0.660 (+0.001) \\
\textbf{Qwen3-4B} & 0.507 (+0.017) & 0.804 (+0.012) & 0.840 (+0.026) & 0.684 (+0.003) & 0.402 (- 0.004) & 0.661 (+0.004) & 0.650 (+0.010) \\
\textbf{Qwen3-8B-Base} & 0.532 (+0.002) & 0.817 (+0.005) & 0.850 (- 0.017) & 0.785 (+0.003) & 0.410 (+0.012) & 0.723 (+0.002) & 0.686 (+0.001) \\
\textbf{Qwen3-8B} & 0.555 (+0.008) & 0.835 (+0.005) & 0.880 (+0.002) & 0.750 (+0.006) & 0.416 (+0.010) & 0.677 (+0.017) & 0.686 (+0.008) \\ \midrule
\textbf{Llama-3.1-8B} & 0.513 (+0.015) & 0.814 (+0.014) & 0.502 (+0.027) & 0.790 (+0.007) & 0.450 (+0.016) & 0.739 (+0.004) & 0.635 (+0.014) \\
\textbf{\begin{tabular}[c]{@{}l@{}}Llama-3.1-8B-\\ Instruct\end{tabular}} & 0.515 (+0.015) & 0.819 (+0.006) & 0.784 (- 0.012) & 0.792 (- 0.004) & 0.432 (+0.008) & 0.735 (+0.004) & 0.680 (+0.003) \\
\textbf{Llama-3.2-1B} & 0.317 (- 0.012) & 0.654 (- 0.010) & 0.068 (- 0.011) & 0.636 (- 0.004) & 0.372 (+0.000) & 0.603 (+0.010) & 0.442 (- 0.005) \\
\textbf{\begin{tabular}[c]{@{}l@{}}Llama-3.2-1B-\\ Instruct\end{tabular}} & 0.358 (- 0.007) & 0.683 (+0.002) & 0.334 (- 0.002) & 0.607 (- 0.006) & 0.342 (+0.006) & 0.597 (+0.004) & 0.487 (- 0.001) \\
\textbf{Llama-3.2-3B} & 0.422 (- 0.007) & 0.743 (- 0.003) & 0.262 (+0.025) & 0.736 (+0.007) & 0.428 (- 0.004) & 0.698 (- 0.005) & 0.548 (+0.002) \\
\textbf{\begin{tabular}[c]{@{}l@{}}Llama-3.2-3B-\\ Instruct\end{tabular}} & 0.436 (+0.003) & 0.741 (+0.007) & 0.649 (- 0.008) & 0.705 (- 0.003) & 0.362 (+0.002) & 0.678 (- 0.001) & 0.595 (+0.000) \\ \midrule
\textbf{Mistral-7B-v0.3} & 0.488 (+0.015) & 0.796 (+0.011) & 0.371 (- 0.053) & 0.804 (+0.009) & 0.436 (- 0.006) & 0.736 (- 0.002) & 0.605 (- 0.004) \\
\textbf{\begin{tabular}[c]{@{}l@{}}Mistral-7B-Instruct-\\ v0.3\end{tabular}} & 0.572 (- 0.008) & 0.842 (- 0.007) & 0.498 (- 0.043) & 0.829 (- 0.004) & 0.474 (- 0.016) & 0.744 (- 0.014) & 0.660 (- 0.015) \\ \midrule 
 &  &  &  &  &  & \multicolumn{1}{c}{\textbf{total average}} & \textbf{0.601 (+0.001)} \\ \cmidrule(l){7-8} 
\end{tabular}%
}
\vspace*{-0.5\baselineskip}
\caption{General benchmark performance after supervised fine-tuning (SFT) on \bench with the \textbf{definition instruction}.
Here, \textit{acc} denotes accuracy, \textit{acc\_norm} is normalized accuracy, and \textit{exact\_match} requires an exact string match with the reference answer.  
For each model and task, the main value reports the performance of the pre-SFT model, while the value in parentheses indicates the change after applying SFT.}
\label{tab:general_sft_def}
\end{table*}

\begin{table*}[htbp]
\centering
\resizebox{\textwidth}{!}{%
\begin{tabular}{@{}lrrrrrrr@{}}
\toprule
\multicolumn{1}{c}{\textbf{tasks}} & \multicolumn{1}{c}{\textbf{ARC-Challenge}} & \multicolumn{1}{c}{\textbf{ARC-Easy}} & \multicolumn{1}{c}{\textbf{GSM8K}} & \multicolumn{1}{c}{\textbf{HellaSwag}} & \multicolumn{1}{c}{\textbf{OpenBookQA}} & \multicolumn{1}{c}{\textbf{WinoGrande}} & \multicolumn{1}{c}{\multirow{2}{*}{\textbf{average}}} \\ \cmidrule(r){1-7}
\multicolumn{1}{c}{\textbf{metric}} & \multicolumn{1}{c}{\textbf{acc}} & \multicolumn{1}{c}{\textbf{acc}} & \multicolumn{1}{c}{\textbf{exact\_match}} & \multicolumn{1}{c}{\textbf{acc\_norm}} & \multicolumn{1}{c}{\textbf{acc\_norm}} & \multicolumn{1}{c}{\textbf{acc}} & \multicolumn{1}{c}{} \\ \midrule \midrule
\textbf{gemma-2-2b} & 0.468 (+0.000) & 0.797 (+0.000) & 0.259 (+0.000) & 0.740 (+0.000) & 0.422 (+0.000) & 0.680 (+0.000) & 0.561 (+0.000) \\
\textbf{gemma-2-2b-it} & 0.498 (- 0.001) & 0.809 (- 0.002) & 0.438 (+0.000) & 0.716 (- 0.001) & 0.434 (+0.002) & 0.680 (+0.002) & 0.596 (+0.000) \\
\textbf{gemma-2-9b} & 0.616 (+0.004) & 0.874 (+0.001) & 0.672 (+0.035) & 0.800 (+0.013) & 0.472 (- 0.008) & 0.741 (+0.006) & 0.696 (+0.009) \\
\textbf{gemma-2-9b-it} & 0.629 (+0.006) & 0.858 (+0.011) & 0.820 (+0.005) & 0.800 (+0.001) & 0.502 (+0.002) & 0.762 (- 0.004) & 0.729 (+0.004) \\ \midrule
\textbf{Qwen3-0.6B-Base} & 0.326 (+0.000) & 0.668 (+0.000) & 0.510 (+0.000) & 0.517 (+0.000) & 0.340 (+0.000) & 0.594 (+0.000) & 0.493 (+0.000) \\
\textbf{Qwen3-0.6B} & 0.317 (- 0.009) & 0.607 (+0.016) & 0.409 (+0.014) & 0.473 (- 0.012) & 0.318 (- 0.002) & 0.560 (- 0.001) & 0.447 (+0.001) \\
\textbf{Qwen3-1.7B-Base} & 0.416 (+0.002) & 0.733 (+0.003) & 0.723 (- 0.035) & 0.665 (+0.000) & 0.392 (- 0.004) & 0.639 (+0.002) & 0.595 (- 0.005) \\
\textbf{Qwen3-1.7B} & 0.403 (+0.003) & 0.726 (+0.021) & 0.692 (- 0.024) & 0.604 (+0.003) & 0.366 (+0.000) & 0.608 (+0.009) & 0.567 (+0.002) \\
\textbf{Qwen3-4B-Base} & 0.477 (+0.026) & 0.789 (+0.009) & 0.842 (- 0.036) & 0.737 (+0.005) & 0.408 (- 0.004) & 0.706 (+0.004) & 0.660 (+0.001) \\
\textbf{Qwen3-4B} & 0.507 (+0.008) & 0.804 (+0.009) & 0.840 (+0.008) & 0.684 (+0.008) & 0.402 (+0.006) & 0.661 (+0.006) & 0.650 (+0.008) \\
\textbf{Qwen3-8B-Base} & 0.532 (+0.024) & 0.817 (+0.011) & 0.850 (- 0.040) & 0.785 (+0.004) & 0.410 (+0.014) & 0.723 (- 0.002) & 0.686 (+0.002) \\
\textbf{Qwen3-8B} & 0.555 (+0.002) & 0.835 (+0.000) & 0.880 (- 0.002) & 0.750 (+0.005) & 0.416 (+0.006) & 0.677 (+0.016) & 0.686 (+0.005) \\ \midrule
\textbf{Llama-3.1-8B} & 0.513 (+0.003) & 0.814 (+0.012) & 0.502 (- 0.001) & 0.790 (+0.010) & 0.450 (+0.018) & 0.739 (+0.001) & 0.635 (+0.007) \\
\textbf{\begin{tabular}[c]{@{}l@{}}Llama-3.1-8B-\\ Instruct\end{tabular}} & 0.515 (+0.009) & 0.819 (+0.003) & 0.784 (- 0.013) & 0.792 (- 0.006) & 0.432 (+0.006) & 0.735 (+0.009) & 0.680 (+0.001) \\
\textbf{Llama-3.2-1B} & 0.317 (- 0.009) & 0.654 (- 0.005) & 0.068 (- 0.024) & 0.636 (- 0.008) & 0.372 (- 0.004) & 0.603 (+0.017) & 0.442 (- 0.006) \\
\textbf{\begin{tabular}[c]{@{}l@{}}Llama-3.2-1B-\\ Instruct\end{tabular}} & 0.358 (- 0.004) & 0.683 (+0.002) & 0.334 (- 0.008) & 0.607 (- 0.007) & 0.342 (+0.008) & 0.597 (+0.000) & 0.487 (- 0.002) \\
\textbf{Llama-3.2-3B} & 0.422 (- 0.005) & 0.743 (- 0.003) & 0.262 (+0.014) & 0.736 (+0.004) & 0.428 (- 0.006) & 0.698 (+0.004) & 0.548 (+0.001) \\
\textbf{\begin{tabular}[c]{@{}l@{}}Llama-3.2-3B-\\ Instruct\end{tabular}} & 0.436 (+0.001) & 0.741 (+0.006) & 0.649 (+0.001) & 0.705 (- 0.001) & 0.362 (+0.000) & 0.678 (+0.000) & 0.595 (+0.001) \\ \midrule
\textbf{Mistral-7B-v0.3} & 0.488 (+0.013) & 0.796 (+0.011) & 0.371 (- 0.094) & 0.804 (+0.004) & 0.436 (+0.008) & 0.736 (+0.005) & 0.605 (- 0.009) \\
\textbf{\begin{tabular}[c]{@{}l@{}}Mistral-7B-Instruct-\\ v0.3\end{tabular}} & 0.572 (- 0.009) & 0.842 (- 0.004) & 0.498 (- 0.057) & 0.829 (- 0.003) & 0.474 (- 0.006) & 0.744 (- 0.014) & 0.660 (- 0.016) \\ \midrule
 &  &  &  &  &  & \multicolumn{1}{c}{\textbf{total average}} & \textbf{0.601 (+0.000)} \\ \cmidrule(l){7-8} 
\end{tabular}%
}
\vspace*{-0.5\baselineskip}
\caption{General benchmark performance after supervised fine-tuning (SFT) on \bench with the \textbf{detailed instruction}.
Here, \textit{acc} denotes accuracy, \textit{acc\_norm} is normalized accuracy, and \textit{exact\_match} requires an exact string match with the reference answer.  
For each model and task, the main value reports the performance of the pre-SFT model, while the value in parentheses indicates the change after applying SFT.}
\label{tab:general_sft_detail}
\end{table*}

\begin{table*}[htbp]
\centering
\resizebox{\textwidth}{!}{%
\begin{tabular}{@{}cccr|rrr|rrrr@{}}
\toprule
\multirow{2}{*}{\textbf{Model}} & \multirow{2}{*}{\textbf{\begin{tabular}[c]{@{}c@{}}Training\\ Setting\end{tabular}}} & \multirow{2}{*}{\textbf{N Shot}} & \multicolumn{1}{c|}{\multirow{2}{*}{\textbf{\begin{tabular}[c]{@{}c@{}}Error \\ Rate\\ (1-acc)\end{tabular}}}} & \multicolumn{3}{c|}{\textbf{Incorrect Choice Distribution}} & \multicolumn{4}{c}{\textbf{Local Negation Confusion Rate}} \\ \cmidrule(l){5-11} 
 &  &  & \multicolumn{1}{c|}{} & \multicolumn{1}{c}{\textbf{\begin{tabular}[c]{@{}c@{}}Local\\ Negation\\ (\%)\end{tabular}}} & \multicolumn{1}{c}{\textbf{\begin{tabular}[c]{@{}c@{}}Contra-\\ diction\\ (\%)\end{tabular}}} & \multicolumn{1}{c|}{\textbf{\begin{tabular}[c]{@{}c@{}}Para-\\ phrase\\ (\%)\end{tabular}}} & \multicolumn{1}{c}{\textbf{\begin{tabular}[c]{@{}c@{}}Relative\\ Clause\\ (\%)\end{tabular}}} & \multicolumn{1}{c}{\textbf{\begin{tabular}[c]{@{}c@{}}Participle\\ Clause\\ (\%)\end{tabular}}} & \multicolumn{1}{c}{\textbf{\begin{tabular}[c]{@{}c@{}}Compound\\ Sentence\\ (\%)\end{tabular}}} & \multicolumn{1}{c}{\textbf{\begin{tabular}[c]{@{}c@{}}Adverbial\\ Clause\\ (\%)\end{tabular}}} \\ \midrule
\multirow{5}{*}{\textbf{gemma-2-2b}} & \multirow{4}{*}{\textbf{baseline}} & \textbf{zero-shot} & 0.564 & 74.96 & 19.41 & 5.63 & 25.64 & 32.47 & 63.27 & 53.87 \\
 &  & \textbf{1-shot} & 0.523 & 76.52 & 19.55 & 3.94 & 28.85 & 30.84 & 58.84 & 47.42 \\
 &  & \textbf{5-shot} & 0.420 & 81.47 & 17.01 & 1.51 & 21.47 & 25.00 & 51.02 & 44.19 \\
 &  & \textbf{10-shot} & 0.399 & 81.71 & 16.30 & 1.99 & 21.47 & 21.43 & 46.94 & 45.16 \\ \cmidrule(l){2-11} 
 & \textbf{after SFT} & \textbf{zero-shot} & 0.219 & 82.25 & 17.03 & 0.72 & 11.86 & 12.99 & 26.19 & 23.55 \\ \midrule
\multirow{5}{*}{\textbf{gemma-2-2b-it}} & \multirow{4}{*}{\textbf{baseline}} & \textbf{zero-shot} & 0.485 & 75.29 & 22.09 & 2.62 & 22.12 & 23.38 & 48.64 & 56.77 \\
 &  & \textbf{1-shot} & 0.480 & 77.36 & 21.32 & 1.32 & 29.17 & 24.68 & 50.00 & 49.68 \\
 &  & \textbf{5-shot} & 0.427 & 78.07 & 20.07 & 1.86 & 23.08 & 23.05 & 46.94 & 44.84 \\
 &  & \textbf{10-shot} & 0.408 & 77.24 & 21.60 & 1.17 & 21.47 & 20.78 & 41.50 & 46.45 \\ \cmidrule(l){2-11}
 & \textbf{after SFT} & \textbf{zero-shot} & 0.259 & 82.87 & 15.60 & 1.53 & 11.86 & 15.58 & 28.57 & 32.90 \\ \midrule
\multirow{5}{*}{\textbf{gemma-2-9b}} & \multirow{4}{*}{\textbf{baseline}} & \textbf{zero-shot} & 0.530 & 75.30 & 19.61 & 5.09 & 24.36 & 31.49 & 61.56 & 48.06 \\
 &  & \textbf{1-shot} & 0.500 & 78.73 & 17.62 & 3.65 & 26.28 & 29.87 & 58.84 & 48.06 \\
 &  & \textbf{5-shot} & 0.431 & 83.79 & 14.00 & 2.21 & 22.44 & 27.27 & 50.00 & 49.68 \\
 &  & \textbf{10-shot} & 0.402 & 84.22 & 14.60 & 1.18 & 21.15 & 27.27 & 44.22 & 47.42 \\ \cmidrule(l){2-11}
 & \textbf{after SFT} & \textbf{zero-shot} & 0.229 & 85.47 & 12.11 & 2.42 & 10.26 & 13.96 & 22.45 & 34.19 \\ \midrule
\multirow{5}{*}{\textbf{gemma-2-9b-it}} & \multirow{4}{*}{\textbf{baseline}} & \textbf{zero-shot} & 0.493 & 76.01 & 22.06 & 1.93 & 24.68 & 27.92 & 52.72 & 49.68 \\
 &  & \textbf{1-shot} & 0.439 & 77.26 & 21.66 & 1.08 & 24.36 & 24.68 & 46.94 & 44.52 \\
 &  & \textbf{5-shot} & 0.352 & 78.38 & 20.72 & 0.90 & 17.31 & 17.86 & 31.97 & 46.77 \\
 &  & \textbf{10-shot} & 0.302 & 77.95 & 21.00 & 1.05 & 14.10 & 16.23 & 26.19 & 40.65 \\
 & \textbf{after SFT} & \textbf{zero-shot} & 0.233 & 87.41 & 10.88 & 1.70 & 9.29 & 14.94 & 26.19 & 33.87 \\ \bottomrule
\end{tabular}%
}
\caption{Error rates, incorrect choice distributions, and local negation confusion rates for the \textbf{Gemma2} family under zero-shot, few-shot, and SFT conditions, evaluated in the \textbf{completion-based setting} using \textbf{definition instruction.}}
\label{tab:neg_gemma_def_completion}
\end{table*}

\begin{table*}[htbp]
\centering
\resizebox{\textwidth}{!}{%
\begin{tabular}{@{}cccr|rrr|rrrr@{}}
\toprule
\multirow{2}{*}{\textbf{Model}} & \multirow{2}{*}{\textbf{\begin{tabular}[c]{@{}c@{}}Training\\ Setting\end{tabular}}} & \multirow{2}{*}{\textbf{N Shot}} & \multicolumn{1}{c|}{\multirow{2}{*}{\textbf{\begin{tabular}[c]{@{}c@{}}Error \\ Rate\\ (1-acc)\end{tabular}}}} & \multicolumn{3}{c|}{\textbf{Incorrect Choice Distribution}} & \multicolumn{4}{c}{\textbf{Local Negation Confusion Rate}} \\ \cmidrule(l){5-11} 
 &  &  & \multicolumn{1}{c|}{} & \multicolumn{1}{c}{\textbf{\begin{tabular}[c]{@{}c@{}}Local\\ Negation\\ (\%)\end{tabular}}} & \multicolumn{1}{c}{\textbf{\begin{tabular}[c]{@{}c@{}}Contra-\\ diction\\ (\%)\end{tabular}}} & \multicolumn{1}{c|}{\textbf{\begin{tabular}[c]{@{}c@{}}Para-\\ phrase\\ (\%)\end{tabular}}} & \multicolumn{1}{c}{\textbf{\begin{tabular}[c]{@{}c@{}}Relative\\ Clause\\ (\%)\end{tabular}}} & \multicolumn{1}{c}{\textbf{\begin{tabular}[c]{@{}c@{}}Participle\\ Clause\\ (\%)\end{tabular}}} & \multicolumn{1}{c}{\textbf{\begin{tabular}[c]{@{}c@{}}Compound\\ Sentence\\ (\%)\end{tabular}}} & \multicolumn{1}{c}{\textbf{\begin{tabular}[c]{@{}c@{}}Adverbial\\ Clause\\ (\%)\end{tabular}}} \\ \midrule
\multirow{5}{*}{\textbf{gemma-2-2b}} & \multirow{4}{*}{\textbf{baseline}} & \textbf{zero-shot} & 0.765 & 33.20 & 26.66 & 40.15 & 22.12 & 29.55 & 29.25 & 23.87 \\
 &  & \textbf{1-shot} & 0.668 & 40.38 & 13.54 & 46.08 & 21.79 & 26.62 & 34.01 & 29.03 \\
 &  & \textbf{5-shot} & 0.623 & 53.69 & 12.47 & 33.84 & 27.88 & 34.42 & 40.48 & 35.48 \\
 &  & \textbf{10-shot} & 0.599 & 58.81 & 10.86 & 30.33 & 31.09 & 35.39 & 43.20 & 35.81 \\ \cmidrule(l){2-11} 
 & \textbf{after SFT} & \textbf{zero-shot} & 0.740 & 32.48 & 34.62 & 32.90 & 22.76 & 27.92 & 25.51 & 22.90 \\ \midrule
\multirow{5}{*}{\textbf{gemma-2-2b-it}} & \multirow{4}{*}{\textbf{baseline}} & \textbf{zero-shot} & 0.496 & 74.88 & 8.80 & 16.32 & 37.18 & 26.30 & 66.67 & 24.19 \\
 &  & \textbf{1-shot} & 0.496 & 70.93 & 11.34 & 17.73 & 30.45 & 30.84 & 55.78 & 29.03 \\
 &  & \textbf{5-shot} & 0.473 & 67.67 & 12.56 & 19.77 & 31.09 & 33.12 & 43.88 & 24.52 \\
 &  & \textbf{10-shot} & 0.467 & 69.10 & 12.39 & 18.51 & 34.62 & 31.82 & 42.18 & 24.84 \\ \cmidrule(l){2-11}
 & \textbf{after SFT} & \textbf{zero-shot} & 0.457 & 68.75 & 6.77 & 24.48 & 29.17 & 30.84 & 50.68 & 19.68 \\ \midrule
\multirow{5}{*}{\textbf{gemma-2-9b}} & \multirow{4}{*}{\textbf{baseline}} & \textbf{zero-shot} & 0.532 & 62.89 & 7.60 & 29.51 & 36.22 & 29.87 & 57.82 & 15.16 \\
 &  & \textbf{1-shot} & 0.472 & 62.35 & 4.71 & 32.94 & 34.94 & 24.35 & 48.30 & 14.52 \\
 &  & \textbf{5-shot} & 0.402 & 71.79 & 6.11 & 22.09 & 33.01 & 23.70 & 45.92 & 17.10 \\
 &  & \textbf{10-shot} & 0.378 & 71.43 & 4.62 & 23.95 & 33.33 & 20.78 & 40.82 & 16.77 \\ \cmidrule(l){2-11}
 & \textbf{after SFT} & \textbf{zero-shot} & 0.249 & 64.01 & 16.24 & 19.75 & 21.15 & 11.36 & 23.81 & 9.68 \\ \midrule
\multirow{5}{*}{\textbf{gemma-2-9b-it}} & \multirow{4}{*}{\textbf{baseline}} & \textbf{zero-shot} & 0.447 & 73.00 & 22.02 & 4.97 & 39.42 & 22.40 & 51.70 & 21.61 \\
 &  & \textbf{1-shot} & 0.370 & 80.69 & 13.95 & 5.36 & 35.26 & 22.73 & 42.86 & 22.58 \\
 &  & \textbf{5-shot} & 0.286 & 83.66 & 11.63 & 4.71 & 28.85 & 17.53 & 31.63 & 20.97 \\
 &  & \textbf{10-shot} & 0.272 & 82.51 & 12.54 & 4.96 & 28.85 & 17.53 & 28.23 & 18.06 \\ \cmidrule(l){2-11}
 & \textbf{after SFT} & \textbf{zero-shot} & 0.186 & 73.62 & 21.70 & 4.68 & 11.86 & 10.71 & 25.51 & 9.03 \\ \bottomrule
\end{tabular}%
}
\caption{Error rates, incorrect choice distributions, and local negation confusion rates for the \textbf{Gemma2} family under zero-shot, few-shot, and SFT conditions, evaluated in the \textbf{option-selection setting} using \textbf{definition instruction.}}
\label{tab:neg_gemma_def_option}
\end{table*}

\begin{table*}[htbp]
\centering
\resizebox{\textwidth}{!}{%
\begin{tabular}{@{}cccr|rrr|rrrr@{}}
\toprule
\multirow{2}{*}{\textbf{Model}} & \multirow{2}{*}{\textbf{\begin{tabular}[c]{@{}c@{}}Training\\ Setting\end{tabular}}} & \multirow{2}{*}{\textbf{N Shot}} & \multicolumn{1}{c|}{\multirow{2}{*}{\textbf{\begin{tabular}[c]{@{}c@{}}Error \\ Rate\\ (1-acc)\end{tabular}}}} & \multicolumn{3}{c|}{\textbf{Incorrect Choice Distribution}} & \multicolumn{4}{c}{\textbf{Local Negation Confusion Rate}} \\ \cmidrule(l){5-11} 
 &  &  & \multicolumn{1}{c|}{} & \multicolumn{1}{c}{\textbf{\begin{tabular}[c]{@{}c@{}}Local\\ Negation\\ (\%)\end{tabular}}} & \multicolumn{1}{c}{\textbf{\begin{tabular}[c]{@{}c@{}}Contra-\\ diction\\ (\%)\end{tabular}}} & \multicolumn{1}{c|}{\textbf{\begin{tabular}[c]{@{}c@{}}Para-\\ phrase\\ (\%)\end{tabular}}} & \multicolumn{1}{c}{\textbf{\begin{tabular}[c]{@{}c@{}}Relative\\ Clause\\ (\%)\end{tabular}}} & \multicolumn{1}{c}{\textbf{\begin{tabular}[c]{@{}c@{}}Participle\\ Clause\\ (\%)\end{tabular}}} & \multicolumn{1}{c}{\textbf{\begin{tabular}[c]{@{}c@{}}Compound\\ Sentence\\ (\%)\end{tabular}}} & \multicolumn{1}{c}{\textbf{\begin{tabular}[c]{@{}c@{}}Adverbial\\ Clause\\ (\%)\end{tabular}}} \\ \midrule
\multirow{5}{*}{\textbf{gemma-2-2b}} & \multirow{4}{*}{\textbf{baseline}} & \textbf{zero-shot} & 0.562 & 74.89 & 19.04 & 6.06 & 26.60 & 34.74 & 60.88 & 52.26 \\
 &  & \textbf{1-shot} & 0.507 & 76.37 & 19.41 & 4.23 & 26.60 & 30.84 & 56.46 & 46.45 \\
 &  & \textbf{5-shot} & 0.425 & 82.09 & 16.04 & 1.87 & 20.83 & 25.97 & 52.72 & 45.16 \\
 &  & \textbf{10-shot} & 0.392 & 81.98 & 15.99 & 2.02 & 20.19 & 22.40 & 46.26 & 44.19 \\ \cmidrule(l){2-11} 
 & \textbf{after SFT} & \textbf{zero-shot} & 0.232 & 82.59 & 16.72 & 0.68 & 11.22 & 13.64 & 27.21 & 27.42 \\ \midrule
\multirow{5}{*}{\textbf{gemma-2-2b-it}} & \multirow{4}{*}{\textbf{baseline}} & \textbf{zero-shot} & 0.495 & 78.37 & 18.75 & 2.88 & 24.68 & 27.60 & 54.76 & 53.55 \\
 &  & \textbf{1-shot} & 0.493 & 79.74 & 18.81 & 1.45 & 29.17 & 28.25 & 52.38 & 52.90 \\
 &  & \textbf{5-shot} & 0.450 & 80.81 & 17.61 & 1.58 & 25.64 & 24.35 & 51.70 & 49.03 \\
 &  & \textbf{10-shot} & 0.422 & 80.26 & 18.42 & 1.32 & 22.76 & 22.40 & 46.60 & 48.39 \\ \cmidrule(l){2-11} 
 & \textbf{after SFT} & \textbf{zero-shot} & 0.267 & 81.90 & 16.62 & 1.48 & 12.50 & 14.61 & 32.65 & 30.97 \\ \midrule
\multirow{5}{*}{\textbf{gemma-2-9b}} & \multirow{4}{*}{\textbf{baseline}} & \textbf{zero-shot} & 0.531 & 74.63 & 20.60 & 4.78 & 25.00 & 33.77 & 62.59 & 43.23 \\
 &  & \textbf{1-shot} & 0.487 & 78.50 & 17.59 & 3.91 & 25.96 & 30.52 & 59.18 & 42.90 \\
 &  & \textbf{5-shot} & 0.425 & 83.21 & 14.37 & 2.43 & 21.47 & 25.65 & 51.36 & 48.06 \\
 &  & \textbf{10-shot} & 0.393 & 83.23 & 15.35 & 1.41 & 20.83 & 25.97 & 43.54 & 44.84 \\ \cmidrule(l){2-11} 
 & \textbf{after SFT} & \textbf{zero-shot} & 0.228 & 86.06 & 11.85 & 2.09 & 8.97 & 14.61 & 23.81 & 33.55 \\ \midrule
\multirow{5}{*}{\textbf{gemma-2-9b-it}} & \multirow{4}{*}{\textbf{baseline}} & \textbf{zero-shot} & 0.458 & 76.95 & 21.14 & 1.91 & 18.59 & 27.60 & 56.80 & 43.23 \\
 &  & \textbf{1-shot} & 0.401 & 77.62 & 21.58 & 0.79 & 20.51 & 20.78 & 45.24 & 42.26 \\
 &  & \textbf{5-shot} & 0.345 & 78.85 & 20.23 & 0.92 & 16.35 & 17.53 & 33.67 & 44.84 \\
 &  & \textbf{10-shot} & 0.308 & 78.87 & 19.85 & 1.29 & 14.10 & 17.21 & 26.53 & 42.26 \\ \cmidrule(l){2-11} 
 & \textbf{after SFT} & \textbf{zero-shot} & 0.236 & 85.91 & 11.41 & 2.68 & 10.26 & 13.64 & 26.87 & 33.23 \\ \bottomrule
\end{tabular}%
}
\caption{Error rates, incorrect choice distributions, and local negation confusion rates for the \textbf{Gemma2} family under zero-shot, few-shot, and SFT conditions, evaluated in the \textbf{completion-based setting} using \textbf{detailed instruction.}}
\label{tab:neg_gemma_detail_completion}
\end{table*}

\begin{table*}[htbp]
\centering
\resizebox{\textwidth}{!}{%
\begin{tabular}{@{}cccr|rrr|rrrr@{}}
\toprule
\multirow{2}{*}{\textbf{Model}} & \multirow{2}{*}{\textbf{\begin{tabular}[c]{@{}c@{}}Training\\ Setting\end{tabular}}} & \multirow{2}{*}{\textbf{N Shot}} & \multicolumn{1}{c|}{\multirow{2}{*}{\textbf{\begin{tabular}[c]{@{}c@{}}Error \\ Rate\\ (1-acc)\end{tabular}}}} & \multicolumn{3}{c|}{\textbf{Incorrect Choice Distribution}} & \multicolumn{4}{c}{\textbf{Local Negation Confusion Rate}} \\ \cmidrule(l){5-11} 
 &  &  & \multicolumn{1}{c|}{} & \multicolumn{1}{c}{\textbf{\begin{tabular}[c]{@{}c@{}}Local\\ Negation\\ (\%)\end{tabular}}} & \multicolumn{1}{c}{\textbf{\begin{tabular}[c]{@{}c@{}}Contra-\\ diction\\ (\%)\end{tabular}}} & \multicolumn{1}{c|}{\textbf{\begin{tabular}[c]{@{}c@{}}Para-\\ phrase\\ (\%)\end{tabular}}} & \multicolumn{1}{c}{\textbf{\begin{tabular}[c]{@{}c@{}}Relative\\ Clause\\ (\%)\end{tabular}}} & \multicolumn{1}{c}{\textbf{\begin{tabular}[c]{@{}c@{}}Participle\\ Clause\\ (\%)\end{tabular}}} & \multicolumn{1}{c}{\textbf{\begin{tabular}[c]{@{}c@{}}Compound\\ Sentence\\ (\%)\end{tabular}}} & \multicolumn{1}{c}{\textbf{\begin{tabular}[c]{@{}c@{}}Adverbial\\ Clause\\ (\%)\end{tabular}}} \\ \midrule
\multirow{5}{*}{\textbf{gemma-2-2b}} & \multirow{4}{*}{\textbf{baseline}} & \textbf{zero-shot} & 0.724 & 35.93 & 29.46 & 34.61 & 23.72 & 31.49 & 29.25 & 22.90 \\
 &  & \textbf{1-shot} & 0.660 & 42.91 & 15.02 & 42.07 & 22.76 & 28.90 & 35.37 & 30.00 \\
 &  & \textbf{5-shot} & 0.629 & 53.47 & 12.48 & 34.05 & 28.21 & 34.74 & 40.82 & 35.16 \\
 &  & \textbf{10-shot} & 0.612 & 57.64 & 11.53 & 30.83 & 32.05 & 34.42 & 41.84 & 37.42 \\ \cmidrule(l){2-11} 
 & \textbf{after SFT} & \textbf{zero-shot} & 0.741 & 32.55 & 34.58 & 32.87 & 22.76 & 27.92 & 25.51 & 23.23 \\ \midrule
\multirow{5}{*}{\textbf{gemma-2-2b-it}} & \multirow{4}{*}{\textbf{baseline}} & \textbf{zero-shot} & 0.531 & 82.06 & 7.47 & 10.46 & 42.95 & 38.96 & 67.69 & 30.97 \\
 &  & \textbf{1-shot} & 0.539 & 70.84 & 8.84 & 20.32 & 32.37 & 31.82 & 63.27 & 30.97 \\
 &  & \textbf{5-shot} & 0.508 & 64.06 & 12.81 & 23.12 & 30.13 & 32.47 & 45.92 & 26.13 \\
 &  & \textbf{10-shot} & 0.504 & 61.64 & 12.11 & 26.26 & 31.41 & 29.22 & 43.20 & 24.84 \\ \cmidrule(l){2-11} 
 & \textbf{after SFT} & \textbf{zero-shot} & 0.337 & 72.00 & 10.12 & 17.88 & 21.15 & 25.65 & 35.71 & 18.06 \\  \midrule
\multirow{5}{*}{\textbf{gemma-2-9b}} & \multirow{4}{*}{\textbf{baseline}} & \textbf{zero-shot} & 0.534 & 73.70 & 8.47 & 17.83 & 42.63 & 32.47 & 66.33 & 21.94 \\
 &  & \textbf{1-shot} & 0.476 & 64.83 & 6.50 & 28.67 & 36.22 & 25.32 & 50.00 & 16.45 \\
 &  & \textbf{5-shot} & 0.422 & 70.11 & 5.45 & 24.44 & 33.65 & 22.73 & 48.30 & 18.06 \\
 &  & \textbf{10-shot} & 0.403 & 69.88 & 3.94 & 26.18 & 34.62 & 20.45 & 44.22 & 17.42 \\ \cmidrule(l){2-11} 
 & \textbf{after SFT} & \textbf{zero-shot} & 0.203 & 75.00 & 11.72 & 13.28 & 22.12 & 12.01 & 16.33 & 12.26 \\  \midrule
\multirow{5}{*}{\textbf{gemma-2-9b-it}} & \multirow{4}{*}{\textbf{baseline}} & \textbf{zero-shot} & 0.431 & 83.27 & 11.58 & 5.15 & 35.90 & 23.70 & 63.61 & 26.13 \\
 &  & \textbf{1-shot} & 0.361 & 83.30 & 10.77 & 5.93 & 35.58 & 21.75 & 43.54 & 23.55 \\
 &  & \textbf{5-shot} & 0.305 & 84.11 & 10.42 & 5.47 & 30.45 & 21.43 & 32.31 & 21.61 \\
 &  & \textbf{10-shot} & 0.284 & 82.12 & 12.85 & 5.03 & 30.13 & 18.51 & 28.57 & 19.03 \\ \cmidrule(l){2-11} 
 & \textbf{after SFT} & \textbf{zero-shot} & 0.195 & 69.92 & 17.07 & 13.01 & 16.67 & 13.96 & 12.59 & 12.90 \\ \bottomrule
\end{tabular}%
}
\caption{Error rates, incorrect choice distributions, and local negation confusion rates for the \textbf{Gemma2} family under zero-shot, few-shot, and SFT conditions, evaluated in the \textbf{option-selection setting} using \textbf{detailed instruction.}}
\label{tab:neg_gemma_detail_option}
\end{table*}

\begin{table*}[htbp]
\centering
\resizebox{\textwidth}{!}{%
\begin{tabular}{@{}cccr|rrr|rrrr@{}}
\toprule
\multirow{2}{*}{\textbf{Model}} & \multirow{2}{*}{\textbf{\begin{tabular}[c]{@{}c@{}}Training\\ Setting\end{tabular}}} & \multirow{2}{*}{\textbf{N Shot}} & \multicolumn{1}{c|}{\multirow{2}{*}{\textbf{\begin{tabular}[c]{@{}c@{}}Error \\ Rate\\ (1-acc)\end{tabular}}}} & \multicolumn{3}{c|}{\textbf{Incorrect Choice Distribution}} & \multicolumn{4}{c}{\textbf{Local Negation Confusion Rate}} \\ \cmidrule(l){5-11} 
 &  &  & \multicolumn{1}{c|}{} & \multicolumn{1}{c}{\textbf{\begin{tabular}[c]{@{}c@{}}Local\\ Negation\\ (\%)\end{tabular}}} & \multicolumn{1}{c}{\textbf{\begin{tabular}[c]{@{}c@{}}Contra-\\ diction\\ (\%)\end{tabular}}} & \multicolumn{1}{c|}{\textbf{\begin{tabular}[c]{@{}c@{}}Para-\\ phrase\\ (\%)\end{tabular}}} & \multicolumn{1}{c}{\textbf{\begin{tabular}[c]{@{}c@{}}Relative\\ Clause\\ (\%)\end{tabular}}} & \multicolumn{1}{c}{\textbf{\begin{tabular}[c]{@{}c@{}}Participle\\ Clause\\ (\%)\end{tabular}}} & \multicolumn{1}{c}{\textbf{\begin{tabular}[c]{@{}c@{}}Compound\\ Sentence\\ (\%)\end{tabular}}} & \multicolumn{1}{c}{\textbf{\begin{tabular}[c]{@{}c@{}}Adverbial\\ Clause\\ (\%)\end{tabular}}} \\ \midrule
\multirow{5}{*}{\textbf{Qwen3-0.6B-Base}} & \multirow{4}{*}{\textbf{baseline}} & \textbf{zero-shot} & 0.547 & 72.32 & 23.48 & 4.20 & 29.49 & 34.74 & 50.68 & 48.71 \\
 &  & \textbf{1-shot} & 0.487 & 74.27 & 22.15 & 3.58 & 31.09 & 28.90 & 46.60 & 42.90 \\
 &  & \textbf{5-shot} & 0.412 & 77.07 & 19.46 & 3.47 & 23.40 & 21.75 & 43.54 & 42.58 \\
 &  & \textbf{10-shot} & 0.390 & 79.27 & 18.90 & 1.83 & 22.76 & 22.08 & 40.82 & 42.26 \\ \cmidrule(l){2-11} 
 & \textbf{after SFT} & \textbf{zero-shot} & 0.292 & 83.97 & 15.22 & 0.82 & 18.59 & 17.53 & 30.27 & 34.84 \\ \midrule
\multirow{5}{*}{\textbf{Qwen3-0.6B}} & \multirow{4}{*}{\textbf{baseline}} & \textbf{zero-shot} & 0.584 & 74.46 & 20.52 & 5.03 & 26.92 & 32.14 & 64.63 & 56.45 \\
 &  & \textbf{1-shot} & 0.571 & 76.81 & 19.17 & 4.03 & 29.81 & 34.42 & 62.24 & 55.16 \\
 &  & \textbf{5-shot} & 0.515 & 79.82 & 16.95 & 3.24 & 25.96 & 28.57 & 57.48 & 58.06 \\
 &  & \textbf{10-shot} & 0.470 & 80.74 & 15.71 & 3.55 & 25.00 & 25.97 & 48.64 & 57.10 \\
 \cmidrule(l){2-11}   & \textbf{after SFT} & \textbf{zero-shot} & 0.355 & 84.82 & 13.84 & 1.34 & 20.51 & 20.78 & 38.10 & 45.16 \\
\midrule \multirow{5}{*}{\textbf{Qwen3-1.7B-Base}} & \multirow{4}{*}{\textbf{baseline}} & \textbf{zero-shot} & 0.519 & 73.70 & 22.02 & 4.28 & 28.21 & 34.74 & 56.80 & 38.71 \\
 &  & \textbf{1-shot} & 0.467 & 74.87 & 22.92 & 2.21 & 27.88 & 30.19 & 50.34 & 36.45 \\
 &  & \textbf{5-shot} & 0.416 & 76.57 & 21.14 & 2.29 & 22.12 & 26.95 & 42.86 & 40.00 \\
 &  & \textbf{10-shot} & 0.396 & 79.56 & 18.84 & 1.60 & 23.72 & 24.35 & 41.84 & 40.32 \\
 \cmidrule(l){2-11}   & \textbf{after SFT} & \textbf{zero-shot} & 0.301 & 89.47 & 9.74 & 0.79 & 19.23 & 24.35 & 36.05 & 31.94 \\
\midrule \multirow{5}{*}{\textbf{Qwen3-1.7B}} & \multirow{4}{*}{\textbf{baseline}} & \textbf{zero-shot} & 0.602 & 67.46 & 24.64 & 7.91 & 28.53 & 37.99 & 65.31 & 36.77 \\
 &  & \textbf{1-shot} & 0.601 & 71.50 & 24.01 & 4.49 & 31.41 & 34.74 & 64.29 & 47.74 \\
 &  & \textbf{5-shot} & 0.517 & 76.61 & 20.64 & 2.75 & 26.92 & 28.25 & 58.84 & 50.65 \\
 &  & \textbf{10-shot} & 0.483 & 78.33 & 19.70 & 1.97 & 27.88 & 24.35 & 54.08 & 50.32 \\
 \cmidrule(l){2-11}   & \textbf{after SFT} & \textbf{zero-shot} & 0.332 & 83.77 & 14.56 & 1.67 & 15.71 & 24.68 & 37.41 & 37.42 \\ \midrule
 \multirow{5}{*}{\textbf{Qwen3-4B-Base}} & \multirow{4}{*}{\textbf{baseline}} & \textbf{zero-shot} & 0.546 & 76.34 & 20.46 & 3.19 & 31.73 & 32.47 & 62.59 & 46.13 \\
 &  & \textbf{1-shot} & 0.515 & 74.73 & 22.03 & 3.24 & 30.13 & 30.52 & 54.42 & 44.19 \\
 &  & \textbf{5-shot} & 0.467 & 74.19 & 22.41 & 3.40 & 26.28 & 27.27 & 48.64 & 41.29 \\
 &  & \textbf{10-shot} & 0.443 & 76.21 & 21.65 & 2.15 & 25.64 & 25.65 & 47.96 & 40.65 \\
 \cmidrule(l){2-11}   & \textbf{after SFT} & \textbf{zero-shot} & 0.267 & 89.61 & 10.09 & 0.30 & 16.03 & 20.13 & 30.95 & 31.94 \\
\midrule \multirow{5}{*}{\textbf{Qwen3-4B}} & \multirow{4}{*}{\textbf{baseline}} & \textbf{zero-shot} & 0.590 & 73.66 & 20.83 & 5.51 & 31.73 & 40.58 & 68.37 & 39.68 \\
 &  & \textbf{1-shot} & 0.534 & 76.97 & 18.13 & 4.90 & 35.26 & 33.77 & 62.24 & 39.03 \\
 &  & \textbf{5-shot} & 0.435 & 80.66 & 17.34 & 2.01 & 25.00 & 25.97 & 52.72 & 41.61 \\
 &  & \textbf{10-shot} & 0.408 & 81.91 & 16.34 & 1.75 & 23.72 & 26.95 & 48.30 & 39.35 \\
 \cmidrule(l){2-11}   & \textbf{after SFT} & \textbf{zero-shot} & 0.285 & 85.79 & 12.81 & 1.39 & 16.03 & 18.83 & 24.83 & 40.97 \\
\midrule \multirow{5}{*}{\textbf{Qwen3-8B-Base}} & \multirow{4}{*}{\textbf{baseline}} & \textbf{zero-shot} & 0.526 & 73.76 & 22.47 & 3.77 & 27.56 & 33.12 & 61.56 & 38.71 \\
 &  & \textbf{1-shot} & 0.502 & 72.67 & 23.85 & 3.48 & 27.24 & 28.57 & 56.12 & 39.35 \\
 &  & \textbf{5-shot} & 0.420 & 73.91 & 22.68 & 3.40 & 20.83 & 23.05 & 44.22 & 40.32 \\
 &  & \textbf{10-shot} & 0.401 & 77.62 & 20.40 & 1.98 & 22.44 & 25.32 & 40.14 & 40.65 \\
 \cmidrule(l){2-11}   & \textbf{after SFT} & \textbf{zero-shot} & 0.297 & 85.07 & 13.60 & 1.33 & 16.67 & 19.81 & 35.03 & 33.23 \\
\midrule \multirow{5}{*}{\textbf{Qwen3-8B}} & \multirow{4}{*}{\textbf{baseline}} & \textbf{zero-shot} & 0.558 & 71.59 & 22.87 & 5.54 & 27.24 & 34.74 & 65.31 & 38.71 \\
 &  & \textbf{1-shot} & 0.501 & 77.22 & 18.20 & 4.59 & 32.37 & 32.14 & 61.22 & 34.84 \\
 &  & \textbf{5-shot} & 0.436 & 80.73 & 16.73 & 2.55 & 26.92 & 30.84 & 48.98 & 39.03 \\
 &  & \textbf{10-shot} & 0.397 & 81.20 & 17.00 & 1.80 & 25.32 & 26.95 & 44.56 & 36.45 \\
 \cmidrule(l){2-11}   & \textbf{after SFT} & \textbf{zero-shot} & 0.278 & 83.48 & 15.10 & 1.42 & 14.74 & 16.56 & 31.97 & 32.90 \\
\midrule \multirow{4}{*}{\textbf{Qwen3-14B-Base}} & \multirow{4}{*}{\textbf{baseline}} & \textbf{zero-shot} & 0.512 & 70.08 & 23.57 & 6.36 & 24.68 & 30.84 & 53.74 & 39.35 \\
 &  & \textbf{1-shot} & 0.450 & 72.01 & 23.06 & 4.93 & 24.68 & 26.62 & 45.58 & 37.42 \\
 &  & \textbf{5-shot} & 0.380 & 75.57 & 21.50 & 2.92 & 20.83 & 22.73 & 37.76 & 37.42 \\
 &  & \textbf{10-shot} & 0.347 & 76.89 & 21.05 & 2.06 & 20.51 & 20.13 & 31.97 & 37.42 \\
\midrule \multirow{4}{*}{\textbf{Qwen3-14B}} & \multirow{4}{*}{\textbf{baseline}} & \textbf{zero-shot} & 0.560 & 72.52 & 21.10 & 6.37 & 27.88 & 32.47 & 65.31 & 42.90 \\
 &  & \textbf{1-shot} & 0.500 & 78.92 & 16.64 & 4.44 & 32.37 & 31.17 & 59.18 & 40.97 \\
 &  & \textbf{5-shot} & 0.429 & 80.41 & 17.19 & 2.40 & 25.64 & 26.95 & 47.28 & 42.90 \\
 &  & \textbf{10-shot} & 0.381 & 82.29 & 15.62 & 2.08 & 23.72 & 25.00 & 36.05 & 44.52 \\
\midrule \multirow{4}{*}{\textbf{Qwen3-32B}} & \multirow{4}{*}{\textbf{baseline}} & \textbf{zero-shot} & 0.531 & 76.68 & 17.64 & 5.68 & 27.24 & 34.42 & 67.01 & 40.32 \\
 &  & \textbf{1-shot} & 0.462 & 77.49 & 17.70 & 4.81 & 26.28 & 28.25 & 55.78 & 38.06 \\
 &  & \textbf{5-shot} & 0.347 & 81.46 & 15.79 & 2.75 & 18.91 & 22.73 & 39.80 & 35.48 \\
 &  & \textbf{10-shot} & 0.316 & 83.17 & 14.32 & 2.51 & 19.87 & 20.78 & 32.31 & 35.48 \\ \bottomrule
\end{tabular}%
}
\caption{Error rates, incorrect choice distributions, and local negation confusion rates for the \textbf{Qwen3} family under zero-shot, few-shot, and SFT conditions, evaluated in the \textbf{completion-based setting} using \textbf{definition instruction.}}
\label{tab:neg_qwen_def_completion}
\end{table*}

\begin{table*}[htbp]
\centering
\resizebox{\textwidth}{!}{%
\begin{tabular}{@{}cccr|rrr|rrrr@{}}
\toprule
\multirow{2}{*}{\textbf{Model}} & \multirow{2}{*}{\textbf{\begin{tabular}[c]{@{}c@{}}Training\\ Setting\end{tabular}}} & \multirow{2}{*}{\textbf{N Shot}} & \multicolumn{1}{c|}{\multirow{2}{*}{\textbf{\begin{tabular}[c]{@{}c@{}}Error \\ Rate\\ (1-acc)\end{tabular}}}} & \multicolumn{3}{c|}{\textbf{Incorrect Choice Distribution}} & \multicolumn{4}{c}{\textbf{Local Negation Confusion Rate}} \\ \cmidrule(l){5-11} 
 &  &  & \multicolumn{1}{c|}{} & \multicolumn{1}{c}{\textbf{\begin{tabular}[c]{@{}c@{}}Local\\ Negation\\ (\%)\end{tabular}}} & \multicolumn{1}{c}{\textbf{\begin{tabular}[c]{@{}c@{}}Contra-\\ diction\\ (\%)\end{tabular}}} & \multicolumn{1}{c|}{\textbf{\begin{tabular}[c]{@{}c@{}}Para-\\ phrase\\ (\%)\end{tabular}}} & \multicolumn{1}{c}{\textbf{\begin{tabular}[c]{@{}c@{}}Relative\\ Clause\\ (\%)\end{tabular}}} & \multicolumn{1}{c}{\textbf{\begin{tabular}[c]{@{}c@{}}Participle\\ Clause\\ (\%)\end{tabular}}} & \multicolumn{1}{c}{\textbf{\begin{tabular}[c]{@{}c@{}}Compound\\ Sentence\\ (\%)\end{tabular}}} & \multicolumn{1}{c}{\textbf{\begin{tabular}[c]{@{}c@{}}Adverbial\\ Clause\\ (\%)\end{tabular}}} \\ \midrule
\multirow{5}{*}{\textbf{Qwen3-0.6B-Base}} & \multirow{4}{*}{\textbf{baseline}} & \textbf{zero-shot} & 0.651 & 50.55 & 7.06 & 42.39 & 25.32 & 37.99 & 39.80 & 32.90 \\
 &  & \textbf{1-shot} & 0.603 & 55.26 & 6.84 & 37.89 & 25.64 & 37.01 & 48.98 & 26.45 \\
 &  & \textbf{5-shot} & 0.577 & 55.57 & 3.30 & 41.13 & 27.56 & 33.12 & 47.28 & 24.84 \\
 &  & \textbf{10-shot} & 0.525 & 60.12 & 3.78 & 36.10 & 29.81 & 30.52 & 45.92 & 24.52 \\ \cmidrule(l){2-11} 
 & \textbf{after SFT} & \textbf{zero-shot} & 0.521 & 67.43 & 11.42 & 21.16 & 32.37 & 39.94 & 42.18 & 30.65 \\ \midrule
\multirow{5}{*}{\textbf{Qwen3-0.6B}} & \multirow{4}{*}{\textbf{baseline}} & \textbf{zero-shot} & 0.562 & 69.96 & 7.90 & 22.14 & 37.82 & 38.64 & 54.08 & 32.26 \\
 &  & \textbf{1-shot} & 0.532 & 59.31 & 14.75 & 25.93 & 26.28 & 29.87 & 47.28 & 27.42 \\
 &  & \textbf{5-shot} & 0.493 & 71.66 & 5.80 & 22.54 & 32.37 & 30.52 & 54.76 & 28.71 \\
 &  & \textbf{10-shot} & 0.469 & 69.88 & 6.60 & 23.52 & 32.37 & 29.87 & 49.32 & 24.19 \\ \cmidrule(l){2-11} 
 & \textbf{after SFT} & \textbf{zero-shot} & 0.445 & 67.02 & 10.34 & 22.64 & 24.68 & 27.27 & 47.96 & 23.87 \\ \midrule
\multirow{5}{*}{\textbf{Qwen3-1.7B-Base}} & \multirow{4}{*}{\textbf{baseline}} & \textbf{zero-shot} & 0.474 & 65.22 & 9.87 & 24.92 & 31.41 & 23.70 & 52.04 & 21.29 \\
 &  & \textbf{1-shot} & 0.444 & 72.50 & 13.57 & 13.93 & 33.97 & 30.84 & 38.78 & 29.35 \\
 &  & \textbf{5-shot} & 0.361 & 75.82 & 12.53 & 11.65 & 29.81 & 25.32 & 36.39 & 21.61 \\
 &  & \textbf{10-shot} & 0.313 & 76.14 & 11.42 & 12.44 & 25.96 & 20.78 & 34.69 & 17.10 \\ \cmidrule(l){2-11} 
 & \textbf{after SFT} & \textbf{zero-shot} & 0.507 & 58.84 & 12.83 & 28.33 & 33.97 & 26.30 & 43.54 & 19.68 \\ \midrule
\multirow{5}{*}{\textbf{Qwen3-1.7B}} & \multirow{4}{*}{\textbf{baseline}} & \textbf{zero-shot} & 0.468 & 70.85 & 15.76 & 13.39 & 26.92 & 31.82 & 46.26 & 32.26 \\
 &  & \textbf{1-shot} & 0.451 & 67.31 & 14.24 & 18.45 & 24.04 & 28.57 & 47.62 & 25.81 \\
 &  & \textbf{5-shot} & 0.424 & 72.47 & 11.80 & 15.73 & 26.60 & 23.70 & 51.70 & 25.48 \\
 &  & \textbf{10-shot} & 0.421 & 72.13 & 9.98 & 17.89 & 25.96 & 23.70 & 53.40 & 23.23 \\ \cmidrule(l){2-11} 
 & \textbf{after SFT} & \textbf{zero-shot} & 0.392 & 68.83 & 22.27 & 8.91 & 30.13 & 27.92 & 30.27 & 22.90 \\ \midrule
\multirow{5}{*}{\textbf{Qwen3-4B-Base}} & \multirow{4}{*}{\textbf{baseline}} & \textbf{zero-shot} & 0.398 & 78.69 & 15.74 & 5.58 & 38.14 & 22.40 & 45.58 & 23.55 \\
 &  & \textbf{1-shot} & 0.386 & 77.82 & 19.30 & 2.87 & 40.71 & 29.22 & 36.73 & 17.42 \\
 &  & \textbf{5-shot} & 0.350 & 75.74 & 19.27 & 4.99 & 40.38 & 26.95 & 24.15 & 17.42 \\
 &  & \textbf{10-shot} & 0.336 & 77.36 & 17.45 & 5.19 & 39.10 & 26.95 & 24.49 & 16.45 \\ \cmidrule(l){2-11} 
 & \textbf{after SFT} & \textbf{zero-shot} & 0.291 & 89.37 & 8.45 & 2.18 & 32.05 & 18.18 & 43.20 & 14.52 \\ \midrule
\multirow{5}{*}{\textbf{Qwen3-4B}} & \multirow{4}{*}{\textbf{baseline}} & \textbf{zero-shot} & 0.439 & 82.28 & 15.55 & 2.17 & 38.46 & 30.19 & 56.46 & 24.52 \\
 &  & \textbf{1-shot} & 0.375 & 80.97 & 17.34 & 1.69 & 33.97 & 25.32 & 46.26 & 20.32 \\
 &  & \textbf{5-shot} & 0.321 & 84.20 & 12.35 & 3.46 & 32.69 & 23.70 & 33.67 & 21.61 \\
 &  & \textbf{10-shot} & 0.301 & 83.64 & 13.19 & 3.17 & 30.77 & 24.68 & 26.87 & 21.29 \\ \cmidrule(l){2-11} 
 & \textbf{after SFT} & \textbf{zero-shot} & 0.278 & 87.14 & 9.43 & 3.43 & 28.53 & 22.73 & 37.76 & 11.29 \\ \midrule
\multirow{5}{*}{\textbf{Qwen3-8B-Base}} & \multirow{4}{*}{\textbf{baseline}} & \textbf{zero-shot} & 0.310 & 79.03 & 13.81 & 7.16 & 29.17 & 20.78 & 33.33 & 18.06 \\
 &  & \textbf{1-shot} & 0.313 & 76.96 & 13.42 & 9.62 & 31.73 & 26.95 & 23.81 & 16.77 \\
 &  & \textbf{5-shot} & 0.254 & 76.25 & 15.00 & 8.75 & 28.21 & 21.10 & 16.67 & 13.55 \\
 &  & \textbf{10-shot} & 0.250 & 78.41 & 13.02 & 8.57 & 27.56 & 20.45 & 16.67 & 15.81 \\ \cmidrule(l){2-11} 
 & \textbf{after SFT} & \textbf{zero-shot} & 0.232 & 72.7 & 20.82 & 6.48 & 19.55 & 17.21 & 18.71 & 14.19 \\ \midrule
\multirow{5}{*}{\textbf{Qwen3-8B}} & \multirow{4}{*}{\textbf{baseline}} & \textbf{zero-shot} & 0.411 & 84.17 & 12.55 & 3.28 & 41.67 & 31.82 & 49.32 & 20.32 \\
 &  & \textbf{1-shot} & 0.326 & 82.00 & 13.14 & 4.87 & 35.90 & 26.30 & 33.33 & 14.84 \\
 &  & \textbf{5-shot} & 0.279 & 83.81 & 10.80 & 5.40 & 34.29 & 25.32 & 21.09 & 15.48 \\
 &  & \textbf{10-shot} & 0.256 & 82.35 & 11.15 & 6.50 & 30.77 & 23.38 & 16.67 & 15.81 \\ \cmidrule(l){2-11} 
 & \textbf{after SFT} & \textbf{zero-shot} & 0.217 & 84.62 & 11.72 & 3.66 & 25.00 & 22.40 & 15.65 & 12.26 \\ \midrule
\multirow{4}{*}{\textbf{Qwen3-14B-Base}} & \multirow{4}{*}{\textbf{baseline}} & \textbf{zero-shot} & 0.292 & 81.52 & 15.22 & 3.26 & 23.72 & 16.56 & 39.12 & 19.35 \\
 &  & \textbf{1-shot} & 0.236 & 75.17 & 18.79 & 6.04 & 21.15 & 16.56 & 26.53 & 9.35 \\
 &  & \textbf{5-shot} & 0.190 & 73.22 & 23.01 & 3.77 & 18.59 & 14.61 & 13.27 & 10.65 \\
 &  & \textbf{10-shot} & 0.177 & 75.34 & 21.97 & 2.69 & 17.95 & 14.61 & 9.52 & 12.58 \\ \midrule
\multirow{4}{*}{\textbf{Qwen3-14B}} & \multirow{4}{*}{\textbf{baseline}} & \textbf{zero-shot} & 0.340 & 79.67 & 17.52 & 2.80 & 32.37 & 22.08 & 45.92 & 11.94 \\
 &  & \textbf{1-shot} & 0.316 & 82.96 & 12.28 & 4.76 & 30.77 & 21.75 & 43.20 & 13.23 \\
 &  & \textbf{5-shot} & 0.263 & 85.20 & 12.99 & 1.81 & 31.09 & 25.32 & 26.19 & 9.68 \\
 &  & \textbf{10-shot} & 0.250 & 84.76 & 12.70 & 2.54 & 33.01 & 21.75 & 20.75 & 11.61 \\ \midrule
\multirow{4}{*}{\textbf{Qwen3-32B}} & \multirow{4}{*}{\textbf{baseline}} & \textbf{zero-shot} & 0.270 & 81.23 & 16.72 & 2.05 & 25.00 & 24.35 & 26.53 & 14.84 \\
 &  & \textbf{1-shot} & 0.232 & 80.82 & 18.15 & 1.03 & 24.68 & 23.05 & 19.73 & 9.68 \\
 &  & \textbf{5-shot} & 0.201 & 81.10 & 18.50 & 0.39 & 24.36 & 19.48 & 14.97 & 8.39 \\
 &  & \textbf{10-shot} & 0.181 & 77.63 & 21.93 & 0.44 & 19.55 & 16.88 & 12.24 & 9.03 \\ \bottomrule
\end{tabular}%
}
\caption{Error rates, incorrect choice distributions, and local negation confusion rates for the \textbf{Qwen3} family under zero-shot, few-shot, and SFT conditions, evaluated in the \textbf{option-selection setting} using \textbf{definition instruction.}}
\label{tab:neg_qwen_def_option}
\end{table*}

\begin{table*}[]
\centering
\resizebox{\textwidth}{!}{%
\begin{tabular}{@{}cccr|rrr|rrrr@{}}
\toprule
\multirow{2}{*}{\textbf{Model}} & \multirow{2}{*}{\textbf{\begin{tabular}[c]{@{}c@{}}Training\\ Setting\end{tabular}}} & \multirow{2}{*}{\textbf{N Shot}} & \multicolumn{1}{c|}{\multirow{2}{*}{\textbf{\begin{tabular}[c]{@{}c@{}}Error \\ Rate\\ (1-acc)\end{tabular}}}} & \multicolumn{3}{c|}{\textbf{Incorrect Choice Distribution}} & \multicolumn{4}{c}{\textbf{Local Negation Confusion Rate}} \\ \cmidrule(l){5-11} 
 &  &  & \multicolumn{1}{c|}{} & \multicolumn{1}{c}{\textbf{\begin{tabular}[c]{@{}c@{}}Local\\ Negation\\ (\%)\end{tabular}}} & \multicolumn{1}{c}{\textbf{\begin{tabular}[c]{@{}c@{}}Contra-\\ diction\\ (\%)\end{tabular}}} & \multicolumn{1}{c|}{\textbf{\begin{tabular}[c]{@{}c@{}}Para-\\ phrase\\ (\%)\end{tabular}}} & \multicolumn{1}{c}{\textbf{\begin{tabular}[c]{@{}c@{}}Relative\\ Clause\\ (\%)\end{tabular}}} & \multicolumn{1}{c}{\textbf{\begin{tabular}[c]{@{}c@{}}Participle\\ Clause\\ (\%)\end{tabular}}} & \multicolumn{1}{c}{\textbf{\begin{tabular}[c]{@{}c@{}}Compound\\ Sentence\\ (\%)\end{tabular}}} & \multicolumn{1}{c}{\textbf{\begin{tabular}[c]{@{}c@{}}Adverbial\\ Clause\\ (\%)\end{tabular}}} \\ \midrule
\multirow{5}{*}{\textbf{Qwen3-0.6B-Base}} & \multirow{4}{*}{\textbf{baseline}} & \textbf{zero-shot} & 0.577 & 71.53 & 22.97 & 5.50 & 29.81 & 35.39 & 54.42 & 50.97 \\
 &  & \textbf{1-shot} & 0.502 & 73.46 & 21.80 & 4.74 & 29.17 & 30.84 & 50.34 & 42.26 \\
 &  & \textbf{5-shot} & 0.426 & 77.28 & 19.18 & 3.54 & 23.40 & 23.70 & 44.22 & 44.84 \\
 &  & \textbf{10-shot} & 0.401 & 79.01 & 18.61 & 2.38 & 22.12 & 23.70 & 42.18 & 42.90 \\ \cmidrule(l){2-11} 
 & \textbf{after SFT} & \textbf{zero-shot} & 0.277 & 83.67 & 15.47 & 0.86 & 17.31 & 17.53 & 28.23 & 32.58 \\ \midrule
\multirow{5}{*}{\textbf{Qwen3-0.6B}} & \multirow{4}{*}{\textbf{baseline}} & \textbf{zero-shot} & 0.587 & 77.70 & 17.57 & 4.73 & 27.24 & 41.56 & 62.93 & 57.10 \\
 &  & \textbf{1-shot} & 0.590 & 76.48 & 18.82 & 4.70 & 30.77 & 38.64 & 64.29 & 53.23 \\
 &  & \textbf{5-shot} & 0.517 & 78.53 & 17.18 & 4.29 & 25.00 & 31.82 & 57.82 & 53.55 \\
 &  & \textbf{10-shot} & 0.471 & 81.31 & 15.15 & 3.54 & 25.64 & 27.92 & 52.38 & 52.58 \\ \cmidrule(l){2-11} 
 & \textbf{after SFT} & \textbf{zero-shot} & 0.321 & 84.44 & 14.07 & 1.48 & 17.63 & 20.13 & 35.03 & 39.35 \\ \midrule
\multirow{5}{*}{\textbf{Qwen3-1.7B-Base}} & \multirow{4}{*}{\textbf{baseline}} & \textbf{zero-shot} & 0.538 & 71.24 & 22.71 & 6.05 & 29.81 & 33.12 & 58.50 & 37.42 \\
 &  & \textbf{1-shot} & 0.479 & 74.17 & 22.85 & 2.98 & 28.21 & 30.52 & 53.06 & 35.48 \\
 &  & \textbf{5-shot} & 0.417 & 77.38 & 20.53 & 2.09 & 23.40 & 26.30 & 41.84 & 41.94 \\
 &  & \textbf{10-shot} & 0.401 & 80.40 & 18.22 & 1.39 & 23.72 & 24.68 & 41.84 & 42.90 \\ \cmidrule(l){2-11} 
 & \textbf{after SFT} & \textbf{zero-shot} & 0.276 & 85.63 & 12.93 & 1.44 & 17.95 & 24.03 & 31.97 & 23.87 \\ \midrule
\multirow{5}{*}{\textbf{Qwen3-1.7B}} & \multirow{4}{*}{\textbf{baseline}} & \textbf{zero-shot} & 0.581 & 67.76 & 25.27 & 6.97 & 27.88 & 37.66 & 63.27 & 34.52 \\
 &  & \textbf{1-shot} & 0.570 & 71.35 & 24.34 & 4.31 & 29.49 & 33.77 & 61.22 & 44.19 \\
 &  & \textbf{5-shot} & 0.504 & 76.54 & 21.10 & 2.36 & 26.60 & 26.95 & 58.16 & 48.06 \\
 &  & \textbf{10-shot} & 0.485 & 77.45 & 20.42 & 2.12 & 27.56 & 25.32 & 54.42 & 48.39 \\ \cmidrule(l){2-11} 
 & \textbf{after SFT} & \textbf{zero-shot} & 0.358 & 82.71 & 15.74 & 1.55 & 18.59 & 26.30 & 39.12 & 38.39 \\ \midrule
\multirow{5}{*}{\textbf{Qwen3-4B-Base}} & \multirow{4}{*}{\textbf{baseline}} & \textbf{zero-shot} & 0.515 & 75.50 & 21.57 & 2.93 & 27.56 & 30.84 & 55.10 & 47.42 \\
 &  & \textbf{1-shot} & 0.494 & 75.12 & 22.15 & 2.73 & 29.49 & 29.55 & 52.72 & 41.94 \\
 &  & \textbf{5-shot} & 0.449 & 74.20 & 22.79 & 3.00 & 25.00 & 27.27 & 44.22 & 41.29 \\
 &  & \textbf{10-shot} & 0.440 & 76.22 & 21.80 & 1.98 & 25.64 & 25.97 & 45.58 & 41.61 \\ \cmidrule(l){2-11} 
 & \textbf{after SFT} & \textbf{zero-shot} & 0.271 & 86.26 & 12.87 & 0.88 & 16.67 & 19.48 & 31.63 & 29.03 \\ \midrule
\multirow{5}{*}{\textbf{Qwen3-4B}} & \multirow{4}{*}{\textbf{baseline}} & \textbf{zero-shot} & 0.562 & 73.73 & 20.34 & 5.93 & 28.21 & 38.96 & 65.31 & 39.35 \\
 &  & \textbf{1-shot} & 0.495 & 74.68 & 20.19 & 5.13 & 28.85 & 29.87 & 57.82 & 36.77 \\
 &  & \textbf{5-shot} & 0.431 & 78.12 & 19.12 & 2.76 & 23.08 & 25.65 & 51.70 & 39.35 \\
 &  & \textbf{10-shot} & 0.398 & 79.08 & 18.53 & 2.39 & 21.79 & 25.65 & 46.60 & 36.45 \\ \cmidrule(l){2-11} 
 & \textbf{after SFT} & \textbf{zero-shot} & 0.313 & 83.76 & 14.47 & 1.78 & 16.67 & 21.10 & 28.91 & 41.29 \\ \midrule
\multirow{5}{*}{\textbf{Qwen3-8B-Base}} & \multirow{4}{*}{\textbf{baseline}} & \textbf{zero-shot} & 0.545 & 74.09 & 21.69 & 4.22 & 29.17 & 33.44 & 61.22 & 43.55 \\
 &  & \textbf{1-shot} & 0.489 & 73.42 & 23.18 & 3.40 & 27.56 & 28.25 & 55.44 & 37.74 \\
 &  & \textbf{5-shot} & 0.436 & 73.09 & 23.27 & 3.64 & 22.44 & 24.35 & 44.90 & 40.32 \\
 &  & \textbf{10-shot} & 0.412 & 76.35 & 21.15 & 2.50 & 22.76 & 25.65 & 40.48 & 41.29 \\ \cmidrule(l){2-11} 
 & \textbf{after SFT} & \textbf{zero-shot} & 0.286 & 86.43 & 12.47 & 1.11 & 16.35 & 20.45 & 34.01 & 31.61 \\ \midrule
\multirow{5}{*}{\textbf{Qwen3-8B}} & \multirow{4}{*}{\textbf{baseline}} & \textbf{zero-shot} & 0.527 & 71.84 & 23.04 & 5.12 & 26.60 & 35.06 & 61.22 & 34.19 \\
 &  & \textbf{1-shot} & 0.484 & 75.41 & 20.49 & 4.10 & 31.73 & 30.19 & 57.14 & 32.26 \\
 &  & \textbf{5-shot} & 0.442 & 80.43 & 17.59 & 1.97 & 27.56 & 30.52 & 49.32 & 39.68 \\
 &  & \textbf{10-shot} & 0.408 & 81.36 & 16.70 & 1.94 & 25.64 & 28.57 & 45.24 & 38.06 \\ \cmidrule(l){2-11} 
 & \textbf{after SFT} & \textbf{zero-shot} & 0.260 & 83.84 & 14.33 & 1.83 & 13.78 & 17.21 & 27.55 & 31.61 \\ \midrule
\multirow{4}{*}{\textbf{Qwen3-14B-Base}} & \multirow{4}{*}{\textbf{baseline}} & \textbf{zero-shot} & 0.493 & 70.69 & 23.67 & 5.64 & 25.00 & 31.17 & 46.60 & 41.29 \\
 &  & \textbf{1-shot} & 0.443 & 74.37 & 22.58 & 3.05 & 25.64 & 27.27 & 46.26 & 37.10 \\
 &  & \textbf{5-shot} & 0.369 & 76.99 & 20.65 & 2.37 & 20.19 & 22.08 & 35.37 & 39.68 \\
 &  & \textbf{10-shot} & 0.335 & 77.07 & 21.75 & 1.18 & 20.51 & 19.81 & 28.91 & 37.42 \\ \midrule
\multirow{4}{*}{\textbf{Qwen3-14B}} & \multirow{4}{*}{\textbf{baseline}} & \textbf{zero-shot} & 0.512 & 69.35 & 23.37 & 7.28 & 25.64 & 33.12 & 56.46 & 32.26 \\
 &  & \textbf{1-shot} & 0.470 & 75.89 & 18.89 & 5.23 & 31.09 & 30.84 & 55.78 & 30.32 \\
 &  & \textbf{5-shot} & 0.397 & 79.60 & 18.60 & 1.80 & 23.72 & 25.65 & 44.90 & 36.45 \\
 &  & \textbf{10-shot} & 0.361 & 79.56 & 18.24 & 2.20 & 23.72 & 24.03 & 33.67 & 37.10 \\ \midrule
\multirow{4}{*}{\textbf{Qwen3-32B}} & \multirow{4}{*}{\textbf{baseline}} & \textbf{zero-shot} & 0.486 & 70.31 & 20.72 & 8.97 & 24.68 & 35.06 & 60.88 & 21.61 \\
 &  & \textbf{1-shot} & 0.418 & 74.95 & 18.60 & 6.45 & 23.72 & 27.60 & 53.06 & 25.81 \\
 &  & \textbf{5-shot} & 0.341 & 81.40 & 15.58 & 3.02 & 20.51 & 23.38 & 40.14 & 30.97 \\
 &  & \textbf{10-shot} & 0.309 & 82.01 & 15.17 & 2.83 & 19.23 & 21.75 & 32.99 & 30.65 \\ \bottomrule
\end{tabular}%
}
\caption{Error rates, incorrect choice distributions, and local negation confusion rates for the \textbf{Qwen3} family under zero-shot, few-shot, and SFT conditions, evaluated in the \textbf{completion-based setting} using \textbf{detailed instruction.}}
\label{tab:neg_qwen_detail_completion}
\end{table*}

\begin{table*}[htbp]
\centering
\resizebox{\textwidth}{!}{%
\begin{tabular}{@{}cccr|rrr|rrrr@{}}
\toprule
\multirow{2}{*}{\textbf{Model}} & \multirow{2}{*}{\textbf{\begin{tabular}[c]{@{}c@{}}Training\\ Setting\end{tabular}}} & \multirow{2}{*}{\textbf{N Shot}} & \multicolumn{1}{c|}{\multirow{2}{*}{\textbf{\begin{tabular}[c]{@{}c@{}}Error \\ Rate\\ (1-acc)\end{tabular}}}} & \multicolumn{3}{c|}{\textbf{Incorrect Choice Distribution}} & \multicolumn{4}{c}{\textbf{Local Negation Confusion Rate}} \\ \cmidrule(l){5-11} 
 &  &  & \multicolumn{1}{c|}{} & \multicolumn{1}{c}{\textbf{\begin{tabular}[c]{@{}c@{}}Local\\ Negation\\ (\%)\end{tabular}}} & \multicolumn{1}{c}{\textbf{\begin{tabular}[c]{@{}c@{}}Contra-\\ diction\\ (\%)\end{tabular}}} & \multicolumn{1}{c|}{\textbf{\begin{tabular}[c]{@{}c@{}}Para-\\ phrase\\ (\%)\end{tabular}}} & \multicolumn{1}{c}{\textbf{\begin{tabular}[c]{@{}c@{}}Relative\\ Clause\\ (\%)\end{tabular}}} & \multicolumn{1}{c}{\textbf{\begin{tabular}[c]{@{}c@{}}Participle\\ Clause\\ (\%)\end{tabular}}} & \multicolumn{1}{c}{\textbf{\begin{tabular}[c]{@{}c@{}}Compound\\ Sentence\\ (\%)\end{tabular}}} & \multicolumn{1}{c}{\textbf{\begin{tabular}[c]{@{}c@{}}Adverbial\\ Clause\\ (\%)\end{tabular}}} \\ \midrule
\multirow{5}{*}{\textbf{Qwen3-0.6B-Base}} & \multirow{4}{*}{\textbf{baseline}} & \textbf{zero-shot} & 0.658 & 48.80 & 4.46 & 46.75 & 27.88 & 37.66 & 38.78 & 28.39 \\
 &  & \textbf{1-shot} & 0.606 & 53.14 & 7.46 & 39.40 & 25.00 & 36.04 & 47.62 & 24.84 \\
 &  & \textbf{5-shot} & 0.588 & 52.77 & 3.24 & 43.99 & 26.60 & 33.12 & 43.20 & 25.48 \\
 &  & \textbf{10-shot} & 0.537 & 61.00 & 3.55 & 35.45 & 30.45 & 33.77 & 46.60 & 24.84 \\ \cmidrule(l){2-11} 
 & \textbf{after SFT} & \textbf{zero-shot} & 0.675 & 37.25 & 30.79 & 31.96 & 23.08 & 27.92 & 31.63 & 21.29 \\ \midrule
\multirow{5}{*}{\textbf{Qwen3-0.6B}} & \multirow{4}{*}{\textbf{baseline}} & \textbf{zero-shot} & 0.541 & 74.93 & 7.92 & 17.16 & 43.91 & 40.91 & 57.48 & 25.48 \\
 &  & \textbf{1-shot} & 0.588 & 53.37 & 15.36 & 31.27 & 25.96 & 29.87 & 45.58 & 28.71 \\
 &  & \textbf{5-shot} & 0.535 & 65.04 & 6.81 & 28.15 & 31.09 & 31.49 & 52.72 & 29.03 \\
 &  & \textbf{10-shot} & 0.502 & 65.72 & 6.64 & 27.65 & 31.73 & 29.87 & 52.72 & 22.58 \\ \cmidrule(l){2-11} 
 & \textbf{after SFT} & \textbf{zero-shot} & 0.474 & 64.55 & 20.57 & 14.88 & 27.56 & 29.22 & 46.60 & 23.55 \\ \midrule
\multirow{5}{*}{\textbf{Qwen3-1.7B-Base}} & \multirow{4}{*}{\textbf{baseline}} & \textbf{zero-shot} & 0.517 & 57.21 & 13.50 & 29.29 & 26.92 & 28.25 & 49.32 & 18.39 \\
 &  & \textbf{1-shot} & 0.457 & 69.97 & 13.89 & 16.15 & 31.73 & 32.79 & 40.14 & 27.42 \\
 &  & \textbf{5-shot} & 0.382 & 71.78 & 12.86 & 15.35 & 29.49 & 25.97 & 38.10 & 20.00 \\
 &  & \textbf{10-shot} & 0.343 & 72.29 & 11.55 & 16.17 & 25.32 & 23.70 & 36.73 & 17.10 \\ \cmidrule(l){2-11} 
 & \textbf{after SFT} & \textbf{zero-shot} & 0.391 & 68.15 & 13.18 & 18.66 & 30.45 & 26.30 & 41.16 & 12.58 \\ \midrule
\multirow{5}{*}{\textbf{Qwen3-1.7B}} & \multirow{4}{*}{\textbf{baseline}} & \textbf{zero-shot} & 0.553 & 71.74 & 18.94 & 9.33 & 37.50 & 37.99 & 53.74 & 34.84 \\
 &  & \textbf{1-shot} & 0.508 & 68.12 & 14.06 & 17.81 & 30.45 & 31.82 & 55.44 & 25.81 \\
 &  & \textbf{5-shot} & 0.450 & 72.31 & 10.41 & 17.28 & 27.56 & 25.97 & 56.80 & 24.84 \\
 &  & \textbf{10-shot} & 0.455 & 71.25 & 10.10 & 18.64 & 30.13 & 25.65 & 55.78 & 23.23 \\ \cmidrule(l){2-11} 
 & \textbf{after SFT} & \textbf{zero-shot} & 0.427 & 66.05 & 26.90 & 7.05 & 29.81 & 29.87 & 34.35 & 22.58 \\ \midrule
\multirow{5}{*}{\textbf{Qwen3-4B-Base}} & \multirow{4}{*}{\textbf{baseline}} & \textbf{zero-shot} & 0.416 & 79.01 & 14.89 & 6.11 & 36.86 & 24.68 & 48.64 & 25.81 \\
 &  & \textbf{1-shot} & 0.413 & 76.20 & 18.23 & 5.57 & 41.35 & 27.92 & 42.86 & 18.06 \\
 &  & \textbf{5-shot} & 0.366 & 75.49 & 17.57 & 6.94 & 39.42 & 25.97 & 30.27 & 18.06 \\
 &  & \textbf{10-shot} & 0.355 & 76.06 & 17.00 & 6.94 & 40.71 & 26.30 & 27.21 & 16.77 \\ \cmidrule(l){2-11} 
 & \textbf{after SFT} & \textbf{zero-shot} & 0.290 & 84.38 & 11.78 & 3.84 & 28.85 & 15.91 & 39.80 & 16.77 \\ \midrule
\multirow{5}{*}{\textbf{Qwen3-4B}} & \multirow{4}{*}{\textbf{baseline}} & \textbf{zero-shot} & 0.424 & 81.27 & 17.23 & 1.50 & 37.18 & 29.22 & 51.36 & 24.84 \\
 &  & \textbf{1-shot} & 0.364 & 79.52 & 18.52 & 1.96 & 35.26 & 24.03 & 41.16 & 19.35 \\
 &  & \textbf{5-shot} & 0.347 & 82.88 & 14.61 & 2.51 & 32.37 & 26.30 & 38.10 & 22.26 \\
 &  & \textbf{10-shot} & 0.317 & 82.00 & 14.50 & 3.50 & 33.97 & 23.70 & 29.93 & 19.68 \\ \cmidrule(l){2-11} 
 & \textbf{after SFT} & \textbf{zero-shot} & 0.289 & 81.32 & 17.58 & 1.10 & 28.21 & 21.43 & 35.37 & 12.26 \\ \midrule
\multirow{5}{*}{\textbf{Qwen3-8B-Base}} & \multirow{4}{*}{\textbf{baseline}} & \textbf{zero-shot} & 0.332 & 78.04 & 12.41 & 9.55 & 29.81 & 24.68 & 36.05 & 16.77 \\
 &  & \textbf{1-shot} & 0.305 & 77.86 & 13.80 & 8.33 & 30.45 & 26.62 & 25.51 & 15.16 \\
 &  & \textbf{5-shot} & 0.258 & 75.08 & 16.31 & 8.62 & 27.24 & 20.13 & 18.37 & 13.87 \\
 &  & \textbf{10-shot} & 0.251 & 78.23 & 12.93 & 8.83 & 27.24 & 19.81 & 18.03 & 15.81 \\ \cmidrule(l){2-11} 
 & \textbf{after SFT} & \textbf{zero-shot} & 0.177 & 74.89 & 16.14 & 8.97 & 17.63 & 14.61 & 9.52 & 12.58 \\ \midrule
\multirow{5}{*}{\textbf{Qwen3-8B}} & \multirow{4}{*}{\textbf{baseline}} & \textbf{zero-shot} & 0.381 & 86.07 & 12.27 & 1.66 & 43.59 & 33.12 & 37.41 & 21.29 \\
 &  & \textbf{1-shot} & 0.305 & 83.38 & 14.03 & 2.60 & 36.54 & 24.03 & 26.87 & 17.42 \\
 &  & \textbf{5-shot} & 0.297 & 81.87 & 13.60 & 4.53 & 34.62 & 26.30 & 21.77 & 17.42 \\
 &  & \textbf{10-shot} & 0.276 & 81.32 & 12.93 & 5.75 & 31.73 & 23.70 & 18.71 & 18.06 \\ \cmidrule(l){2-11} 
 & \textbf{after SFT} & \textbf{zero-shot} & 0.240 & 85.48 & 10.56 & 3.96 & 29.17 & 21.43 & 19.05 & 14.84 \\ \midrule
\multirow{4}{*}{\textbf{Qwen3-14B-Base}} & \multirow{4}{*}{\textbf{baseline}} & \textbf{zero-shot} & 0.264 & 79.58 & 15.92 & 4.50 & 20.83 & 16.56 & 27.89 & 21.61 \\
 &  & \textbf{1-shot} & 0.242 & 73.11 & 21.97 & 4.92 & 21.79 & 16.88 & 20.75 & 13.55 \\
 &  & \textbf{5-shot} & 0.210 & 75.09 & 21.51 & 3.40 & 22.12 & 15.58 & 13.95 & 13.23 \\
 &  & \textbf{10-shot} & 0.182 & 77.29 & 20.52 & 2.18 & 18.59 & 15.58 & 8.50 & 14.84 \\ \midrule
\multirow{4}{*}{\textbf{Qwen3-14B}} & \multirow{4}{*}{\textbf{baseline}} & \textbf{zero-shot} & 0.280 & 79.04 & 19.55 & 1.42 & 28.85 & 23.05 & 25.17 & 14.19 \\
 &  & \textbf{1-shot} & 0.291 & 80.65 & 15.53 & 3.81 & 31.73 & 21.10 & 29.93 & 14.19 \\
 &  & \textbf{5-shot} & 0.241 & 86.51 & 11.18 & 2.30 & 30.13 & 20.78 & 20.75 & 14.19 \\
 &  & \textbf{10-shot} & 0.241 & 84.21 & 13.82 & 1.97 & 31.73 & 21.75 & 16.33 & 13.55 \\ \midrule
\multirow{4}{*}{\textbf{Qwen3-32B}} & \multirow{4}{*}{\textbf{baseline}} & \textbf{zero-shot} & 0.240 & 78.81 & 19.54 & 1.66 & 20.83 & 24.35 & 18.03 & 14.52 \\
 &  & \textbf{1-shot} & 0.189 & 80.67 & 18.49 & 0.84 & 22.44 & 20.45 & 10.88 & 8.71 \\
 &  & \textbf{5-shot} & 0.194 & 80.82 & 18.78 & 0.41 & 21.47 & 20.45 & 12.93 & 9.68 \\
 &  & \textbf{10-shot} & 0.175 & 80.54 & 19.00 & 0.45 & 19.23 & 18.18 & 9.18 & 11.29 \\ \bottomrule
\end{tabular}%
}
\caption{Error rates, incorrect choice distributions, and local negation confusion rates for the \textbf{Qwen3} family under zero-shot, few-shot, and SFT conditions, evaluated in the \textbf{option-selection setting} using \textbf{detailed instruction.}}
\label{tab:neg_qwen_detail_option}
\end{table*}

\begin{table*}[htbp]
\centering
\resizebox{\textwidth}{!}{%
\begin{tabular}{@{}lllr|rrr|rrrr@{}}
\toprule
\multicolumn{1}{c}{\multirow{2}{*}{\textbf{Model}}} & \multicolumn{1}{c}{\multirow{2}{*}{\textbf{\begin{tabular}[c]{@{}c@{}}Training\\ Setting\end{tabular}}}} & \multicolumn{1}{c}{\multirow{2}{*}{\textbf{N Shot}}} & \multicolumn{1}{c|}{\multirow{2}{*}{\textbf{\begin{tabular}[c]{@{}c@{}}Error \\ Rate\\ (1-acc)\end{tabular}}}} & \multicolumn{3}{c|}{\textbf{Incorrect Choice Distribution}} & \multicolumn{4}{c}{\textbf{Local Negation Confusion Rate}} \\ \cmidrule(l){5-11} 
\multicolumn{1}{c}{} & \multicolumn{1}{c}{} & \multicolumn{1}{c}{} & \multicolumn{1}{c|}{} & \multicolumn{1}{c}{\textbf{\begin{tabular}[c]{@{}c@{}}Local\\ Negation\\ (\%)\end{tabular}}} & \multicolumn{1}{c}{\textbf{\begin{tabular}[c]{@{}c@{}}Contra-\\ diction\\ (\%)\end{tabular}}} & \multicolumn{1}{c|}{\textbf{\begin{tabular}[c]{@{}c@{}}Para-\\ phrase\\ (\%)\end{tabular}}} & \multicolumn{1}{c}{\textbf{\begin{tabular}[c]{@{}c@{}}Relative\\ Clause\\ (\%)\end{tabular}}} & \multicolumn{1}{c}{\textbf{\begin{tabular}[c]{@{}c@{}}Participle\\ Clause\\ (\%)\end{tabular}}} & \multicolumn{1}{c}{\textbf{\begin{tabular}[c]{@{}c@{}}Compound\\ Sentence\\ (\%)\end{tabular}}} & \multicolumn{1}{c}{\textbf{\begin{tabular}[c]{@{}c@{}}Adverbial\\ Clause\\ (\%)\end{tabular}}} \\ \midrule
\multirow{5}{*}{\textbf{Llama-3.1-8B}} & \multirow{4}{*}{\textbf{baseline}} & \textbf{zero-shot} & 0.536 & 76.92 & 17.60 & 5.47 & 25.32 & 30.19 & 60.20 & 55.16 \\
 &  & \textbf{1-shot} & 0.501 & 80.06 & 16.14 & 3.80 & 24.68 & 29.22 & 57.14 & 55.16 \\
 &  & \textbf{5-shot} & 0.391 & 83.77 & 14.20 & 2.03 & 21.79 & 23.05 & 43.54 & 47.10 \\
 &  & \textbf{10-shot} & 0.356 & 83.07 & 15.14 & 1.78 & 19.23 & 21.43 & 38.78 & 42.90 \\ \cmidrule(l){2-11} 
 & \textbf{after SFT} & \textbf{zero-shot} & 0.226 & 88.07 & 10.18 & 1.75 & 10.58 & 15.91 & 24.15 & 31.61 \\ \midrule
\multirow{5}{*}{\textbf{\begin{tabular}[c]{@{}l@{}}Llama-3.1-8B-\\ Instruct\end{tabular}}} & \multirow{4}{*}{\textbf{baseline}} & \textbf{zero-shot} & 0.536 & 73.82 & 23.96 & 2.22 & 25.00 & 31.82 & 58.50 & 48.71 \\
 &  & \textbf{1-shot} & 0.457 & 79.51 & 19.10 & 1.39 & 24.36 & 28.57 & 48.98 & 48.39 \\
 &  & \textbf{5-shot} & 0.367 & 85.31 & 14.25 & 0.43 & 18.91 & 22.40 & 40.14 & 48.06 \\
 &  & \textbf{10-shot} & 0.332 & 85.17 & 14.35 & 0.48 & 20.51 & 20.78 & 32.99 & 42.26 \\ \cmidrule(l){2-11} 
 & \textbf{after SFT} & \textbf{zero-shot} & 0.229 & 86.51 & 12.46 & 1.04 & 10.90 & 14.29 & 23.13 & 33.55 \\ \midrule
\multirow{5}{*}{\textbf{Llama-3.2-1B}} & \multirow{4}{*}{\textbf{baseline}} & \textbf{zero-shot} & 0.591 & 71.01 & 24.03 & 4.97 & 21.79 & 29.55 & 63.95 & 58.71 \\
 &  & \textbf{1-shot} & 0.570 & 72.32 & 23.37 & 4.31 & 25.32 & 32.47 & 63.95 & 49.35 \\
 &  & \textbf{5-shot} & 0.514 & 75.77 & 20.52 & 3.70 & 20.83 & 27.27 & 63.95 & 49.68 \\
 &  & \textbf{10-shot} & 0.468 & 78.14 & 18.64 & 3.22 & 19.87 & 25.97 & 59.86 & 46.13 \\ \cmidrule(l){2-11} 
 & \textbf{after SFT} & \textbf{zero-shot} & 0.250 & 76.83 & 21.59 & 1.59 & 13.46 & 14.61 & 24.15 & 27.10 \\ \midrule
\multirow{5}{*}{\textbf{\begin{tabular}[c]{@{}l@{}}Llama-3.2-1B-\\ Instruct\end{tabular}}} & \multirow{4}{*}{\textbf{baseline}} & \textbf{zero-shot} & 0.569 & 68.48 & 27.89 & 3.63 & 25.00 & 30.84 & 58.16 & 47.42 \\
 &  & \textbf{1-shot} & 0.526 & 69.83 & 27.75 & 2.41 & 25.32 & 26.95 & 54.76 & 45.16 \\
 &  & \textbf{5-shot} & 0.472 & 70.92 & 25.38 & 3.70 & 21.15 & 24.68 & 51.36 & 41.61 \\
 &  & \textbf{10-shot} & 0.447 & 72.70 & 23.94 & 3.37 & 22.12 & 22.40 & 47.96 & 42.26 \\ \cmidrule(l){2-11} 
 & \textbf{after SFT} & \textbf{zero-shot} & 0.239 & 81.06 & 17.94 & 1.00 & 13.14 & 15.26 & 21.43 & 30.00 \\ \midrule
\multirow{5}{*}{\textbf{Llama-3.2-3B}} & \multirow{4}{*}{\textbf{baseline}} & \textbf{zero-shot} & 0.565 & 74.58 & 19.94 & 5.48 & 24.36 & 31.49 & 62.93 & 55.81 \\
 &  & \textbf{1-shot} & 0.538 & 78.17 & 18.88 & 2.95 & 31.09 & 31.82 & 60.88 & 50.32 \\
 &  & \textbf{5-shot} & 0.429 & 79.85 & 17.56 & 2.59 & 22.44 & 27.27 & 47.96 & 44.19 \\
 &  & \textbf{10-shot} & 0.413 & 81.96 & 15.16 & 2.88 & 23.72 & 26.95 & 45.92 & 43.55 \\ \cmidrule(l){2-11} 
 & \textbf{after SFT} & \textbf{zero-shot} & 0.244 & 84.36 & 14.33 & 1.30 & 10.58 & 16.88 & 24.83 & 32.58 \\ \midrule
\multirow{5}{*}{\textbf{\begin{tabular}[c]{@{}l@{}}Llama-3.2-3B-\\ Instruct\end{tabular}}} & \multirow{4}{*}{\textbf{baseline}} & \textbf{zero-shot} & 0.548 & 76.70 & 21.71 & 1.59 & 27.88 & 33.12 & 54.42 & 58.39 \\
 &  & \textbf{1-shot} & 0.536 & 76.33 & 22.34 & 1.33 & 31.41 & 31.17 & 52.72 & 53.87 \\
 &  & \textbf{5-shot} & 0.448 & 78.76 & 20.18 & 1.06 & 24.68 & 26.95 & 40.82 & 53.23 \\
 &  & \textbf{10-shot} & 0.414 & 79.12 & 19.73 & 1.15 & 24.04 & 24.35 & 37.41 & 49.35 \\ \cmidrule(l){2-11} 
 & \textbf{after SFT} & \textbf{zero-shot} & 0.205 & 84.50 & 13.18 & 2.33 & 8.33 & 10.06 & 25.17 & 28.06 \\ \bottomrule
\end{tabular}%
}
\caption{Error rates, incorrect choice distributions, and local negation confusion rates for the \textbf{Llama3} family under zero-shot, few-shot, and SFT conditions, evaluated in the \textbf{completion-based setting} using \textbf{definition instruction.}}
\label{tab:neg_llama_def_completion}
\end{table*}

\begin{table*}[htbp]
\centering
\resizebox{\textwidth}{!}{%
\begin{tabular}{@{}lllr|rrr|rrrr@{}}
\toprule
\multicolumn{1}{c}{\multirow{2}{*}{\textbf{Model}}} & \multicolumn{1}{c}{\multirow{2}{*}{\textbf{\begin{tabular}[c]{@{}c@{}}Training\\ Setting\end{tabular}}}} & \multicolumn{1}{c}{\multirow{2}{*}{\textbf{N Shot}}} & \multicolumn{1}{c|}{\multirow{2}{*}{\textbf{\begin{tabular}[c]{@{}c@{}}Error \\ Rate\\ (1-acc)\end{tabular}}}} & \multicolumn{3}{c|}{\textbf{Incorrect Choice Distribution}} & \multicolumn{4}{c}{\textbf{Local Negation Confusion Rate}} \\ \cmidrule(l){5-11} 
\multicolumn{1}{c}{} & \multicolumn{1}{c}{} & \multicolumn{1}{c}{} & \multicolumn{1}{c|}{} & \multicolumn{1}{c}{\textbf{\begin{tabular}[c]{@{}c@{}}Local\\ Negation\\ (\%)\end{tabular}}} & \multicolumn{1}{c}{\textbf{\begin{tabular}[c]{@{}c@{}}Contra-\\ diction\\ (\%)\end{tabular}}} & \multicolumn{1}{c|}{\textbf{\begin{tabular}[c]{@{}c@{}}Para-\\ phrase\\ (\%)\end{tabular}}} & \multicolumn{1}{c}{\textbf{\begin{tabular}[c]{@{}c@{}}Relative\\ Clause\\ (\%)\end{tabular}}} & \multicolumn{1}{c}{\textbf{\begin{tabular}[c]{@{}c@{}}Participle\\ Clause\\ (\%)\end{tabular}}} & \multicolumn{1}{c}{\textbf{\begin{tabular}[c]{@{}c@{}}Compound\\ Sentence\\ (\%)\end{tabular}}} & \multicolumn{1}{c}{\textbf{\begin{tabular}[c]{@{}c@{}}Adverbial\\ Clause\\ (\%)\end{tabular}}} \\ \midrule
\multirow{5}{*}{\textbf{Llama-3.1-8B}} & \multirow{4}{*}{\textbf{baseline}} & \textbf{zero-shot} & 0.526 & 62.29 & 8.14 & 29.56 & 28.21 & 24.68 & 54.42 & 28.71 \\
 &  & \textbf{1-shot} & 0.493 & 70.42 & 6.11 & 23.47 & 36.22 & 30.52 & 47.28 & 29.68 \\
 &  & \textbf{5-shot} & 0.422 & 84.59 & 4.32 & 11.09 & 41.35 & 26.95 & 49.32 & 30.00 \\
 &  & \textbf{10-shot} & 0.372 & 88.27 & 4.48 & 7.25 & 36.86 & 25.32 & 40.48 & 32.90 \\ \cmidrule(l){2-11} 
 & \textbf{after SFT} & \textbf{zero-shot} & 0.441 & 62.77 & 17.63 & 19.60 & 27.24 & 20.45 & 41.16 & 25.81 \\ \midrule
\multirow{5}{*}{\textbf{\begin{tabular}[c]{@{}l@{}}Llama-3.1-8B-\\ Instruct\end{tabular}}} & \multirow{4}{*}{\textbf{baseline}} & \textbf{zero-shot} & 0.462 & 70.67 & 18.52 & 10.81 & 27.56 & 24.68 & 53.40 & 30.00 \\
 &  & \textbf{1-shot} & 0.404 & 70.53 & 18.07 & 11.39 & 26.92 & 23.70 & 41.84 & 25.48 \\
 &  & \textbf{5-shot} & 0.332 & 81.62 & 11.22 & 7.16 & 25.64 & 18.83 & 40.48 & 27.42 \\
 &  & \textbf{10-shot} & 0.325 & 85.85 & 9.27 & 4.88 & 28.85 & 19.81 & 37.41 & 29.35 \\ \cmidrule(l){2-11} 
 & \textbf{after SFT} & \textbf{zero-shot} & 0.356 & 71.94 & 15.37 & 12.69 & 25.00 & 19.81 & 44.56 & 17.10 \\ \midrule
\multirow{5}{*}{\textbf{Llama-3.2-1B}} & \multirow{4}{*}{\textbf{baseline}} & \textbf{zero-shot} & 0.741 & 32.44 & 34.69 & 32.87 & 22.76 & 27.92 & 25.51 & 22.90 \\
 &  & \textbf{1-shot} & 0.719 & 30.35 & 33.44 & 36.20 & 20.83 & 20.78 & 25.85 & 22.58 \\
 &  & \textbf{5-shot} & 0.751 & 33.16 & 34.85 & 32.00 & 24.36 & 24.68 & 27.89 & 25.81 \\
 &  & \textbf{10-shot} & 0.738 & 33.08 & 33.83 & 33.08 & 25.00 & 25.65 & 25.85 & 24.19 \\ \cmidrule(l){2-11} 
 & \textbf{after SFT} & \textbf{zero-shot} & 0.741 & 32.44 & 34.69 & 32.87 & 22.76 & 27.92 & 25.51 & 22.90 \\ \midrule
\multirow{5}{*}{\textbf{\begin{tabular}[c]{@{}l@{}}Llama-3.2-1B-\\ Instruct\end{tabular}}} & \multirow{4}{*}{\textbf{baseline}} & \textbf{zero-shot} & 0.699 & 39.68 & 27.89 & 32.43 & 26.28 & 34.09 & 29.59 & 24.52 \\
 &  & \textbf{1-shot} & 0.608 & 53.59 & 19.17 & 27.25 & 25.32 & 39.29 & 41.50 & 28.71 \\
 &  & \textbf{5-shot} & 0.645 & 47.36 & 22.88 & 29.77 & 27.56 & 38.64 & 31.97 & 27.74 \\
 &  & \textbf{10-shot} & 0.678 & 43.98 & 25.26 & 30.76 & 28.21 & 33.77 & 33.33 & 27.74 \\ \cmidrule(l){2-11} 
 & \textbf{after SFT} & \textbf{zero-shot} & 0.714 & 35.78 & 29.22 & 35.00 & 23.72 & 30.52 & 26.53 & 24.52 \\ \midrule
\multirow{5}{*}{\textbf{Llama-3.2-3B}} & \multirow{4}{*}{\textbf{baseline}} & \textbf{zero-shot} & 0.638 & 50.37 & 26.37 & 23.26 & 23.08 & 35.39 & 40.82 & 33.55 \\
 &  & \textbf{1-shot} & 0.555 & 68.43 & 14.00 & 17.57 & 33.33 & 34.74 & 46.94 & 41.94 \\
 &  & \textbf{5-shot} & 0.527 & 79.70 & 10.83 & 9.47 & 38.14 & 37.66 & 51.02 & 46.77 \\
 &  & \textbf{10-shot} & 0.516 & 83.23 & 10.15 & 6.62 & 40.38 & 36.36 & 49.66 & 50.65 \\ \cmidrule(l){2-11} 
 & \textbf{after SFT} & \textbf{zero-shot} & 0.615 & 46.97 & 28.77 & 24.26 & 22.44 & 30.52 & 35.71 & 30.65 \\ \midrule
\multirow{5}{*}{\textbf{\begin{tabular}[c]{@{}l@{}}Llama-3.2-3B-\\ Instruct\end{tabular}}} & \multirow{4}{*}{\textbf{baseline}} & \textbf{zero-shot} & 0.488 & 80.00 & 7.97 & 12.03 & 27.24 & 35.39 & 64.29 & 35.16 \\
 &  & \textbf{1-shot} & 0.458 & 77.68 & 9.86 & 12.46 & 24.36 & 28.25 & 60.88 & 34.52 \\
 &  & \textbf{5-shot} & 0.502 & 74.25 & 8.85 & 16.90 & 29.49 & 27.27 & 67.35 & 30.97 \\
 &  & \textbf{10-shot} & 0.498 & 74.04 & 8.44 & 17.52 & 31.09 & 26.62 & 64.29 & 31.29 \\ \cmidrule(l){2-11} 
 & \textbf{after SFT} & \textbf{zero-shot} & 0.349 & 70.45 & 6.59 & 22.95 & 15.71 & 16.88 & 45.58 & 24.19 \\ \bottomrule
\end{tabular}%
}
\caption{Error rates, incorrect choice distributions, and local negation confusion rates for the \textbf{Llama3} family under zero-shot, few-shot, and SFT conditions, evaluated in the \textbf{option-selection setting} using \textbf{definition instruction.}}
\label{tab:neg_llama_def_option}
\end{table*}

\begin{table*}[htbp]
\centering
\resizebox{\textwidth}{!}{%
\begin{tabular}{@{}lllr|rrr|rrrr@{}}
\toprule
\multicolumn{1}{c}{\multirow{2}{*}{\textbf{Model}}} & \multicolumn{1}{c}{\multirow{2}{*}{\textbf{\begin{tabular}[c]{@{}c@{}}Training\\ Setting\end{tabular}}}} & \multicolumn{1}{c}{\multirow{2}{*}{\textbf{N Shot}}} & \multicolumn{1}{c|}{\multirow{2}{*}{\textbf{\begin{tabular}[c]{@{}c@{}}Error \\ Rate\\ (1-acc)\end{tabular}}}} & \multicolumn{3}{c|}{\textbf{Incorrect Choice Distribution}} & \multicolumn{4}{c}{\textbf{Local Negation Confusion Rate}} \\ \cmidrule(l){5-11} 
\multicolumn{1}{c}{} & \multicolumn{1}{c}{} & \multicolumn{1}{c}{} & \multicolumn{1}{c|}{} & \multicolumn{1}{c}{\textbf{\begin{tabular}[c]{@{}c@{}}Local\\ Negation\\ (\%)\end{tabular}}} & \multicolumn{1}{c}{\textbf{\begin{tabular}[c]{@{}c@{}}Contra-\\ diction\\ (\%)\end{tabular}}} & \multicolumn{1}{c|}{\textbf{\begin{tabular}[c]{@{}c@{}}Para-\\ phrase\\ (\%)\end{tabular}}} & \multicolumn{1}{c}{\textbf{\begin{tabular}[c]{@{}c@{}}Relative\\ Clause\\ (\%)\end{tabular}}} & \multicolumn{1}{c}{\textbf{\begin{tabular}[c]{@{}c@{}}Participle\\ Clause\\ (\%)\end{tabular}}} & \multicolumn{1}{c}{\textbf{\begin{tabular}[c]{@{}c@{}}Compound\\ Sentence\\ (\%)\end{tabular}}} & \multicolumn{1}{c}{\textbf{\begin{tabular}[c]{@{}c@{}}Adverbial\\ Clause\\ (\%)\end{tabular}}} \\ \midrule
\multirow{5}{*}{\textbf{Llama-3.1-8B}} & \multirow{4}{*}{\textbf{baseline}} & \textbf{zero-shot} & 0.549 & 73.99 & 19.08 & 6.94 & 26.60 & 28.57 & 61.56 & 51.61 \\
 &  & \textbf{1-shot} & 0.516 & 75.42 & 20.28 & 4.30 & 25.96 & 25.32 & 58.50 & 51.61 \\
 &  & \textbf{5-shot} & 0.424 & 79.07 & 17.76 & 3.18 & 22.12 & 23.05 & 47.62 & 46.13 \\
 &  & \textbf{10-shot} & 0.381 & 81.08 & 16.63 & 2.29 & 20.51 & 22.08 & 41.50 & 43.87 \\ \cmidrule(l){2-11} 
 & \textbf{after SFT} & \textbf{zero-shot} & 0.198 & 88.00 & 11.60 & 0.40 & 9.62 & 13.31 & 21.09 & 28.06 \\ \midrule
\multirow{5}{*}{\textbf{\begin{tabular}[c]{@{}l@{}}Llama-3.1-8B-\\ Instruct\end{tabular}}} & \multirow{4}{*}{\textbf{baseline}} & \textbf{zero-shot} & 0.468 & 71.86 & 26.44 & 1.69 & 23.08 & 29.87 & 47.62 & 38.71 \\
 &  & \textbf{1-shot} & 0.408 & 78.99 & 19.65 & 1.36 & 25.00 & 26.95 & 43.20 & 38.06 \\
 &  & \textbf{5-shot} & 0.347 & 84.70 & 14.84 & 0.46 & 19.55 & 23.05 & 37.76 & 41.29 \\
 &  & \textbf{10-shot} & 0.316 & 84.46 & 15.04 & 0.50 & 21.15 & 19.48 & 31.29 & 38.39 \\ \cmidrule(l){2-11} 
 & \textbf{after SFT} & \textbf{zero-shot} & 0.211 & 87.97 & 10.90 & 1.13 & 8.97 & 13.31 & 21.09 & 33.23 \\ \midrule
\multirow{5}{*}{\textbf{Llama-3.2-1B}} & \multirow{4}{*}{\textbf{baseline}} & \textbf{zero-shot} & 0.597 & 69.85 & 24.70 & 5.44 & 24.04 & 32.47 & 67.01 & 49.68 \\
 &  & \textbf{1-shot} & 0.565 & 72.75 & 22.89 & 4.35 & 26.28 & 30.52 & 64.63 & 49.03 \\
 &  & \textbf{5-shot} & 0.510 & 76.21 & 20.22 & 3.58 & 20.83 & 26.95 & 64.97 & 48.71 \\
 &  & \textbf{10-shot} & 0.470 & 78.08 & 18.55 & 3.37 & 20.51 & 25.65 & 60.20 & 46.13 \\ \cmidrule(l){2-11} 
 & \textbf{after SFT} & \textbf{zero-shot} & 0.253 & 80.25 & 18.50 & 1.25 & 14.42 & 15.91 & 24.83 & 28.71 \\ \midrule
\multirow{5}{*}{\textbf{\begin{tabular}[c]{@{}l@{}}Llama-3.2-1B-\\ Instruct\end{tabular}}} & \multirow{4}{*}{\textbf{baseline}} & \textbf{zero-shot} & 0.584 & 68.89 & 27.72 & 3.40 & 27.56 & 34.42 & 57.48 & 47.10 \\
 &  & \textbf{1-shot} & 0.528 & 69.67 & 27.18 & 3.15 & 24.36 & 27.60 & 53.06 & 47.42 \\
 &  & \textbf{5-shot} & 0.478 & 71.31 & 24.71 & 3.98 & 20.83 & 24.03 & 53.40 & 43.23 \\
 &  & \textbf{10-shot} & 0.453 & 72.33 & 23.99 & 3.68 & 21.79 & 21.43 & 50.00 & 42.58 \\ \cmidrule(l){2-11} 
 & \textbf{after SFT} & \textbf{zero-shot} & 0.297 & 79.41 & 18.98 & 1.60 & 14.74 & 18.83 & 33.33 & 30.65 \\ \midrule
\multirow{5}{*}{\textbf{Llama-3.2-3B}} & \multirow{4}{*}{\textbf{baseline}} & \textbf{zero-shot} & 0.553 & 73.17 & 19.66 & 7.17 & 23.72 & 35.06 & 60.20 & 48.71 \\
 &  & \textbf{1-shot} & 0.522 & 77.05 & 19.30 & 3.65 & 27.88 & 30.52 & 59.86 & 48.39 \\
 &  & \textbf{5-shot} & 0.427 & 80.67 & 16.54 & 2.79 & 22.12 & 28.25 & 45.92 & 46.13 \\
 &  & \textbf{10-shot} & 0.413 & 82.73 & 14.40 & 2.88 & 25.00 & 27.60 & 44.56 & 44.19 \\ \cmidrule(l){2-11} 
 & \textbf{after SFT} & \textbf{zero-shot} & 0.240 & 81.13 & 16.89 & 1.99 & 9.29 & 14.94 & 24.49 & 31.61 \\ \midrule
\multirow{5}{*}{\textbf{\begin{tabular}[c]{@{}l@{}}Llama-3.2-3B-\\ Instruct\end{tabular}}} & \multirow{4}{*}{\textbf{baseline}} & \textbf{zero-shot} & 0.512 & 76.63 & 21.98 & 1.39 & 25.64 & 34.09 & 47.62 & 54.84 \\
 &  & \textbf{1-shot} & 0.527 & 77.26 & 21.54 & 1.20 & 30.45 & 30.84 & 49.66 & 57.10 \\
 &  & \textbf{5-shot} & 0.456 & 80.35 & 18.61 & 1.04 & 24.36 & 28.57 & 42.86 & 55.48 \\
 &  & \textbf{10-shot} & 0.424 & 81.68 & 17.20 & 1.12 & 24.68 & 27.27 & 38.78 & 52.26 \\ \cmidrule(l){2-11} 
 & \textbf{after SFT} & \textbf{zero-shot} & 0.209 & 82.89 & 15.21 & 1.90 & 9.62 & 12.34 & 26.87 & 22.90 \\ \bottomrule
\end{tabular}%
}
\caption{Error rates, incorrect choice distributions, and local negation confusion rates for the \textbf{Llama3} family under zero-shot, few-shot, and SFT conditions, evaluated in the \textbf{completion-based setting} using \textbf{detailed instruction.}}
\label{tab:neg_llama_detail_completion}
\end{table*}

\begin{table*}[htbp]
\centering
\resizebox{\textwidth}{!}{%
\begin{tabular}{@{}lllr|rrr|rrrr@{}}
\toprule
\multicolumn{1}{c}{\multirow{2}{*}{\textbf{Model}}} & \multicolumn{1}{c}{\multirow{2}{*}{\textbf{\begin{tabular}[c]{@{}c@{}}Training\\ Setting\end{tabular}}}} & \multicolumn{1}{c}{\multirow{2}{*}{\textbf{N Shot}}} & \multicolumn{1}{c|}{\multirow{2}{*}{\textbf{\begin{tabular}[c]{@{}c@{}}Error \\ Rate\\ (1-acc)\end{tabular}}}} & \multicolumn{3}{c|}{\textbf{Incorrect Choice Distribution}} & \multicolumn{4}{c}{\textbf{Local Negation Confusion Rate}} \\ \cmidrule(l){5-11} 
\multicolumn{1}{c}{} & \multicolumn{1}{c}{} & \multicolumn{1}{c}{} & \multicolumn{1}{c|}{} & \multicolumn{1}{c}{\textbf{\begin{tabular}[c]{@{}c@{}}Local\\ Negation\\ (\%)\end{tabular}}} & \multicolumn{1}{c}{\textbf{\begin{tabular}[c]{@{}c@{}}Contra-\\ diction\\ (\%)\end{tabular}}} & \multicolumn{1}{c|}{\textbf{\begin{tabular}[c]{@{}c@{}}Para-\\ phrase\\ (\%)\end{tabular}}} & \multicolumn{1}{c}{\textbf{\begin{tabular}[c]{@{}c@{}}Relative\\ Clause\\ (\%)\end{tabular}}} & \multicolumn{1}{c}{\textbf{\begin{tabular}[c]{@{}c@{}}Participle\\ Clause\\ (\%)\end{tabular}}} & \multicolumn{1}{c}{\textbf{\begin{tabular}[c]{@{}c@{}}Compound\\ Sentence\\ (\%)\end{tabular}}} & \multicolumn{1}{c}{\textbf{\begin{tabular}[c]{@{}c@{}}Adverbial\\ Clause\\ (\%)\end{tabular}}} \\ \midrule
\multirow{5}{*}{\textbf{Llama-3.1-8B}} & \multirow{4}{*}{\textbf{baseline}} & \textbf{zero-shot} & 0.569 & 66.43 & 8.64 & 24.93 & 32.69 & 31.49 & 59.18 & 33.55 \\
 &  & \textbf{1-shot} & 0.519 & 73.85 & 5.20 & 20.95 & 40.71 & 35.71 & 48.30 & 33.55 \\
 &  & \textbf{5-shot} & 0.448 & 85.84 & 3.89 & 10.27 & 43.27 & 31.82 & 49.32 & 34.52 \\
 &  & \textbf{10-shot} & 0.385 & 88.25 & 4.12 & 7.63 & 38.14 & 27.92 & 40.48 & 33.55 \\ \cmidrule(l){2-11} 
 & \textbf{after SFT} & \textbf{zero-shot} & 0.328 & 69.81 & 18.84 & 11.35 & 22.76 & 20.78 & 35.71 & 15.81 \\ \midrule
\multirow{5}{*}{\textbf{\begin{tabular}[c]{@{}l@{}}Llama-3.1-8B-\\ Instruct\end{tabular}}} & \multirow{4}{*}{\textbf{baseline}} & \textbf{zero-shot} & 0.486 & 71.45 & 14.85 & 13.70 & 33.01 & 25.97 & 56.12 & 29.03 \\
 &  & \textbf{1-shot} & 0.402 & 67.26 & 17.16 & 15.58 & 28.85 & 21.10 & 36.39 & 25.48 \\
 &  & \textbf{5-shot} & 0.353 & 77.53 & 11.24 & 11.24 & 27.56 & 19.81 & 39.12 & 26.77 \\
 &  & \textbf{10-shot} & 0.349 & 83.41 & 9.09 & 7.50 & 31.41 & 19.48 & 39.12 & 30.32 \\ \cmidrule(l){2-11} 
 & \textbf{after SFT} & \textbf{zero-shot} & 0.270 & 65.69 & 21.70 & 12.61 & 14.74 & 14.61 & 29.25 & 15.16 \\ \midrule
\multirow{5}{*}{\textbf{Llama-3.2-1B}} & \multirow{4}{*}{\textbf{baseline}} & \textbf{zero-shot} & 0.741 & 32.44 & 34.69 & 32.87 & 22.76 & 27.92 & 25.51 & 22.90 \\
 &  & \textbf{1-shot} & 0.726 & 29.29 & 33.66 & 37.05 & 21.47 & 18.83 & 24.49 & 22.90 \\
 &  & \textbf{5-shot} & 0.744 & 33.05 & 34.75 & 32.20 & 24.04 & 26.30 & 25.85 & 25.16 \\
 &  & \textbf{10-shot} & 0.742 & 32.48 & 34.19 & 33.33 & 23.72 & 25.00 & 25.85 & 24.84 \\ \cmidrule(l){2-11} 
 & \textbf{after SFT} & \textbf{zero-shot} & 0.741 & 32.44 & 34.69 & 32.87 & 22.76 & 27.92 & 25.51 & 22.90 \\ \midrule
\multirow{5}{*}{\textbf{\begin{tabular}[c]{@{}l@{}}Llama-3.2-1B-\\ Instruct\end{tabular}}} & \multirow{4}{*}{\textbf{baseline}} & \textbf{zero-shot} & 0.700 & 40.43 & 27.86 & 31.71 & 25.96 & 35.39 & 29.59 & 25.81 \\
 &  & \textbf{1-shot} & 0.638 & 50.31 & 22.36 & 27.33 & 24.68 & 38.31 & 40.14 & 29.68 \\
 &  & \textbf{5-shot} & 0.658 & 46.63 & 23.37 & 30.00 & 27.24 & 38.64 & 31.63 & 29.03 \\
 &  & \textbf{10-shot} & 0.681 & 42.03 & 25.73 & 32.25 & 27.88 & 33.12 & 31.97 & 25.16 \\ \cmidrule(l){2-11} 
 & \textbf{after SFT} & \textbf{zero-shot} & 0.738 & 32.69 & 33.23 & 34.09 & 22.44 & 27.92 & 25.85 & 23.23 \\ \midrule
\multirow{5}{*}{\textbf{Llama-3.2-3B}} & \multirow{4}{*}{\textbf{baseline}} & \textbf{zero-shot} & 0.626 & 58.81 & 18.76 & 22.43 & 27.24 & 41.56 & 47.96 & 35.48 \\
 &  & \textbf{1-shot} & 0.581 & 67.21 & 13.80 & 18.99 & 33.65 & 39.61 & 44.90 & 42.90 \\
 &  & \textbf{5-shot} & 0.526 & 79.94 & 10.71 & 9.35 & 37.18 & 38.64 & 51.02 & 46.77 \\
 &  & \textbf{10-shot} & 0.512 & 83.41 & 9.30 & 7.29 & 38.78 & 36.69 & 50.34 & 50.32 \\ \cmidrule(l){2-11} 
 & \textbf{after SFT} & \textbf{zero-shot} & 0.636 & 47.51 & 26.68 & 25.81 & 22.76 & 36.36 & 35.37 & 30.32 \\ \midrule
\multirow{5}{*}{\textbf{\begin{tabular}[c]{@{}l@{}}Llama-3.2-3B-\\ Instruct\end{tabular}}} & \multirow{4}{*}{\textbf{baseline}} & \textbf{zero-shot} & 0.512 & 79.26 & 9.91 & 10.84 & 31.09 & 36.69 & 65.65 & 35.16 \\
 &  & \textbf{1-shot} & 0.466 & 80.07 & 10.22 & 9.71 & 28.53 & 30.19 & 60.88 & 35.16 \\
 &  & \textbf{5-shot} & 0.509 & 76.64 & 8.88 & 14.49 & 30.77 & 29.55 & 67.35 & 34.52 \\
 &  & \textbf{10-shot} & 0.509 & 74.61 & 7.79 & 17.60 & 32.05 & 28.25 & 65.65 & 31.94 \\ \cmidrule(l){2-11} 
 & \textbf{after SFT} & \textbf{zero-shot} & 0.418 & 58.82 & 10.25 & 30.93 & 17.63 & 18.18 & 48.64 & 18.06 \\ \bottomrule
\end{tabular}%
}
\caption{Error rates, incorrect choice distributions, and local negation confusion rates for the \textbf{Llama3} family under zero-shot, few-shot, and SFT conditions, evaluated in the \textbf{option-selection setting} using \textbf{detailed instruction.}}
\label{tab:neg_llama_detail_option}
\end{table*}

\begin{table*}[htbp]
\centering
\resizebox{0.9\textwidth}{!}{%
\begin{tabular}{@{}cccr|rrr|rrrr@{}}
\toprule
\multirow{2}{*}{\textbf{Model}} & \multirow{2}{*}{\textbf{\begin{tabular}[c]{@{}c@{}}Training\\ Setting\end{tabular}}} & \multirow{2}{*}{\textbf{N Shot}} & \multicolumn{1}{c|}{\multirow{2}{*}{\textbf{\begin{tabular}[c]{@{}c@{}}Error \\ Rate\\ (1-acc)\end{tabular}}}} & \multicolumn{3}{c|}{\textbf{Incorrect Choice Distribution}} & \multicolumn{4}{c}{\textbf{Local Negation Confusion Rate}} \\ \cmidrule(l){5-11} 
 &  &  & \multicolumn{1}{c|}{} & \multicolumn{1}{c}{\textbf{\begin{tabular}[c]{@{}c@{}}Local\\ Negation\\ (\%)\end{tabular}}} & \multicolumn{1}{c}{\textbf{\begin{tabular}[c]{@{}c@{}}Contra-\\ diction\\ (\%)\end{tabular}}} & \multicolumn{1}{c|}{\textbf{\begin{tabular}[c]{@{}c@{}}Para-\\ phrase\\ (\%)\end{tabular}}} & \multicolumn{1}{c}{\textbf{\begin{tabular}[c]{@{}c@{}}Relative\\ Clause\\ (\%)\end{tabular}}} & \multicolumn{1}{c}{\textbf{\begin{tabular}[c]{@{}c@{}}Participle\\ Clause\\ (\%)\end{tabular}}} & \multicolumn{1}{c}{\textbf{\begin{tabular}[c]{@{}c@{}}Compound\\ Sentence\\ (\%)\end{tabular}}} & \multicolumn{1}{c}{\textbf{\begin{tabular}[c]{@{}c@{}}Adverbial\\ Clause\\ (\%)\end{tabular}}} \\ \midrule
\multicolumn{1}{l}{\multirow{5}{*}{\textbf{Mistral-7B-v0.3}}} & \multicolumn{1}{l}{\multirow{4}{*}{\textbf{baseline}}} & \multicolumn{1}{l}{\textbf{zero-shot}} & 0.533 & 75.74 & 20.24 & 4.02 & 25.96 & 31.82 & 60.20 & 49.35 \\
\multicolumn{1}{l}{} & \multicolumn{1}{l}{} & \multicolumn{1}{l}{\textbf{1-shot}} & 0.533 & 78.27 & 18.15 & 3.57 & 29.81 & 33.77 & 59.18 & 50.00 \\
\multicolumn{1}{l}{} & \multicolumn{1}{l}{} & \multicolumn{1}{l}{\textbf{5-shot}} & 0.427 & 82.71 & 15.80 & 1.49 & 22.12 & 25.65 & 51.02 & 47.42 \\
\multicolumn{1}{l}{} & \multicolumn{1}{l}{} & \multicolumn{1}{l}{\textbf{10-shot}} & 0.394 & 84.31 & 13.28 & 2.41 & 21.47 & 24.68 & 47.96 & 43.55 \\ \cmidrule(l){2-11} 
\multicolumn{1}{l}{} & \multicolumn{1}{l}{\textbf{after SFT}} & \multicolumn{1}{l}{\textbf{zero-shot}} & 0.217 & 86.81 & 12.09 & 1.10 & 8.97 & 12.66 & 26.53 & 29.68 \\ \midrule
\multicolumn{1}{l}{\multirow{5}{*}{\textbf{\begin{tabular}[c]{@{}l@{}}Mistral-7B-Instruct-\\ v0.3\end{tabular}}}} & \multicolumn{1}{l}{\multirow{4}{*}{\textbf{baseline}}} & \multicolumn{1}{l}{\textbf{zero-shot}} & 0.389 & 65.10 & 34.08 & 0.82 & 16.67 & 18.83 & 29.25 & 39.68 \\
\multicolumn{1}{l}{} & \multicolumn{1}{l}{} & \multicolumn{1}{l}{\textbf{1-shot}} & 0.350 & 71.66 & 27.44 & 0.91 & 16.35 & 18.18 & 33.33 & 35.81 \\
\multicolumn{1}{l}{} & \multicolumn{1}{l}{} & \multicolumn{1}{l}{\textbf{5-shot}} & 0.305 & 72.47 & 27.27 & 0.26 & 12.50 & 15.58 & 28.91 & 34.52 \\
\multicolumn{1}{l}{} & \multicolumn{1}{l}{} & \multicolumn{1}{l}{\textbf{10-shot}} & 0.282 & 73.24 & 25.92 & 0.85 & 12.82 & 13.96 & 26.19 & 32.26 \\ \cmidrule(l){2-11} 
\multicolumn{1}{l}{} & \multicolumn{1}{l}{\textbf{after SFT}} & \multicolumn{1}{l}{\textbf{zero-shot}} & 0.207 & 83.91 & 14.18 & 1.92 & 8.01 & 11.04 & 24.49 & 28.39 \\ \midrule
\multirow{4}{*}{\textbf{\begin{tabular}[c]{@{}c@{}}Mistral-Nemo-\\ Base-2407 (12B)\end{tabular}}} & \multirow{4}{*}{\textbf{baseline}} & \textbf{zero-shot} & 0.540 & 75.62 & 21.44 & 2.94 & 25.64 & 33.44 & 60.20 & 50.00 \\
 &  & \textbf{1-shot} & 0.509 & 76.48 & 20.72 & 2.80 & 27.24 & 30.19 & 56.80 & 47.10 \\
 &  & \textbf{5-shot} & 0.399 & 81.11 & 16.70 & 2.19 & 21.15 & 23.70 & 42.86 & 46.13 \\
 &  & \textbf{10-shot} & 0.346 & 82.11 & 16.74 & 1.15 & 18.91 & 21.75 & 32.65 & 43.87 \\ \midrule
\multirow{4}{*}{\textbf{\begin{tabular}[c]{@{}c@{}}Mistral-Nemo-\\ Instruct-2407 (12B)\end{tabular}}} & \multirow{4}{*}{\textbf{baseline}} & \textbf{zero-shot} & 0.513 & 75.43 & 21.64 & 2.94 & 24.04 & 29.22 & 56.80 & 50.32 \\
 &  & \textbf{1-shot} & 0.468 & 76.95 & 21.36 & 1.69 & 23.08 & 26.30 & 52.38 & 47.42 \\
 &  & \textbf{5-shot} & 0.369 & 78.28 & 20.43 & 1.29 & 19.23 & 20.45 & 38.44 & 41.29 \\
 &  & \textbf{10-shot} & 0.324 & 79.41 & 19.36 & 1.23 & 17.31 & 17.53 & 30.61 & 40.65 \\ \midrule
\multirow{4}{*}{\textbf{\begin{tabular}[c]{@{}c@{}}Mistral-Small-24B-\\ Base-2501\end{tabular}}} & \multirow{4}{*}{\textbf{baseline}} & \textbf{zero-shot} & 0.526 & 72.70 & 21.57 & 5.73 & 24.68 & 29.87 & 60.20 & 43.87 \\
 &  & \textbf{1-shot} & 0.502 & 78.52 & 18.48 & 3.00 & 26.60 & 29.87 & 58.16 & 48.71 \\
 &  & \textbf{5-shot} & 0.394 & 82.70 & 14.69 & 2.62 & 20.51 & 24.35 & 45.24 & 44.84 \\
 &  & \textbf{10-shot} & 0.348 & 83.37 & 14.58 & 2.05 & 17.63 & 21.75 & 37.76 & 42.90 \\ \midrule
\multirow{4}{*}{\textbf{\begin{tabular}[c]{@{}c@{}}Mistral-Small-24B-\\ Instruct-2501\end{tabular}}} & \multirow{4}{*}{\textbf{baseline}} & \textbf{zero-shot} & 0.474 & 78.43 & 19.57 & 2.01 & 25.96 & 29.87 & 53.06 & 45.16 \\
 &  & \textbf{1-shot} & 0.426 & 79.70 & 18.81 & 1.49 & 23.40 & 24.03 & 48.30 & 44.84 \\
 &  & \textbf{5-shot} & 0.355 & 81.43 & 17.67 & 0.89 & 18.91 & 20.78 & 38.10 & 41.61 \\
 &  & \textbf{10-shot} & 0.309 & 83.55 & 15.68 & 0.77 & 16.03 & 18.51 & 31.63 & 40.32 \\ \bottomrule
\end{tabular}%
}
\caption{Error rates, incorrect choice distributions, and local negation confusion rates for the \textbf{Mistral} family under zero-shot, few-shot, and SFT conditions, evaluated in the \textbf{completion-based setting} using \textbf{definition instruction.}}
\label{tab:neg_mistral_def_completion}
\end{table*}

\begin{table*}[htbp]
\centering
\resizebox{0.9\textwidth}{!}{%
\begin{tabular}{@{}cccr|rrr|rrrr@{}}
\toprule
\multirow{2}{*}{\textbf{Model}} & \multirow{2}{*}{\textbf{\begin{tabular}[c]{@{}c@{}}Training\\ Setting\end{tabular}}} & \multirow{2}{*}{\textbf{N Shot}} & \multicolumn{1}{c|}{\multirow{2}{*}{\textbf{\begin{tabular}[c]{@{}c@{}}Error \\ Rate\\ (1-acc)\end{tabular}}}} & \multicolumn{3}{c|}{\textbf{Incorrect Choice Distribution}} & \multicolumn{4}{c}{\textbf{Local Negation Confusion Rate}} \\ \cmidrule(l){5-11} 
 &  &  & \multicolumn{1}{c|}{} & \multicolumn{1}{c}{\textbf{\begin{tabular}[c]{@{}c@{}}Local\\ Negation\\ (\%)\end{tabular}}} & \multicolumn{1}{c}{\textbf{\begin{tabular}[c]{@{}c@{}}Contra-\\ diction\\ (\%)\end{tabular}}} & \multicolumn{1}{c|}{\textbf{\begin{tabular}[c]{@{}c@{}}Para-\\ phrase\\ (\%)\end{tabular}}} & \multicolumn{1}{c}{\textbf{\begin{tabular}[c]{@{}c@{}}Relative\\ Clause\\ (\%)\end{tabular}}} & \multicolumn{1}{c}{\textbf{\begin{tabular}[c]{@{}c@{}}Participle\\ Clause\\ (\%)\end{tabular}}} & \multicolumn{1}{c}{\textbf{\begin{tabular}[c]{@{}c@{}}Compound\\ Sentence\\ (\%)\end{tabular}}} & \multicolumn{1}{c}{\textbf{\begin{tabular}[c]{@{}c@{}}Adverbial\\ Clause\\ (\%)\end{tabular}}} \\ \midrule
\multicolumn{1}{l}{\multirow{5}{*}{\textbf{Mistral-7B-v0.3}}} & \multicolumn{1}{l}{\multirow{4}{*}{\textbf{baseline}}} & \multicolumn{1}{l}{\textbf{zero-shot}} & 0.516 & 51.23 & 17.08 & 31.69 & 24.68 & 20.78 & 37.41 & 26.45 \\
\multicolumn{1}{l}{} & \multicolumn{1}{l}{} & \multicolumn{1}{l}{\textbf{1-shot}} & 0.521 & 53.42 & 10.20 & 36.38 & 27.88 & 23.38 & 36.73 & 27.10 \\
\multicolumn{1}{l}{} & \multicolumn{1}{l}{} & \multicolumn{1}{l}{\textbf{5-shot}} & 0.370 & 73.61 & 7.73 & 18.67 & 25.96 & 21.43 & 31.63 & 33.23 \\
\multicolumn{1}{l}{} & \multicolumn{1}{l}{} & \multicolumn{1}{l}{\textbf{10-shot}} & 0.336 & 73.82 & 7.78 & 18.40 & 25.00 & 19.81 & 26.19 & 31.29 \\ \cmidrule(l){2-11} 
\multicolumn{1}{l}{} & \multicolumn{1}{l}{\textbf{after SFT}} & \multicolumn{1}{l}{\textbf{zero-shot}} & 0.520 & 54.57 & 24.85 & 20.58 & 27.88 & 29.87 & 33.67 & 25.81 \\ \midrule
\multicolumn{1}{l}{\multirow{5}{*}{\textbf{\begin{tabular}[c]{@{}l@{}}Mistral-7B-Instruct-\\ v0.3\end{tabular}}}} & \multicolumn{1}{l}{\multirow{4}{*}{\textbf{baseline}}} & \multicolumn{1}{l}{\textbf{zero-shot}} & 0.343 & 77.37 & 14.32 & 8.31 & 26.28 & 20.78 & 34.35 & 28.39 \\
\multicolumn{1}{l}{} & \multicolumn{1}{l}{} & \multicolumn{1}{l}{\textbf{1-shot}} & 0.355 & 69.42 & 20.09 & 10.49 & 28.53 & 23.05 & 29.25 & 20.97 \\
\multicolumn{1}{l}{} & \multicolumn{1}{l}{} & \multicolumn{1}{l}{\textbf{5-shot}} & 0.347 & 71.00 & 17.81 & 11.19 & 25.64 & 22.08 & 31.29 & 22.90 \\
\multicolumn{1}{l}{} & \multicolumn{1}{l}{} & \multicolumn{1}{l}{\textbf{10-shot}} & 0.351 & 72.46 & 20.77 & 6.77 & 24.68 & 22.08 & 32.65 & 25.81 \\ \cmidrule(l){2-11} 
\multicolumn{1}{l}{} & \multicolumn{1}{l}{\textbf{after SFT}} & \multicolumn{1}{l}{\textbf{zero-shot}} & 0.308 & 77.58 & 17.01 & 5.41 & 23.72 & 19.48 & 35.37 & 20.32 \\ \midrule
\multirow{4}{*}{\textbf{\begin{tabular}[c]{@{}c@{}}Mistral-Nemo-\\ Base-2407 (12B)\end{tabular}}} & \multirow{4}{*}{\textbf{baseline}} & \textbf{zero-shot} & 0.588 & 64.10 & 15.25 & 20.65 & 35.90 & 33.44 & 49.66 & 36.77 \\
 &  & \textbf{1-shot} & 0.488 & 83.41 & 10.24 & 6.34 & 39.10 & 33.12 & 54.42 & 41.61 \\
 &  & \textbf{5-shot} & 0.438 & 89.49 & 6.34 & 4.17 & 39.10 & 32.47 & 48.98 & 41.29 \\
 &  & \textbf{10-shot} & 0.400 & 89.88 & 6.15 & 3.97 & 33.97 & 29.55 & 40.14 & 44.52 \\ \midrule
\multirow{4}{*}{\textbf{\begin{tabular}[c]{@{}c@{}}Mistral-Nemo-\\ Instruct-2407 (12B)\end{tabular}}} & \multirow{4}{*}{\textbf{baseline}} & \textbf{zero-shot} & 0.396 & 89.38 & 7.82 & 2.81 & 33.33 & 25.32 & 52.38 & 35.48 \\
 &  & \textbf{1-shot} & 0.368 & 89.22 & 8.62 & 2.16 & 30.77 & 24.68 & 46.94 & 33.55 \\
 &  & \textbf{5-shot} & 0.345 & 89.89 & 8.51 & 1.61 & 32.05 & 25.65 & 39.80 & 30.65 \\
 &  & \textbf{10-shot} & 0.338 & 90.85 & 7.75 & 1.41 & 32.69 & 29.22 & 35.37 & 29.35 \\ \midrule
\multirow{4}{*}{\textbf{\begin{tabular}[c]{@{}c@{}}Mistral-Small-24B-\\ Base-2501\end{tabular}}} & \multirow{4}{*}{\textbf{baseline}} & \textbf{zero-shot} & 0.432 & 62.20 & 18.72 & 19.08 & 29.81 & 21.75 & 39.12 & 20.65 \\
 &  & \textbf{1-shot} & 0.381 & 73.75 & 19.38 & 6.88 & 32.69 & 25.65 & 32.65 & 24.84 \\
 &  & \textbf{5-shot} & 0.274 & 85.80 & 9.86 & 4.35 & 25.32 & 20.13 & 21.77 & 29.35 \\
 &  & \textbf{10-shot} & 0.219 & 90.58 & 6.88 & 2.54 & 23.40 & 17.86 & 9.86 & 30.00 \\ \midrule
\multirow{4}{*}{\textbf{\begin{tabular}[c]{@{}c@{}}Mistral-Small-24B-\\ Instruct-2501\end{tabular}}} & \multirow{4}{*}{\textbf{baseline}} & \textbf{zero-shot} & 0.314 & 78.03 & 15.91 & 6.06 & 27.24 & 19.81 & 31.63 & 22.58 \\
 &  & \textbf{1-shot} & 0.290 & 81.15 & 14.48 & 4.37 & 30.77 & 18.83 & 27.89 & 19.68 \\
 &  & \textbf{5-shot} & 0.257 & 86.11 & 11.42 & 2.47 & 30.13 & 23.38 & 17.01 & 20.32 \\
 &  & \textbf{10-shot} & 0.217 & 89.05 & 7.66 & 3.28 & 27.24 & 19.81 & 8.50 & 23.55 \\ \bottomrule
\end{tabular}%
}
\caption{Error rates, incorrect choice distributions, and local negation confusion rates for the \textbf{Mistral} family under zero-shot, few-shot, and SFT conditions, evaluated in the \textbf{option-selection setting} using \textbf{definition instruction.}}
\label{tab:neg_mistral_def_option}
\end{table*}

\begin{table*}[htbp]
\centering
\resizebox{0.9\textwidth}{!}{%
\begin{tabular}{@{}cccr|rrr|rrrr@{}}
\toprule
\multirow{2}{*}{\textbf{Model}} & \multirow{2}{*}{\textbf{\begin{tabular}[c]{@{}c@{}}Training\\ Setting\end{tabular}}} & \multirow{2}{*}{\textbf{N Shot}} & \multicolumn{1}{c|}{\multirow{2}{*}{\textbf{\begin{tabular}[c]{@{}c@{}}Error \\ Rate\\ (1-acc)\end{tabular}}}} & \multicolumn{3}{c|}{\textbf{Incorrect Choice Distribution}} & \multicolumn{4}{c}{\textbf{Local Negation Confusion Rate}} \\ \cmidrule(l){5-11} 
 &  &  & \multicolumn{1}{c|}{} & \multicolumn{1}{c}{\textbf{\begin{tabular}[c]{@{}c@{}}Local\\ Negation\\ (\%)\end{tabular}}} & \multicolumn{1}{c}{\textbf{\begin{tabular}[c]{@{}c@{}}Contra-\\ diction\\ (\%)\end{tabular}}} & \multicolumn{1}{c|}{\textbf{\begin{tabular}[c]{@{}c@{}}Para-\\ phrase\\ (\%)\end{tabular}}} & \multicolumn{1}{c}{\textbf{\begin{tabular}[c]{@{}c@{}}Relative\\ Clause\\ (\%)\end{tabular}}} & \multicolumn{1}{c}{\textbf{\begin{tabular}[c]{@{}c@{}}Participle\\ Clause\\ (\%)\end{tabular}}} & \multicolumn{1}{c}{\textbf{\begin{tabular}[c]{@{}c@{}}Compound\\ Sentence\\ (\%)\end{tabular}}} & \multicolumn{1}{c}{\textbf{\begin{tabular}[c]{@{}c@{}}Adverbial\\ Clause\\ (\%)\end{tabular}}} \\ \midrule
\multicolumn{1}{l}{\multirow{5}{*}{\textbf{Mistral-7B-v0.3}}} & \multicolumn{1}{l}{\multirow{4}{*}{\textbf{baseline}}} & \multicolumn{1}{l}{\textbf{zero-shot}} & 0.521 & 74.12 & 20.24 & 5.63 & 26.60 & 32.14 & 55.44 & 45.81 \\
\multicolumn{1}{l}{} & \multicolumn{1}{l}{} & \multicolumn{1}{l}{\textbf{1-shot}} & 0.530 & 78.29 & 17.81 & 3.89 & 29.81 & 32.47 & 61.22 & 48.39 \\
\multicolumn{1}{l}{} & \multicolumn{1}{l}{} & \multicolumn{1}{l}{\textbf{5-shot}} & 0.432 & 82.94 & 15.23 & 1.83 & 23.08 & 24.35 & 52.38 & 48.71 \\
\multicolumn{1}{l}{} & \multicolumn{1}{l}{} & \multicolumn{1}{l}{\textbf{10-shot}} & 0.398 & 83.07 & 14.54 & 2.39 & 21.47 & 24.35 & 47.28 & 43.87 \\ \cmidrule(l){2-11} 
\multicolumn{1}{l}{} & \multicolumn{1}{l}{\textbf{after SFT}} & \multicolumn{1}{l}{\textbf{zero-shot}} & 0.208 & 87.02 & 11.45 & 1.53 & 9.62 & 12.34 & 25.51 & 27.42 \\ \midrule
\multicolumn{1}{l}{\multirow{5}{*}{\textbf{\begin{tabular}[c]{@{}l@{}}Mistral-7B-Instruct-\\ v0.3\end{tabular}}}} & \multicolumn{1}{l}{\multirow{4}{*}{\textbf{baseline}}} & \multicolumn{1}{l}{\textbf{zero-shot}} & 0.338 & 68.78 & 30.52 & 0.70 & 17.31 & 18.18 & 23.47 & 36.77 \\
\multicolumn{1}{l}{} & \multicolumn{1}{l}{} & \multicolumn{1}{l}{\textbf{1-shot}} & 0.337 & 69.88 & 28.94 & 1.18 & 15.06 & 19.16 & 31.63 & 31.61 \\
\multicolumn{1}{l}{} & \multicolumn{1}{l}{} & \multicolumn{1}{l}{\textbf{5-shot}} & 0.296 & 73.19 & 26.54 & 0.27 & 13.14 & 16.56 & 28.57 & 31.29 \\
\multicolumn{1}{l}{} & \multicolumn{1}{l}{} & \multicolumn{1}{l}{\textbf{10-shot}} & 0.282 & 73.03 & 26.12 & 0.84 & 13.14 & 13.64 & 25.85 & 32.58 \\ \cmidrule(l){2-11} 
\multicolumn{1}{l}{} & \multicolumn{1}{l}{\textbf{after SFT}} & \multicolumn{1}{l}{\textbf{zero-shot}} & 0.205 & 83.40 & 15.06 & 1.54 & 8.01 & 9.42 & 23.81 & 29.68 \\ \midrule
\multirow{4}{*}{\textbf{\begin{tabular}[c]{@{}c@{}}Mistral-Nemo-\\ Base-2407 (12B)\end{tabular}}} & \multirow{4}{*}{\textbf{baseline}} & \textbf{zero-shot} & 0.535 & 71.81 & 24.04 & 4.15 & 25.96 & 30.84 & 53.74 & 48.39 \\
 &  & \textbf{1-shot} & 0.504 & 75.28 & 20.79 & 3.94 & 27.88 & 30.19 & 53.40 & 45.48 \\
 &  & \textbf{5-shot} & 0.409 & 78.68 & 18.41 & 2.91 & 21.47 & 24.35 & 42.52 & 44.84 \\
 &  & \textbf{10-shot} & 0.355 & 81.43 & 17.00 & 1.57 & 19.87 & 22.40 & 33.33 & 43.55 \\ \midrule
\multirow{4}{*}{\textbf{\begin{tabular}[c]{@{}c@{}}Mistral-Nemo-\\ Instruct-2407 (12B)\end{tabular}}} & \multirow{4}{*}{\textbf{baseline}} & \textbf{zero-shot} & 0.489 & 71.47 & 25.28 & 3.24 & 24.36 & 29.55 & 48.30 & 42.58 \\
 &  & \textbf{1-shot} & 0.450 & 75.35 & 22.36 & 2.29 & 21.79 & 26.95 & 45.58 & 46.13 \\
 &  & \textbf{5-shot} & 0.362 & 78.73 & 19.96 & 1.32 & 18.91 & 20.45 & 36.05 & 42.26 \\
 &  & \textbf{10-shot} & 0.326 & 79.56 & 19.46 & 0.97 & 16.99 & 19.16 & 30.27 & 40.65 \\ \midrule
\multirow{4}{*}{\textbf{\begin{tabular}[c]{@{}c@{}}Mistral-Small-24B-\\ Base-2501\end{tabular}}} & \multirow{4}{*}{\textbf{baseline}} & \textbf{zero-shot} & 0.484 & 70.49 & 23.44 & 6.07 & 25.00 & 31.82 & 50.00 & 34.52 \\
 &  & \textbf{1-shot} & 0.466 & 76.32 & 18.91 & 4.77 & 25.64 & 28.25 & 54.08 & 39.35 \\
 &  & \textbf{5-shot} & 0.385 & 81.24 & 15.26 & 3.51 & 19.23 & 23.70 & 41.84 & 44.52 \\
 &  & \textbf{10-shot} & 0.339 & 83.14 & 14.99 & 1.87 & 17.31 & 21.10 & 36.05 & 41.94 \\ \midrule
\multirow{4}{*}{\textbf{\begin{tabular}[c]{@{}c@{}}Mistral-Small-24B-\\ Instruct-2501\end{tabular}}} & \multirow{4}{*}{\textbf{baseline}} & \textbf{zero-shot} & 0.420 & 76.94 & 20.60 & 2.46 & 25.32 & 32.79 & 44.22 & 31.29 \\
 &  & \textbf{1-shot} & 0.360 & 77.97 & 20.04 & 1.98 & 21.79 & 24.03 & 38.78 & 31.61 \\
 &  & \textbf{5-shot} & 0.328 & 82.57 & 16.95 & 0.48 & 17.95 & 19.48 & 35.03 & 39.35 \\
 &  & \textbf{10-shot} & 0.297 & 82.62 & 16.31 & 1.07 & 14.74 & 17.21 & 30.27 & 39.03 \\ \bottomrule
\end{tabular}%
}
\caption{Error rates, incorrect choice distributions, and local negation confusion rates for the \textbf{Mistral} family under zero-shot, few-shot, and SFT conditions, evaluated in the \textbf{completion-based setting} using \textbf{detailed instruction.}}
\label{tab:neg_mistral_detail_completion}
\end{table*}

\begin{table*}[htbp]
\centering
\resizebox{0.9\textwidth}{!}{%
\begin{tabular}{@{}cccr|rrr|rrrr@{}}
\toprule
\multirow{2}{*}{\textbf{Model}} & \multirow{2}{*}{\textbf{\begin{tabular}[c]{@{}c@{}}Training\\ Setting\end{tabular}}} & \multirow{2}{*}{\textbf{N Shot}} & \multicolumn{1}{c|}{\multirow{2}{*}{\textbf{\begin{tabular}[c]{@{}c@{}}Error \\ Rate\\ (1-acc)\end{tabular}}}} & \multicolumn{3}{c|}{\textbf{Incorrect Choice Distribution}} & \multicolumn{4}{c}{\textbf{Local Negation Confusion Rate}} \\ \cmidrule(l){5-11} 
 &  &  & \multicolumn{1}{c|}{} & \multicolumn{1}{c}{\textbf{\begin{tabular}[c]{@{}c@{}}Local\\ Negation\\ (\%)\end{tabular}}} & \multicolumn{1}{c}{\textbf{\begin{tabular}[c]{@{}c@{}}Contra-\\ diction\\ (\%)\end{tabular}}} & \multicolumn{1}{c|}{\textbf{\begin{tabular}[c]{@{}c@{}}Para-\\ phrase\\ (\%)\end{tabular}}} & \multicolumn{1}{c}{\textbf{\begin{tabular}[c]{@{}c@{}}Relative\\ Clause\\ (\%)\end{tabular}}} & \multicolumn{1}{c}{\textbf{\begin{tabular}[c]{@{}c@{}}Participle\\ Clause\\ (\%)\end{tabular}}} & \multicolumn{1}{c}{\textbf{\begin{tabular}[c]{@{}c@{}}Compound\\ Sentence\\ (\%)\end{tabular}}} & \multicolumn{1}{c}{\textbf{\begin{tabular}[c]{@{}c@{}}Adverbial\\ Clause\\ (\%)\end{tabular}}} \\ \midrule
\multicolumn{1}{l}{\multirow{5}{*}{\textbf{Mistral-7B-v0.3}}} & \multicolumn{1}{l}{\multirow{4}{*}{\textbf{baseline}}} & \multicolumn{1}{l}{\textbf{zero-shot}} & 0.498 & 59.39 & 15.13 & 25.48 & 24.04 & 25.97 & 43.88 & 28.71 \\
\multicolumn{1}{l}{} & \multicolumn{1}{l}{} & \multicolumn{1}{l}{\textbf{1-shot}} & 0.543 & 48.03 & 9.64 & 42.34 & 26.28 & 18.51 & 36.39 & 26.77 \\
\multicolumn{1}{l}{} & \multicolumn{1}{l}{} & \multicolumn{1}{l}{\textbf{5-shot}} & 0.364 & 71.02 & 7.63 & 21.35 & 24.36 & 20.45 & 30.95 & 30.97 \\
\multicolumn{1}{l}{} & \multicolumn{1}{l}{} & \multicolumn{1}{l}{\textbf{10-shot}} & 0.339 & 72.20 & 7.71 & 20.09 & 22.76 & 20.13 & 25.85 & 32.26 \\ \cmidrule(l){2-11} 
\multicolumn{1}{l}{} & \multicolumn{1}{l}{\textbf{after SFT}} & \multicolumn{1}{l}{\textbf{zero-shot}} & 0.523 & 51.75 & 23.37 & 24.89 & 26.92 & 28.25 & 30.95 & 25.48 \\ \midrule
\multicolumn{1}{l}{\multirow{5}{*}{\textbf{\begin{tabular}[c]{@{}l@{}}Mistral-7B-Instruct-\\ v0.3\end{tabular}}}} & \multicolumn{1}{l}{\multirow{4}{*}{\textbf{baseline}}} & \multicolumn{1}{l}{\textbf{zero-shot}} & 0.366 & 80.04 & 10.63 & 9.33 & 31.73 & 20.78 & 35.71 & 32.58 \\
\multicolumn{1}{l}{} & \multicolumn{1}{l}{} & \multicolumn{1}{l}{\textbf{1-shot}} & 0.356 & 71.71 & 16.48 & 11.80 & 26.60 & 24.35 & 30.61 & 23.87 \\
\multicolumn{1}{l}{} & \multicolumn{1}{l}{} & \multicolumn{1}{l}{\textbf{5-shot}} & 0.355 & 69.42 & 16.07 & 14.51 & 24.68 & 21.10 & 31.97 & 24.19 \\
\multicolumn{1}{l}{} & \multicolumn{1}{l}{} & \multicolumn{1}{l}{\textbf{10-shot}} & 0.355 & 71.21 & 16.52 & 12.28 & 25.32 & 23.05 & 31.97 & 24.19 \\ \cmidrule(l){2-11} 
\multicolumn{1}{l}{} & \multicolumn{1}{l}{\textbf{after SFT}} & \multicolumn{1}{l}{\textbf{zero-shot}} & 0.290 & 80.27 & 15.62 & 4.11 & 23.72 & 20.45 & 30.61 & 21.29 \\ \midrule
\multirow{4}{*}{\textbf{\begin{tabular}[c]{@{}c@{}}Mistral-Nemo-\\ Base-2407 (12B)\end{tabular}}} & \multirow{4}{*}{\textbf{baseline}} & \textbf{zero-shot} & 0.578 & 66.67 & 14.54 & 18.79 & 37.18 & 40.91 & 47.62 & 33.55 \\
 &  & \textbf{1-shot} & 0.497 & 82.30 & 11.16 & 6.54 & 38.78 & 35.06 & 55.10 & 40.32 \\
 &  & \textbf{5-shot} & 0.443 & 88.35 & 5.91 & 5.73 & 38.46 & 30.52 & 49.32 & 43.23 \\
 &  & \textbf{10-shot} & 0.400 & 89.09 & 6.75 & 4.17 & 33.97 & 28.25 & 40.48 & 44.19 \\ \midrule
\multirow{4}{*}{\textbf{\begin{tabular}[c]{@{}c@{}}Mistral-Nemo-\\ Instruct-2407 (12B)\end{tabular}}} & \multirow{4}{*}{\textbf{baseline}} & \textbf{zero-shot} & 0.374 & 85.56 & 10.83 & 3.61 & 29.49 & 27.92 & 42.86 & 31.94 \\
 &  & \textbf{1-shot} & 0.364 & 86.93 & 11.33 & 1.74 & 30.13 & 27.92 & 39.12 & 33.55 \\
 &  & \textbf{5-shot} & 0.344 & 91.01 & 7.83 & 1.15 & 31.41 & 26.95 & 39.80 & 31.29 \\
 &  & \textbf{10-shot} & 0.340 & 90.91 & 8.16 & 0.93 & 34.29 & 27.92 & 36.05 & 29.35 \\ \midrule
\multirow{4}{*}{\textbf{\begin{tabular}[c]{@{}c@{}}Mistral-Small-24B-\\ Base-2501\end{tabular}}} & \multirow{4}{*}{\textbf{baseline}} & \textbf{zero-shot} & 0.425 & 61.57 & 17.91 & 20.52 & 31.41 & 25.00 & 29.25 & 22.26 \\
 &  & \textbf{1-shot} & 0.357 & 70.00 & 21.78 & 8.22 & 29.81 & 20.13 & 29.59 & 23.55 \\
 &  & \textbf{5-shot} & 0.265 & 83.23 & 10.78 & 5.99 & 24.04 & 20.13 & 16.33 & 30.00 \\
 &  & \textbf{10-shot} & 0.215 & 89.67 & 6.64 & 3.69 & 22.12 & 17.53 & 8.84 & 30.32 \\ \midrule
\multirow{4}{*}{\textbf{\begin{tabular}[c]{@{}c@{}}Mistral-Small-24B-\\ Instruct-2501\end{tabular}}} & \multirow{4}{*}{\textbf{baseline}} & \textbf{zero-shot} & 0.271 & 80.41 & 13.16 & 6.43 & 32.37 & 22.73 & 11.22 & 22.90 \\
 &  & \textbf{1-shot} & 0.243 & 83.01 & 13.4 & 3.59 & 27.56 & 21.43 & 14.97 & 18.71 \\
 &  & \textbf{5-shot} & 0.240 & 87.09 & 9.93 & 2.98 & 27.88 & 21.10 & 12.93 & 23.55 \\
 &  & \textbf{10-shot} & 0.195 & 87.40 & 9.35 & 3.25 & 24.36 & 16.88 & 4.76 & 23.55 \\ \bottomrule
\end{tabular}%
}
\caption{Error rates, incorrect choice distributions, and local negation confusion rates for the \textbf{Mistral} family under zero-shot, few-shot, and SFT conditions, evaluated in the \textbf{option-selection setting} using \textbf{detailed instruction.}}
\label{tab:neg_mistral_detail_option}
\end{table*}

\begin{table*}[htbp]
\centering
\resizebox{\textwidth}{!}{%
\begin{tabular}{@{}lllrr|rrr|rrrr@{}}
\toprule
\multicolumn{1}{c}{\multirow{2}{*}{\textbf{Model}}} & \multicolumn{1}{c}{\multirow{2}{*}{\textbf{\begin{tabular}[c]{@{}c@{}}Training\\ Setting\end{tabular}}}} & \multicolumn{1}{c}{\multirow{2}{*}{\textbf{N Shot}}} & \multicolumn{1}{c}{\multirow{2}{*}{\textbf{\begin{tabular}[c]{@{}c@{}}Error \\ Rate\\ (1-acc)\end{tabular}}}} & \multicolumn{1}{c|}{\multirow{2}{*}{\textbf{\begin{tabular}[c]{@{}c@{}}Answer\\ Format\\ Wrong\end{tabular}}}} & \multicolumn{3}{c|}{\textbf{Incorrect Choice Distribution}} & \multicolumn{4}{c}{\textbf{Local Negation Confusion Rate}} \\ \cmidrule(l){6-12} 
\multicolumn{1}{c}{} & \multicolumn{1}{c}{} & \multicolumn{1}{c}{} & \multicolumn{1}{c}{} & \multicolumn{1}{c|}{} & \multicolumn{1}{c}{\textbf{\begin{tabular}[c]{@{}c@{}}Local\\ Negation\\ (\%)\end{tabular}}} & \multicolumn{1}{c}{\textbf{\begin{tabular}[c]{@{}c@{}}Contra-\\ diction\\ (\%)\end{tabular}}} & \multicolumn{1}{c|}{\textbf{\begin{tabular}[c]{@{}c@{}}Para-\\ phrase\\ (\%)\end{tabular}}} & \multicolumn{1}{c}{\textbf{\begin{tabular}[c]{@{}c@{}}Relative\\ Clause\\ (\%)\end{tabular}}} & \multicolumn{1}{c}{\textbf{\begin{tabular}[c]{@{}c@{}}Participle\\ Clause\\ (\%)\end{tabular}}} & \multicolumn{1}{c}{\textbf{\begin{tabular}[c]{@{}c@{}}Compound\\ Sentence\\ (\%)\end{tabular}}} & \multicolumn{1}{c}{\textbf{\begin{tabular}[c]{@{}c@{}}Adverbial\\ Clause\\ (\%)\end{tabular}}} \\ \midrule
\multirow{4}{*}{\textbf{GPT-4o mini}} & \multirow{4}{*}{\textbf{baseline}} & \textbf{zero-shot} & 0.287 & 0 & 68.23 & 31.77 & 0 & 19.87 & 11.36 & 31.97 & 18.06 \\
 &  & \textbf{1-shot} & 0.251 & 0 & 68.99 & 30.38 & 0.63 & 20.51 & 10.71 & 26.87 & 13.55 \\
 &  & \textbf{5-shot} & 0.231 & 0 & 54.64 & 45.02 & 0.34 & 18.27 & 5.52 & 17.69 & 10.65 \\
 &  & \textbf{10-shot} & 0.205 & 0 & 52.51 & 47.10 & 0.39 & 16.67 & 7.14 & 11.90 & 8.71 \\ \midrule
\multirow{4}{*}{\textbf{GPT-4o}} & \multirow{4}{*}{\textbf{baseline}} & \textbf{zero-shot} & 0.218 & 0 & 80.73 & 19.27 & 0 & 18.91 & 8.44 & 33.67 & 12.26 \\
 &  & \textbf{1-shot} & 0.209 & 0 & 88.26 & 11.36 & 0.38 & 23.08 & 8.77 & 27.55 & 17.10 \\
 &  & \textbf{5-shot} & 0.198 & 0 & 87.55 & 12.05 & 0.40 & 22.12 & 11.36 & 17.35 & 20.32 \\
 &  & \textbf{10-shot} & 0.175 & 0 & 85.45 & 14.09 & 0.45 & 20.19 & 10.39 & 12.24 & 18.39 \\ \midrule
\multirow{4}{*}{\textbf{GPT-4.1 mini}} & \multirow{4}{*}{\textbf{baseline}} & \textbf{zero-shot} & 0.244 & 0 & 90.88 & 9.12 & 0 & 14.42 & 5.19 & 51.36 & 21.61 \\
 &  & \textbf{1-shot} & 0.204 & 0 & 88.33 & 11.67 & 0 & 18.59 & 7.14 & 30.27 & 18.71 \\
 &  & \textbf{5-shot} & 0.159 & 0 & 84.00 & 15.50 & 0.50 & 15.71 & 5.52 & 15.31 & 18.39 \\
 &  & \textbf{10-shot} & 0.151 & 0 & 77.89 & 21.58 & 0.53 & 13.78 & 5.84 & 8.84 & 19.68 \\ \midrule
\multirow{4}{*}{\textbf{GPT-4.1}} & \multirow{4}{*}{\textbf{baseline}} & \textbf{zero-shot} & 0.119 & 0 & 78.67 & 20.00 & 1.33 & 10.90 & 6.49 & 15.65 & 5.81 \\
 &  & \textbf{1-shot} & 0.136 & 0 & 81.87 & 18.13 & 0 & 17.31 & 9.09 & 10.88 & 8.39 \\
 &  & \textbf{5-shot} & 0.128 & 0 & 81.99 & 18.01 & 0 & 18.27 & 9.42 & 5.10 & 10.00 \\
 &  & \textbf{10-shot} & 0.109 & 0 & 79.71 & 19.57 & 0.72 & 15.06 & 8.44 & 2.04 & 10.00 \\ \midrule
\multirow{4}{*}{\textbf{Haiku 4.5}} & \multirow{4}{*}{\textbf{baseline}} & \textbf{zero-shot} & 0.228 & 0 & 88.54 & 11.46 & 0 & 22.12 & 15.26 & 38.78 & 8.06 \\
 &  & \textbf{1-shot} & 0.226 & 0 & 89.12 & 9.82 & 1.05 & 24.36 & 15.26 & 32.65 & 11.29 \\
 &  & \textbf{5-shot} & 0.190 & 0 & 90.79 & 9.21 & 0 & 29.49 & 17.53 & 13.61 & 10.00 \\
 &  & \textbf{10-shot} & 0.172 & 0 & 91.71 & 7.83 & 0.46 & 25.64 & 17.86 & 7.48 & 13.55 \\ \midrule
\multirow{4}{*}{\textbf{Sonnet 4.5}} & \multirow{4}{*}{\textbf{baseline}} & \textbf{zero-shot} & 0.125 & 0 & 85.35 & 12.74 & 1.91 & 16.35 & 14.61 & 8.50 & 4.19 \\
 &  & \textbf{1-shot} & 0.122 & 0 & 92.86 & 6.49 & 0.65 & 19.23 & 16.56 & 6.12 & 4.52 \\
 &  & \textbf{5-shot} & 0.127 & 0 & 81.25 & 12.50 & 6.25 & 16.67 & 12.99 & 4.76 & 7.74 \\
 &  & \textbf{10-shot} & 0.118 & 15 & 82.09 & 10.45 & 7.46 & 13.46 & 9.42 & 5.44 & 7.42 \\ \bottomrule
\end{tabular}%
}
\caption{Error rates, incorrect choice distributions, and local negation confusion rates for the \textbf{API} models under zero-shot and few-shot conditions, using \textbf{definition instruction.}}
\label{tab:neg_api_def}
\end{table*}

\begin{table*}[htbp]
\centering
\resizebox{\textwidth}{!}{%
\begin{tabular}{@{}lllrr|rrr|rrrr@{}}
\toprule
\multicolumn{1}{c}{\multirow{2}{*}{\textbf{Model}}} & \multicolumn{1}{c}{\multirow{2}{*}{\textbf{\begin{tabular}[c]{@{}c@{}}Training\\ Setting\end{tabular}}}} & \multicolumn{1}{c}{\multirow{2}{*}{\textbf{N Shot}}} & \multicolumn{1}{c}{\multirow{2}{*}{\textbf{\begin{tabular}[c]{@{}c@{}}Error \\ Rate\\ (1-acc)\end{tabular}}}} & \multicolumn{1}{c|}{\multirow{2}{*}{\textbf{\begin{tabular}[c]{@{}c@{}}Answer\\ Format\\ Wrong\end{tabular}}}} & \multicolumn{3}{c|}{\textbf{Incorrect Choice Distribution}} & \multicolumn{4}{c}{\textbf{Local Negation Confusion Rate}} \\ \cmidrule(l){6-12} 
\multicolumn{1}{c}{} & \multicolumn{1}{c}{} & \multicolumn{1}{c}{} & \multicolumn{1}{c}{} & \multicolumn{1}{c|}{} & \multicolumn{1}{c}{\textbf{\begin{tabular}[c]{@{}c@{}}Local\\ Negation\\ (\%)\end{tabular}}} & \multicolumn{1}{c}{\textbf{\begin{tabular}[c]{@{}c@{}}Contra-\\ diction\\ (\%)\end{tabular}}} & \multicolumn{1}{c|}{\textbf{\begin{tabular}[c]{@{}c@{}}Para-\\ phrase\\ (\%)\end{tabular}}} & \multicolumn{1}{c}{\textbf{\begin{tabular}[c]{@{}c@{}}Relative\\ Clause\\ (\%)\end{tabular}}} & \multicolumn{1}{c}{\textbf{\begin{tabular}[c]{@{}c@{}}Participle\\ Clause\\ (\%)\end{tabular}}} & \multicolumn{1}{c}{\textbf{\begin{tabular}[c]{@{}c@{}}Compound\\ Sentence\\ (\%)\end{tabular}}} & \multicolumn{1}{c}{\textbf{\begin{tabular}[c]{@{}c@{}}Adverbial\\ Clause\\ (\%)\end{tabular}}} \\ \midrule
\multirow{4}{*}{\textbf{GPT-4o mini}} & \multirow{4}{*}{\textbf{baseline}} & \textbf{zero-shot} & 0.244 & 0 & 82.14 & 17.21 & 0.65 & 14.42 & 12.66 & 38.10 & 18.39 \\
 &  & \textbf{1-shot} & 0.226 & 0 & 81.05 & 17.89 & 1.05 & 19.55 & 8.12 & 33.33 & 15.16 \\
 &  & \textbf{5-shot} & 0.224 & 0 & 63.60 & 35.69 & 0.71 & 19.55 & 4.55 & 19.73 & 15.16 \\
 &  & \textbf{10-shot} & 0.210 & 0 & 60.38 & 39.25 & 0.38 & 16.99 & 7.14 & 15.65 & 12.58 \\ \midrule
\multirow{4}{*}{\textbf{GPT-4o}} & \multirow{4}{*}{\textbf{baseline}} & \textbf{zero-shot} & 0.137 & 0 & 74.57 & 25.43 & 0 & 12.82 & 7.47 & 12.24 & 9.68 \\
 &  & \textbf{1-shot} & 0.147 & 0 & 79.46 & 20.00 & 0.54 & 18.27 & 10.06 & 13.27 & 6.45 \\
 &  & \textbf{5-shot} & 0.172 & 0 & 84.79 & 13.82 & 1.38 & 19.87 & 13.31 & 12.93 & 13.87 \\
 &  & \textbf{10-shot} & 0.185 & 0 & 83.69 & 15.45 & 0.86 & 24.68 & 12.01 & 11.22 & 15.48 \\ \midrule
\multirow{4}{*}{\textbf{GPT-4.1 mini}} & \multirow{4}{*}{\textbf{baseline}} & \textbf{zero-shot} & 0.098 & 0 & 90.32 & 9.68 & 0 & 3.53 & 3.57 & 16.67 & 13.23 \\
 &  & \textbf{1-shot} & 0.133 & 0 & 84.52 & 15.48 & 0 & 14.42 & 5.84 & 18.37 & 8.06 \\
 &  & \textbf{5-shot} & 0.124 & 1 & 83.23 & 15.48 & 1.29 & 12.50 & 4.87 & 11.56 & 13.23 \\
 &  & \textbf{10-shot} & 0.130 & 0 & 75.61 & 23.78 & 0.61 & 12.50 & 4.22 & 11.56 & 12.26 \\ \midrule
\multirow{4}{*}{\textbf{GPT-4.1}} & \multirow{4}{*}{\textbf{baseline}} & \textbf{zero-shot} & 0.064 & 0 & 75.31 & 24.69 & 0 & 6.73 & 3.90 & 6.12 & 3.23 \\
 &  & \textbf{1-shot} & 0.082 & 0 & 91.26 & 8.74 & 0 & 13.14 & 8.12 & 4.42 & 4.84 \\
 &  & \textbf{5-shot} & 0.098 & 0 & 81.30 & 18.70 & 0 & 13.14 & 9.09 & 3.74 & 6.45 \\
 &  & \textbf{10-shot} & 0.098 & 0 & 84.68 & 15.32 & 0 & 13.78 & 8.44 & 1.70 & 10.00 \\ \midrule
\multirow{4}{*}{\textbf{Haiku 4.5}} & \multirow{4}{*}{\textbf{baseline}} & \textbf{zero-shot} & 0.135 & 1 & 89.94 & 10.06 & 0 & 16.99 & 15.58 & 13.27 & 3.87 \\
 &  & \textbf{1-shot} & 0.128 & 0 & 87.58 & 9.32 & 3.11 & 16.67 & 12.01 & 11.56 & 5.81 \\
 &  & \textbf{5-shot} & 0.126 & 0 & 93.71 & 5.66 & 0.63 & 22.12 & 14.29 & 6.80 & 5.16 \\
 &  & \textbf{10-shot} & 0.118 & 0 & 94.63 & 5.37 & 0 & 19.55 & 13.64 & 5.10 & 7.42 \\ \midrule
\multirow{4}{*}{\textbf{Sonnet 4.5}} & \multirow{4}{*}{\textbf{baseline}} & \textbf{zero-shot} & 0.085 & 1 & 91.51 & 7.55 & 0.94 & 10.58 & 12.34 & 6.12 & 2.58 \\
 &  & \textbf{1-shot} & 0.083 & 0 & 91.43 & 8.57 & 0 & 12.50 & 13.96 & 2.04 & 2.58 \\
 &  & \textbf{5-shot} & 0.121 & 26 & 82.68 & 14.17 & 3.15 & 14.74 & 12.01 & 3.74 & 3.55 \\
 &  & \textbf{10-shot} & 0.188 & 150 & 75.86 & 13.79 & 10.34 & 8.65 & 7.14 & 1.70 & 3.87 \\ \bottomrule
\end{tabular}%
}
\caption{Error rates, incorrect choice distributions, and local negation confusion rates for the \textbf{API} models under zero-shot and few-shot conditions, using \textbf{detailed instruction.}}
\label{tab:neg_api_detail}
\end{table*}

\end{document}